\documentclass{article} %
\PassOptionsToPackage{numbers, compress}{natbib}
\usepackage{iclr2027_conference,times}
\iclrfinalcopy

\usepackage{amsmath,amsfonts,bm}

\def\eqref#1{equation~\ref{#1}}

\def\1{\bm{1}}

\DeclareMathAlphabet{\mathsfit}{\encodingdefault}{\sfdefault}{m}{sl}
\SetMathAlphabet{\mathsfit}{bold}{\encodingdefault}{\sfdefault}{bx}{n}

\usepackage{graphicx} \graphicspath{{figures/}}
\usepackage{amsmath,amssymb,mathabx,mathtools,amsthm,nicefrac}
\usepackage[pagebackref,breaklinks,colorlinks]{hyperref}
\usepackage[capitalise,nameinlink]{cleveref}
\usepackage[skip=3pt,font=small]{subcaption}
\usepackage[skip=3pt,font=small]{caption}
\usepackage{acronym}
\usepackage{wrapfig}
\usepackage[dvipsnames,svgnames,x11names,table]{xcolor}
\usepackage{enumerate}
\usepackage{xspace}
\usepackage[misc]{ifsym}
\usepackage{titletoc}
\usepackage{verbatim,fancyvrb,listings}
\usepackage{booktabs,tabularx,colortbl,multirow,multicol,array,makecell,tabularray}
\usepackage{siunitx}
\usepackage{enumitem}

\makeatletter
\DeclareRobustCommand\onedot{\futurelet\@let@token\@onedot}
\def\@onedot{\ifx\@let@token.\else.\null\fi\xspace}
\def\eg{\emph{e.g}\onedot}

\makeatother

\newcommand{\dataset}{\texttt{OpenVidHD-Motion3D}\xspace}
\newcommand{\modelname}{%
  \raisebox{-0.3em}{\includegraphics[trim=250 0 250 0,clip,height=1.2em]{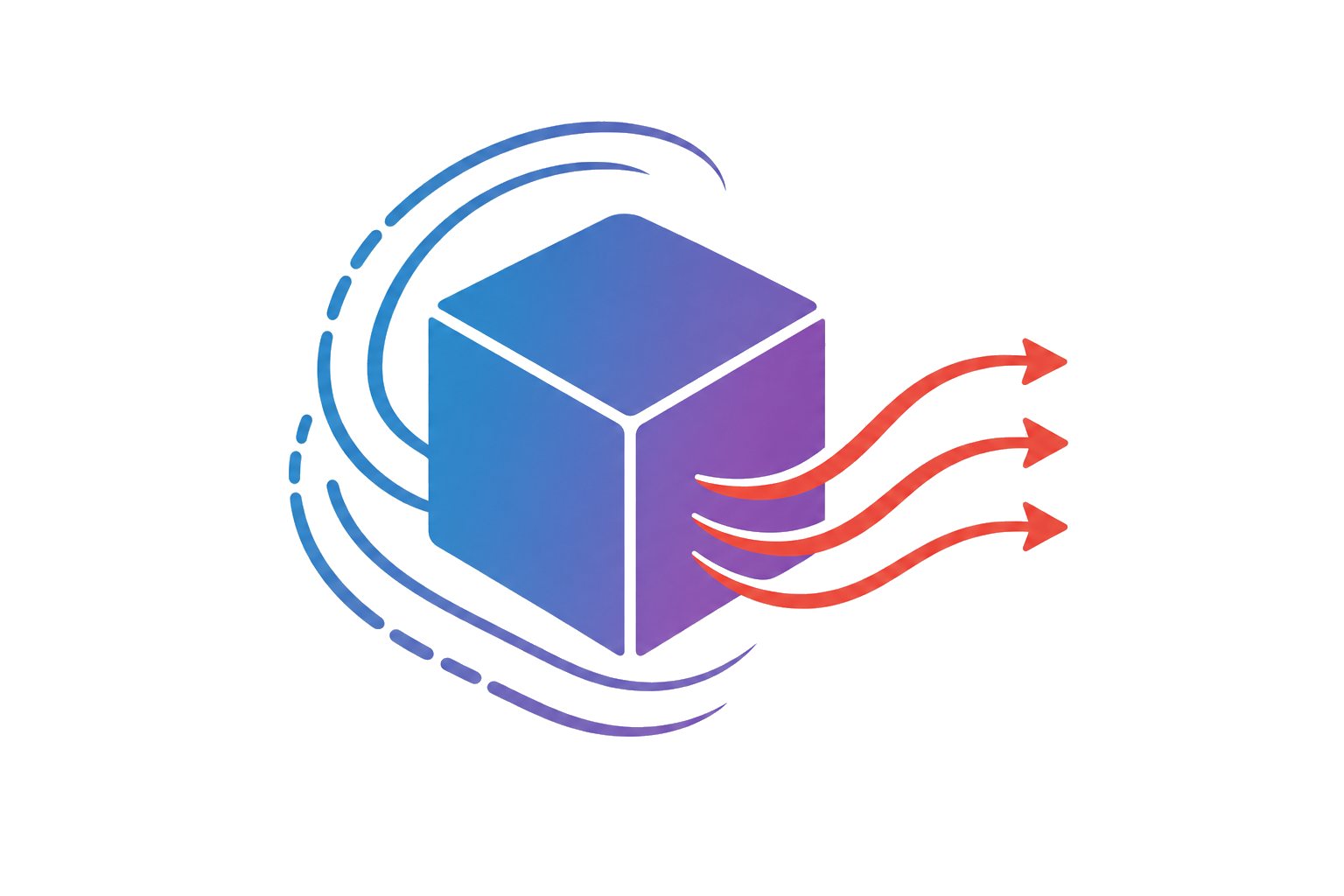}}\;%
  \textsf{\textbf{%
    \textcolor{rb01}{4}%
    \textcolor{rb02}{D}%
    \textcolor{rb03}{S}%
    \textcolor{rb04}{t}%
    \textcolor{rb05}{r}%
    \textcolor{rb06}{e}%
    \textcolor{rb07}{a}%
    \textcolor{rb08}{m}%
    \textcolor{rb09}{C}%
    \textcolor{rb10}{t}%
    \textcolor{rb11}{r}%
    \textcolor{rb12}{l}%
  }}\xspace%
}

\definecolor{rb01}{HTML}{FF6B6B}
\definecolor{rb02}{HTML}{FF8E53}
\definecolor{rb03}{HTML}{FFA940}
\definecolor{rb04}{HTML}{FFD93D}
\definecolor{rb05}{HTML}{6BCB77}
\definecolor{rb06}{HTML}{4ECDC4}
\definecolor{rb07}{HTML}{45B7D1}
\definecolor{rb08}{HTML}{4EA8DE}
\definecolor{rb09}{HTML}{5B86E5}
\definecolor{rb10}{HTML}{7C6CF0}
\definecolor{rb11}{HTML}{9B5DE5}
\definecolor{rb12}{HTML}{C77DFF}

\newcommand{\dashrule}[4]{%
  \par\noindent
  \hbox to #1{%
    \leaders\hbox to #3{\hss\rule{#4}{#2}\hss}\hfill
  }%
  \par
}

\title{\modelname: Interactive Video Generation with Online 4D Control}

\author{%
    \textbf{Shiqian Li}\textsuperscript{1,2},
    \textbf{Chenguo Lin}\textsuperscript{1},
    \textbf{Zhiguang Liu}\textsuperscript{2},
    \textbf{Yu Tang}\textsuperscript{2},
    \textbf{Jiarong Ou}\textsuperscript{2},
    \textbf{Rui Chen}\textsuperscript{2},
    \textbf{Yixin Zhu}\textsuperscript{1,\ \textrm{\Letter}}\\
    \textsuperscript{1} Peking University \quad
    \textsuperscript{2} Tencent Hunyuan \quad
    \textsuperscript{\textrm{\Letter}} Corresponding author\\
    \href{https://4dstreamctrl.github.io/}{https://4dstreamctrl.github.io/}
    \vspace{-18pt}
}

\begin{document}

\maketitle

\begin{figure}[ht!]
    \centering
    \includegraphics[width=\linewidth]{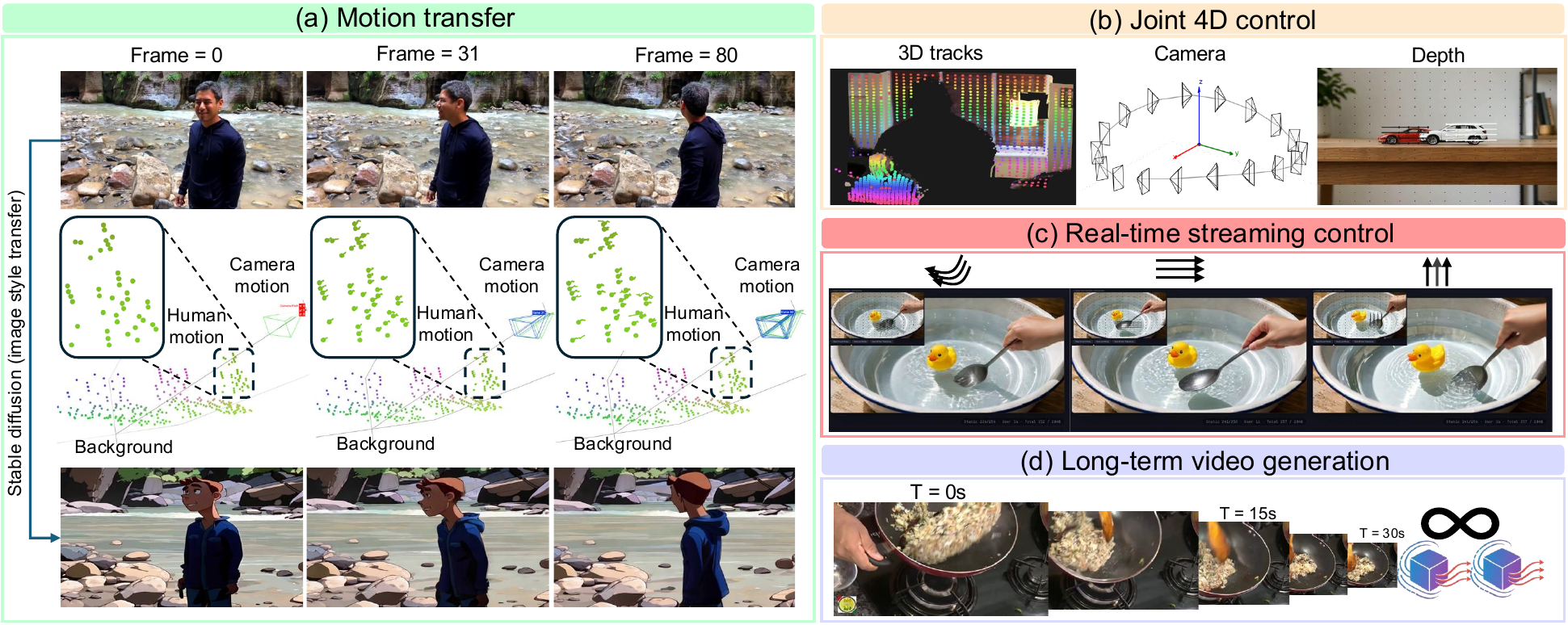}
    \caption{\textbf{Overview of \modelname capabilities.} \textbf{(a) Motion transfer}. Given a source video (top row), we extract decomposed 4D representations of camera motion, human motion, and background structure (middle row), which transfer to a new scene via image style transfer while preserving the original 3D-consistent motion (bottom row). \textbf{ (b) Joint 4D control}. 3D point tracks, camera parameters, and depth are unified into a single representation, enabling simultaneous control of camera and object motion. \textbf{(c) Real-time streaming control}. Users steer the generation with control signals (\eg, dragging objects) as the video streams. \textbf{(d) Long-term video generation}. Through causal autoregressive distillation, \modelname generates arbitrarily long, temporally coherent videos at a memory cost independent of length.}
    \label{fig:overview}
\end{figure}

\begin{abstract}
Generative video models now synthesize footage nearly indistinguishable from reality.
Their promise as interactive tools hinges on fine-grained control of how objects and the camera move over time, yet each existing approach captures only part of this: camera-parameter methods steer the viewpoint but cannot move objects, 2D-trajectory methods act in the image plane and ignore depth and occlusion, and recent 3D methods add geometry but run only offline at a fixed length.
In particular, none combines 3D-consistent control of both camera and objects with real-time, streaming generation.
Here we show that camera motion, object trajectories, and depth can be unified into a single 3D point-track representation, from which one model performs joint camera and object control, depth editing, and motion transfer in a single forward pass.
To learn this interface at scale, we mine in-the-wild video for 3D motion supervision, yielding \dataset, and encode it with a lightweight \emph{Geometric Motion Head} that plugs into a pretrained video diffusion model. Because this encoder is temporally separable, we distill the model into a causal streaming student that generates arbitrarily long video in four denoising steps at memory independent of length. This unified design surpasses prior camera-only, 2D, and offline-3D methods in motion-control precision while covering modalities they address only in isolation.
\modelname runs at 20\,FPS on a single high-end GPU for 480p video and stays temporally coherent over hundreds of frames, enabling, to our knowledge, interactive 4D-controllable streaming generation for the first time.
More broadly, grounding generation in explicit 3D geometry with efficient causal inference points toward interactive world models with closed-loop spatiotemporal control, from controllable simulators to real-time visual imagination for embodied agents.
\end{abstract}

\section{Introduction}

The ultimate goal of controllable video generation is to give users precise command over how objects move, how the camera behaves, and how these dynamics evolve, all in real time. Diffusion-based video models have advanced rapidly \citep{ho2022video,peebles2023scalable}, and open-weight systems such as Wan \citep{wan2025wan} and HunyuanVideo \citep{kong2024hunyuanvideo} now produce footage of striking realism. Yet their dominant interface, natural language with an optional reference image, underspecifies motion: ``the camera orbits the object while the cup slides left'' admits many unintended renderings, leaving practitioners in ``render-and-wait'' loops \citep{kong2024hunyuanvideo,wan2025wan,gu2025das,lee2026generative}. Work that injects explicit motion cues splits across two fronts that no method has united. The first is \emph{what the control signal can express}: camera-parameter methods \citep{he2024cameractrl,wang2024motionctrl} steer the viewpoint but cannot move objects; 2D-trajectory methods \citep{wu2024draganything,geng2025motion,teng2023dragavideo,li2025imageconductor} add object drag but, confined to the image plane, cannot encode depth or occlusion and break under parallax; only recent 3D-aware methods \citep{gu2025das,wang2025ati} restore full geometry via 3D point tracks. The second is \emph{how generation runs}: most of them, 3D-aware included \citep{gu2025das,lee2026generative}, are offline and fixed-length, producing a clip in one slow pass with no way to respond mid-generation. Missing is a method complete on the first front and interactive on the second: 3D-consistent control of both camera and objects, delivered as a real-time, steerable stream.

These shortfalls share one cause: no motion representation is at once geometrically complete and incrementally constructible. We present \modelname, built on the insight that \emph{3D point tracks and camera geometry can be unified into a single conditioning interface} expressing object trajectories, camera motion, and depth together. Projecting camera extrinsics onto background point tracks and encoding per-point depth alongside spatial coordinates makes this interface subsume camera, object, and depth control as special cases of one 3D-track signal, resolving both fronts. For expressiveness, a signal that jointly carries geometry, per-track identity, and depth dispels the ambiguity that defeats image-plane methods, letting one model, in a single forward pass, perform joint camera and object manipulation, cross-subject motion transfer, and depth-aware editing. For interactivity, the same representation admits lossless segment-by-segment encoding, the property that makes streaming possible.

To learn this interface at scale, we build \dataset, the first large-scale 3D motion dataset for in-the-wild video, by mining OpenVid-1M \citep{nan2025openvid} and annotating it with SpatialTrackerV2 \citep{xiao2025spatialtrackerv2}; after filtering it retains roughly 0.4M clips, each with video up to 1080p, a $32\times32$ grid of 3D point tracks, and per-frame camera intrinsics and extrinsics. A lightweight \emph{Geometric Motion Head} encodes each trajectory's coordinates $(x,y,z)$, a sinusoidal point-identity embedding, and monocular depth into dense features on the VAE latent grid, fused with noisy latents by channel concatenation before the Wan2.2 TI2V-5B DiT backbone \citep{wan2025wan}. Because its scatter-add rasterization and strided Conv3D run independently per chunk, encoding one segment is bit-identical to encoding the full sequence, so the offline teacher and streaming model share one motion pathway with no architectural change.

Sharing the encoder is necessary but not sufficient: the teacher's 50-step bidirectional denoising sees the whole clip at once, making it slow and unable to honor signals that arrive mid-playback. We therefore distill it into a causal autoregressive student following the self-forcing \citep{huang2025selfforcing} and DMD \citep{yin2024dmd,yin2024improved} paradigm adapted to our 3D-track interface. The student replaces bidirectional with block-wise causal attention \citep{chen2024diffusion}, keeps a fixed \emph{attention sink} \citep{xiao2023streamingllm} on the first-frame latents for global coherence and a \emph{local sliding window} over recent chunks, and denoises each chunk in 4 steps, a $12.5\times$ reduction. With KV-cache reuse and memory independent of video length, it streams arbitrarily long 480p video at 20\,FPS on a single high-end GPU: a user supplies an initial image and optional prompt, edits 3D trajectories on the fly, and the backend ingests the updated signals and returns decoded frames at interactive speed, closing the loop between human intent and generation for the first time in a 4D-controllable setting.

Extensive experiments show that \modelname attains state-of-the-art motion-control precision and 3D consistency, outperforming prior trajectory- and camera-conditioned methods \citep{wang2024motionctrl,motionstream2025,burgert2025gowiththeflow,gu2025das,wang2025ati,li2025imageconductor}, while its streaming student sustains coherent 350-frame clips (14.6\,s at 24\,FPS) at memory independent of length. Our contributions are threefold:
\begin{itemize}[nosep,leftmargin=*]
    \item A \textbf{unified 3D-track conditioning interface} expressing camera motion, object trajectories, and depth as one signal, subsuming camera, object, and depth control and enabling joint control, cross-subject motion transfer, and depth-aware editing in a single forward pass.
    \item \textbf{\dataset}, the first large-scale in-the-wild 3D motion dataset (roughly 0.4M clips with 3D tracks and per-frame camera parameters), with a lightweight, temporally separable \emph{Geometric Motion Head} that injects these signals into a pretrained video diffusion backbone.
    \item A \textbf{causal streaming distillation} converting the 50-step teacher into a 4-step student with constant-memory, arbitrarily long generation, yielding, to our knowledge, the first real-time 4D-controllable interactive video generation on a single GPU.
\end{itemize}

\section{Related Work}

\paragraph{Motion-conditioned video generation.}
Video diffusion offers a scalable paradigm for generative video modeling \citep{ho2022video,peebles2023scalable}, and recent systems inject explicit motion cues to overcome the underspecification of language-only conditioning. Representative directions include object motion control \citep{wang2024motionctrl}, entity-centric drag interfaces \citep{wu2024draganything}, point-based non-rigid editing \citep{teng2023dragavideo}, and trajectory prompts that steer diffusion with sparse or dense point tracks \citep{geng2025motion,wang2025ati,li2025imageconductor}; open-weight backbones such as Wan \citep{wan2025wan}, including the Wan2.2 TI2V-5B text-and-image-to-video checkpoint, make such interfaces practical to fine-tune. Most of these condition on \emph{image-plane} motion, however, and recent 3D-aware generators \citep{gu2025das} that instead use 3D point tracks remain restricted to offline, fixed-length synthesis. We study 3D trajectory conditioning with \textbf{joint object and camera control} to resolve the ambiguity between object and viewpoint motion, and target \textbf{interactive streaming} rather than one-shot clip generation.

\paragraph{Scalable 3D supervision from video.}
Recovering consistent geometry and cameras from monocular or multi-view video is long-standing, with classical structure-from-motion \citep{schonberger2016colmap} and learning-based correspondence and reconstruction models \citep{wang2024dust3r} improving robustness in the wild, while modern long-range trackers supply scalable point-dynamics signals \citep{karaev2024cotracker3} and large text--video datasets enable training at million-pair scale \citep{nan2025openvid}. We connect these ingredients by constructing \textbf{in-the-wild 3D track supervision with camera intrinsics and extrinsics}, turning internet video into structured geometric motion signals for training trajectory-conditioned generators.

\paragraph{Few-step distillation and streaming video generation.}
Iterative diffusion sampling is too costly for closed-loop interaction, motivating few-step distillation: progressive distillation \citep{salimans2022progressive} and distribution matching \citep{yin2024dmd,yin2024improved} compress a bidirectional teacher into a handful of denoising steps. In parallel, autoregressive and chunk-based generators \citep{yu2023magvit,kondratyuk2024videopoet,yin2025causvid,huang2025selfforcing} synthesize frame blocks sequentially over a rolling context to enable streaming rather than one-shot generation, with a retained \emph{attention sink} \citep{xiao2023streamingllm} stabilizing long rollouts. Closest to our setting, MotionStream \citep{motionstream2025} distills a streaming controllable generator, but conditions on 2D image-plane tracks and so inherits their depth and occlusion ambiguity. We instead distill a causal student over our \textbf{3D-track interface}, carrying 3D-consistent joint control of camera and objects into real-time streaming and moving closer to interactive world-model deployment \citep{pmlr-v235-bruce24a,yang2024unisim} than one-shot full-clip sampling.

\section{Method}

We aim to learn a video generator for 4D control that (i) is trained on in-the-wild 3D motion supervision mined from large-scale internet video, (ii) conditions a pretrained text-and-image-to-video (TI2V) diffusion backbone on 3D motion features (including depth), and (iii) is distilled into a causal streaming student for interactive control at high frame rates.

\subsection{3D Track Dataset Construction}\label{sec:data}

\paragraph{Source videos.}
We start from OpenVid-1M \citep{nan2025openvid}, a large collection of text--video pairs with diverse in-the-wild content. For each clip, we run a fixed offline pipeline to extract per-frame 3D point trajectories together with camera intrinsics and extrinsics, yielding scalable supervision without manual motion annotation.

\paragraph{3D tracking.}
We extract dense 3D tracks with SpatialTrackerV2 \citep{xiao2025spatialtrackerv2}, a feed-forward monocular 3D point tracker that jointly estimates scene geometry, camera motion, and point dynamics. For each clip, we sample query points on a $32\times32$ grid in the first frame and propagate them across all subsequent frames, yielding per-point trajectories in a consistent world coordinate system together with per-frame camera intrinsics $\mathbf{K}_t$ and world-to-camera extrinsics $\mathbf{T}_t$. We then transform each world-space point into the corresponding camera frame via $\mathbf{T}_t$ and store the resulting per-point 3D trajectories $\mathbf{x}_{n,t}\in\mathbb{R}^3$ in camera space, where $n$ indexes tracks and $t$ indexes frames.

Internet videos contain cuts, extreme motion blur, and tracker failures. We discard unreliable and short sequences and keep only high-definition pairs, retaining approximately 0.4M clips after filtering. We normalize camera intrinsics to a canonical resolution and normalize depth to $[0,1]$ for stable training. Each training example therefore comprises RGB frames, a text prompt, 3D tracks, and camera parameters. We refer to this collection as \dataset; it provides structured conditioning targets for training and can also supply control signals at inference time. Full details of the data pipeline and teacher training configuration are given in \cref{sec:supp:teacher}.

\begin{wrapfigure}{r}{0.5\linewidth}
    \centering
    \vspace{-12pt}
    \includegraphics[width=\linewidth]{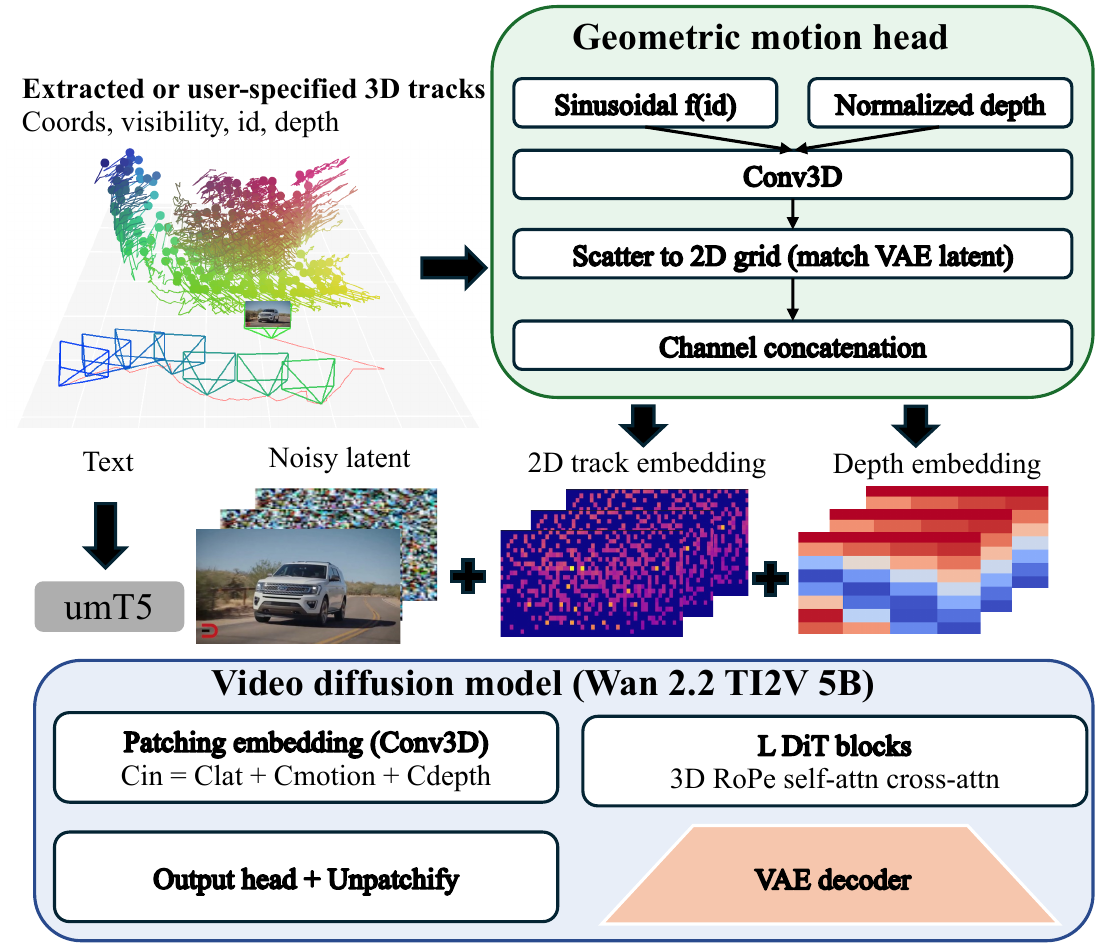}
    \caption{\textbf{Overview of \modelname.} Geometric 3D tracks and text are encoded into 2D and depth motion features, fused with noisy latents in a DiT with 3D RoPE, then decoded to video for precise trajectory, camera, and motion control.}
    \label{fig:model}
    \vspace{-10pt}
\end{wrapfigure}

\subsection{3D Motion-Conditioned Video Model}\label{sec:model}

\paragraph{Overview.}
Our generator is based on a TI2V diffusion transformer (Wan2.2 TI2V-5B) \citep{wan2025wan}; \cref{fig:model} provides an overview. Inspired by MotionStream \citep{motionstream2025}, we inject motion by building a compact motion feature tensor on the VAE latent grid and fusing it with noisy latents \emph{before} DiT patch embedding via \textbf{channel concatenation}. For our 4D control task, we extend this design in two ways: (i) 3D tracks are projected to normalized image coordinates for rasterization; and (ii) scalar depth is encoded in a \textbf{separate branch} rather than fused into per-track identity features prior to scattering, so geometric cues stay disentangled from track identity.

\paragraph{Notation.}
For each batch sample $b\in\{1,\ldots,B\}$, we are given $N$ point tracks in the camera space (\cref{sec:data}). From each camera-space point $\mathbf{x}_{n,t}=(x,y,z)^\top$, we obtain normalized image coordinates and depth via
\begin{equation}
    \mathbf{p}_{n,t}=\pi(\mathbf{K}_t,\mathbf{x}_{n,t})\in[0,1]^2,\qquad d_{n,t}=z,
\end{equation}
where $\pi$ denotes perspective projection followed by resolution normalization. Track $n$ has an integer id $a_n$ and, at each frame $t\in\{1,\ldots,T\}$, provides image coordinates $(p^x_{n,t},p^y_{n,t})^\top$ and an optional scalar depth $d_{n,t}$. Let $(H_\ell,W_\ell)$ denote the latent spatial resolution aligned with the VAE tokenizer grid, and let $F$ denote the number of latent frames fed to the DiT. The motion encoder operates on $T$ raw frames and temporally compresses features to length $F$: the first frame maps to a dedicated latent frame while every subsequent group of four frames is compressed into one, giving $F=(T-1)/4+1$.

\paragraph{Per-track sinusoidal embeddings.}
Each track id $a_n$ is mapped to a fixed sinusoidal positional encoding $\boldsymbol{\phi}_n=\mathrm{PE}(a_n)\in\mathbb{R}^{D}$, using the standard 1D construction (even $D$, log-spaced frequencies) rather than a learned embedding table. The resulting $\boldsymbol{\phi}_n$ is shared across all time steps and depends only on track identity.

\paragraph{Rasterization on the latent grid.}
We map normalized coordinates to discrete latent indices by nearest-neighbor quantization,
\begin{equation}
    x_{n,t}=\Big\lfloor p^x_{n,t}\cdot \max(W_\ell-1,\,1)\Big\rfloor,\qquad
    y_{n,t}=\Big\lfloor p^y_{n,t}\cdot \max(H_\ell-1,\,1)\Big\rfloor,
\end{equation}
and clamp indices to $[0,W_\ell-1]$ and $[0,H_\ell-1]$. We initialize a track raster $\mathbf{C}\in\mathbb{R}^{B\times D\times T\times H_\ell\times W_\ell}$ to zero and scatter-add features at occupied cells:
\begin{equation}
    \mathbf{C}_{b,d,t,y,x}\;=\;
    \sum_{n=1}^{N}\boldsymbol{\phi}_{n,d}\,
    \mathbb{I}\!\left[x=x_{n,t},\,y=y_{n,t}\right],
    \label{eq:track-scatter}
\end{equation}
where $\mathbb{I}[\cdot]$ is the indicator function and multiple tracks landing in the same cell $(t,y,x)$ accumulate by summation. We do not feed an explicit visibility mask into the track branch; occlusion and missing samples are handled implicitly by the depth branch and by empty raster cells.

\paragraph{Track branch.}
The raster $\mathbf{C}$ is processed by a lightweight shared Conv3D head,
\begin{equation}
    \mathbf{F}^{\mathrm{trk}}
    = \mathcal{H}_{\mathrm{trk}}(\mathbf{C})
    = \mathrm{Conv3D}_{1\times1\times1}\!\Big(
        \mathrm{SiLU}\!\big(
        \mathrm{Conv3D}_{4\times1\times1}(\mathbf{C})
        \big)
    \Big),
\end{equation}
where $\mathrm{Conv3D}_{4\times1\times1}$ uses stride $(4,1,1)$ for $\times4$ temporal compression along the frame axis and the final $1\times1\times1$ layer maps to $C_{\mathrm{trk}}$ channels.

\paragraph{Depth branch.}
When enabled, each space--time sample may provide depth $d_{n,t}$. We map depth to disparity $\tilde{d}_{n,t}=1/(d_{n,t}+\varepsilon)$ for numerical stability, embed it via a linear map $\mathbf{e}^{\mathrm{dep}}_{n,t}=\mathbf{W}_d\,\tilde{d}_{n,t}+\mathbf{b}_d\in\mathbb{R}^{D_d}$, and rasterize onto the same indices $(t,y_{n,t},x_{n,t})$ as in \cref{eq:track-scatter} to obtain $\mathbf{C}^{\mathrm{dep}}\in\mathbb{R}^{B\times D_d\times T\times H_\ell\times W_\ell}$. A separate Conv3D head $\mathcal{H}_{\mathrm{dep}}$ with the same $(4,1,1)$ temporal structure produces $\mathbf{F}^{\mathrm{dep}}\in\mathbb{R}^{B\times C_{\mathrm{dep}}\times F\times H_\ell\times W_\ell}$. When depth is unavailable at runtime, we set $\mathbf{F}^{\mathrm{dep}}=\mathbf{0}$, keeping the DiT input channel width fixed.

\paragraph{Fusion with noisy latents.}
Let $\mathbf{z}\in\mathbb{R}^{B\times C_{\mathrm{lat}}\times F\times H_\ell\times W_\ell}$ denote noisy VAE latents. We form the motion conditioning tensor $\mathbf{F}^{\mathrm{cond}}=\mathbf{F}^{\mathrm{trk}}\;\Vert\;\mathbf{F}^{\mathrm{dep}}$ by channel concatenation, then concatenate it with $\mathbf{z}$ to obtain $\tilde{\mathbf{z}}=\mathbf{z}\;\Vert\;\mathbf{F}^{\mathrm{cond}}\in\mathbb{R}^{B\times(C_{\mathrm{lat}}+C_{\mathrm{trk}}+C_{\mathrm{dep}})\times F\times H_\ell\times W_\ell}$, which is fed into the DiT backbone.

\paragraph{Training objective.}
Let $\mathbf{x}_0\in\mathbb{R}^{B\times C_{\mathrm{lat}}\times F\times H_\ell\times W_\ell}$ denote clean VAE latents of a training clip. We fine-tune the pretrained TI2V diffusion transformer with the same flow-matching supervision used by Wan \citep{wan2025wan}. For each step we sample a flow time $\tau$ (equivalently a noise level $\sigma_\tau$), draw Gaussian noise $\boldsymbol{\epsilon}\sim\mathcal{N}(\mathbf{0},\mathbf{I})$, and form the interpolated latent $\mathbf{x}_\tau=(1-\sigma_\tau)\mathbf{x}_0+\sigma_\tau\boldsymbol{\epsilon}$. The network predicts a velocity field $\mathbf{v}_\theta(\mathbf{x}_\tau,\tau,\mathbf{c}_{\text{text}},\mathbf{c}_{\text{img}},\mathbf{F}^{\mathrm{cond}})$, where $\mathbf{c}_{\text{text}}$ is the text embedding, $\mathbf{c}_{\text{img}}$ encodes the first frame for TI2V, and $\mathbf{F}^{\mathrm{cond}}$ is the motion tensor built from 3D tracks (and depth when available). Training minimizes a weighted mean-squared error to the flow target $\mathbf{v}=\boldsymbol{\epsilon}-\mathbf{x}_0$,
\begin{equation}
    \mathcal{L}_{\mathrm{FM}}
    = \mathbb{E}_{\tau,\boldsymbol{\epsilon}}\!\left[
        w(\tau)\,\big\|
        \mathbf{v}_\theta(\mathbf{x}_\tau,\tau,\mathbf{c}_{\text{text}},\mathbf{c}_{\text{img}},\mathbf{F}^{\mathrm{cond}})
        - \mathbf{v}
        \big\|_2^2
    \right],
\end{equation}
with $w(\tau)$ following the scheduler's timestep weighting. For TI2V, the first-frame latent is held fixed when constructing $\mathbf{x}_\tau$, and the loss is evaluated only on future latent frames.

\paragraph{Parameter-efficient fine-tuning.}
To preserve the pretrained generative prior, we adopt LoRA \citep{hu2022lora} on the DiT while keeping its base weights frozen. Low-rank adapters are inserted into the attention and feed-forward projections (i.e., $\mathbf{W}_q,\mathbf{W}_k,\mathbf{W}_v,\mathbf{W}_o$ and FFN layers). In parallel, the motion pathway, comprising the track and depth rasterizers and Conv3D heads ($\mathcal{H}_{\mathrm{trk}}$, $\mathcal{H}_{\mathrm{dep}}$), the depth projector $\mathbf{W}_d$, and the widened patch-embedding input, is trained end-to-end.

\paragraph{Two-stage training schedule.}
To stabilize optimization, we use a two-stage curriculum:
\begin{itemize}[nosep,leftmargin=*]
  \item \textbf{Stage A (low resolution, short horizon).} Train on shorter clips at reduced spatial resolution ($256$p, $49$ frames) to learn coarse alignment between 3D motion cues and video dynamics at lower compute.
  \item \textbf{Stage B (high resolution, long horizon).} Continue at higher resolution and longer duration ($480$p, $81$ frames) for finer motion--appearance alignment and improved long-horizon temporal consistency.
\end{itemize}
This schedule follows common practice in large video-model fine-tuning: first learn the conditioning interface, then scale resolution and length. Our ablations confirm the effectiveness of this two-stage schedule.

\subsection{Streaming Distillation for Interactive 4D Control}

The bidirectional teacher (\cref{sec:model}) produces high-quality 3D-aligned video but requires around 50 denoising steps over the full clip, far too slow for interaction. We distill it into a causal 4-step student following the self-forcing paradigm \citep{huang2025selfforcing,motionstream2025}. The student reuses the teacher's motion encoder and channel-fusion pathway but replaces bidirectional self-attention with \textbf{block-wise causal attention} \citep{chen2024diffusion}: latents are partitioned into chunks $\{\mathbf{z}^i\}_{i=1}^{L}$, and each chunk attends only to its own tokens, an attention sink anchored on the first frame (see \cref{sec:attention_sink} for its effect), and a local window of preceding chunks, with a rolling KV cache ensuring constant compute and memory. A key enabler is that our motion encoder is \textbf{temporally separable by construction}: the per-frame scatter-add rasterization (\cref{eq:track-scatter}) and the non-overlapping $(4,1,1)$ Conv3D head make segment-by-segment encoding bit-identical to full-sequence encoding, requiring no architectural change for streaming.

Following CausVid \citep{yin2025causvid}, we first initialize the student on 4\,k noise--clean ODE pairs from the teacher to stabilize few-step denoising, then train with the DMD objective \citep{yin2024dmd} under temporal self-rollout: each chunk is denoised conditioned on the student's \emph{own} prior outputs via KV cache, mirroring inference. Let $\mathbf{x}$ be the generator's clean rollout output and $\mathbf{x}_\tau{=}(1{-}\sigma_\tau)\mathbf{x}{+}\sigma_\tau\boldsymbol{\epsilon}$ its noised version. A frozen real score and a learnable fake score are jointly optimized,
\begin{equation}
    \mathcal{L}_{\mathrm{critic}}
    = \bigl\|\hat{\mathbf{x}}_{0}^{\mathrm{fake}}(\mathbf{x}_\tau,\tau,\mathbf{c})
    - \mathbf{x}\bigr\|^2, \qquad
    \nabla_{\!\mathbf{x}}\,\mathcal{L}_{\mathrm{gen}}
    = \hat{\mathbf{x}}_{0}^{\mathrm{fake}}(\mathbf{x}_\tau,\tau,\mathbf{c})
    - \hat{\mathbf{x}}_{0}^{\mathrm{real}}(\mathbf{x}_\tau,\tau,\mathbf{c}),
\end{equation}
driving the student's output distribution toward the teacher's guided distribution. Because both scores share the same 3D motion tensor interface, geometric control transfers without additional adapters. At inference, generation proceeds chunk-by-chunk: after each chunk is decoded, the motion encoder recomputes $\mathbf{F}^{\mathrm{cond}}$ from the latest 3D tracks and depth, letting users revise control at any time at constant memory cost. More distillation details are provided in \cref{sec:supp:distill}.

\begin{figure*}[t!]
    \centering
    \setlength{\tabcolsep}{0pt}
    \renewcommand{\arraystretch}{0.8}
    \begin{tabular}{ccccc}
        \multicolumn{5}{c}{\small\textbf{Object control}} \\
        \includegraphics[width=0.2\linewidth]{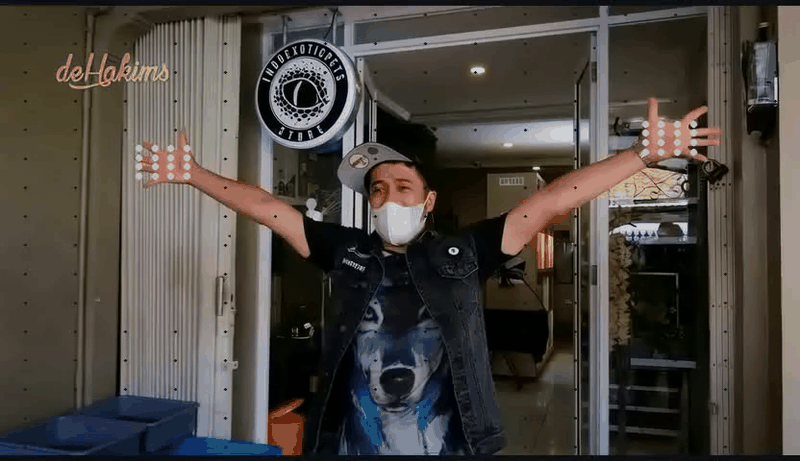} &
        \includegraphics[width=0.2\linewidth]{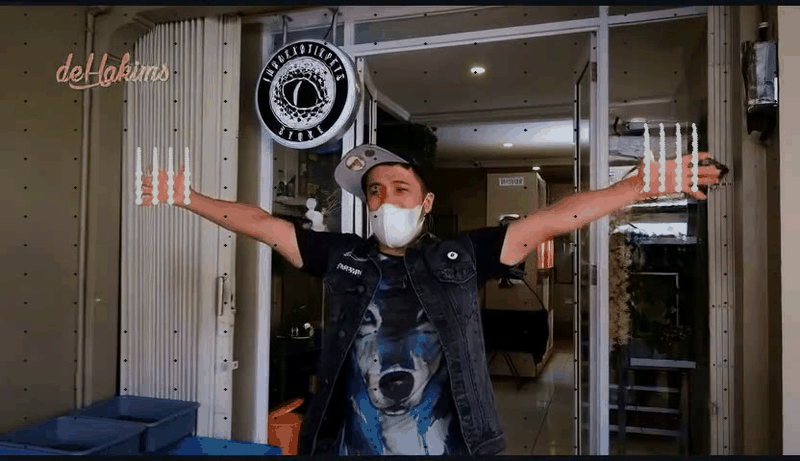} &
        \includegraphics[width=0.2\linewidth]{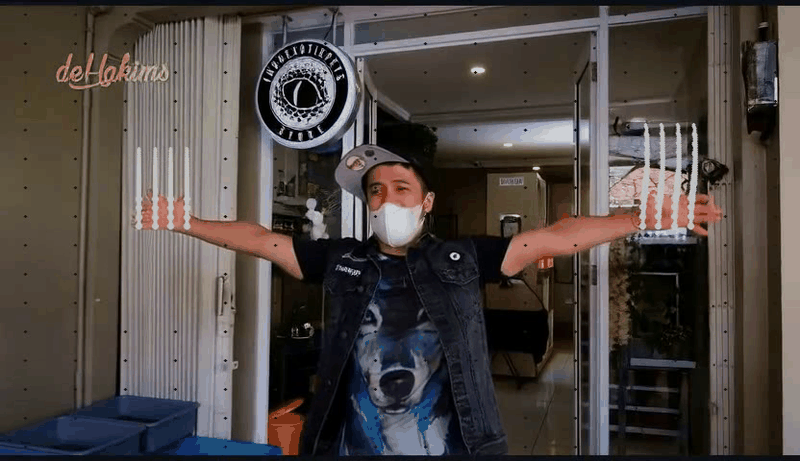} &
        \includegraphics[width=0.2\linewidth]{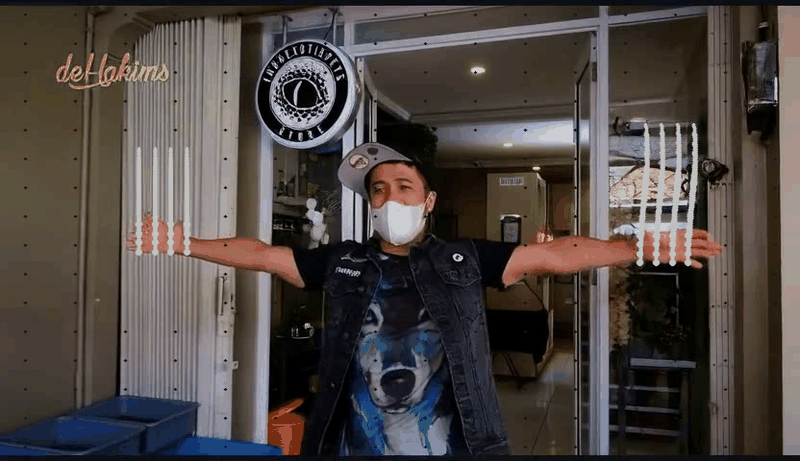} &
        \includegraphics[width=0.2\linewidth]{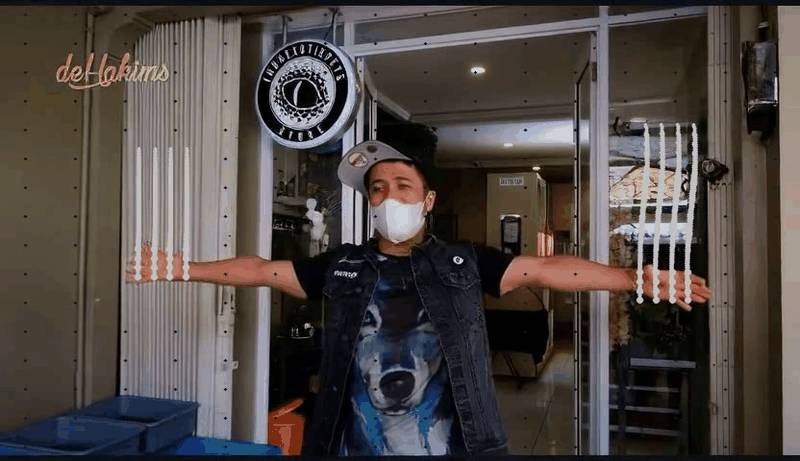} \\
        \includegraphics[width=0.2\linewidth]{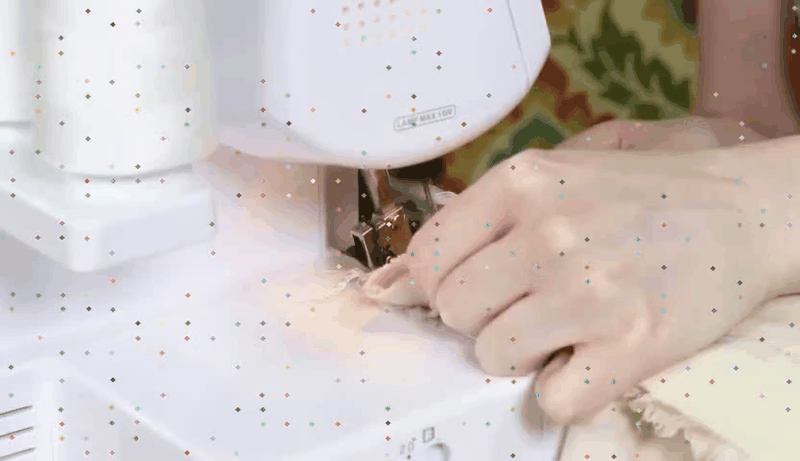} &
        \includegraphics[width=0.2\linewidth]{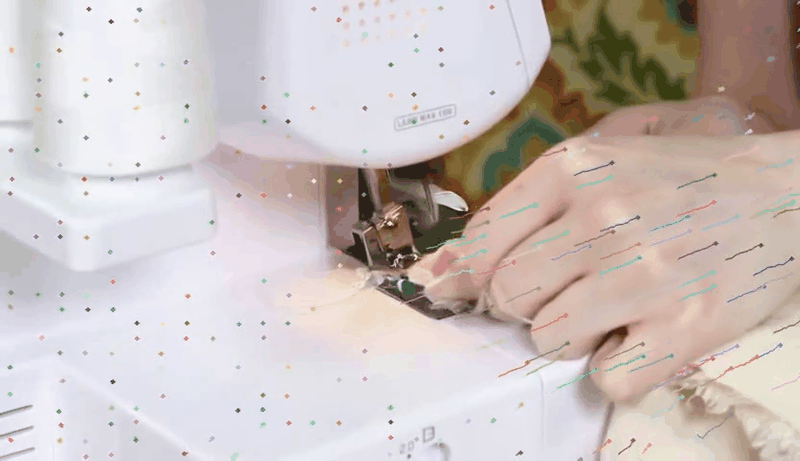} &
        \includegraphics[width=0.2\linewidth]{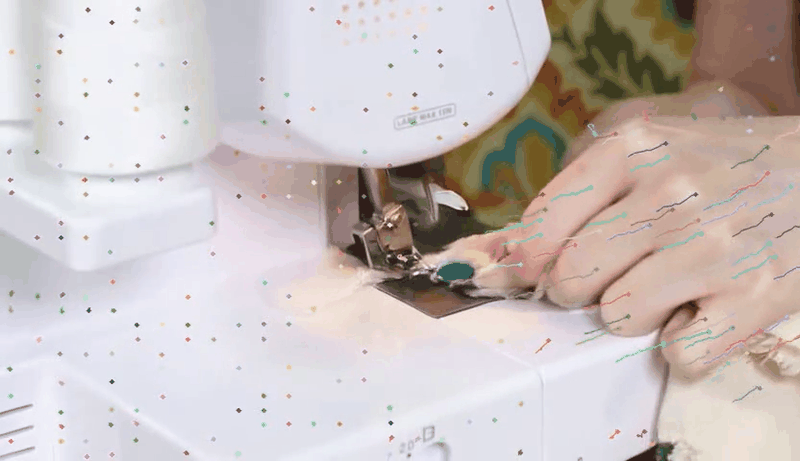} &
        \includegraphics[width=0.2\linewidth]{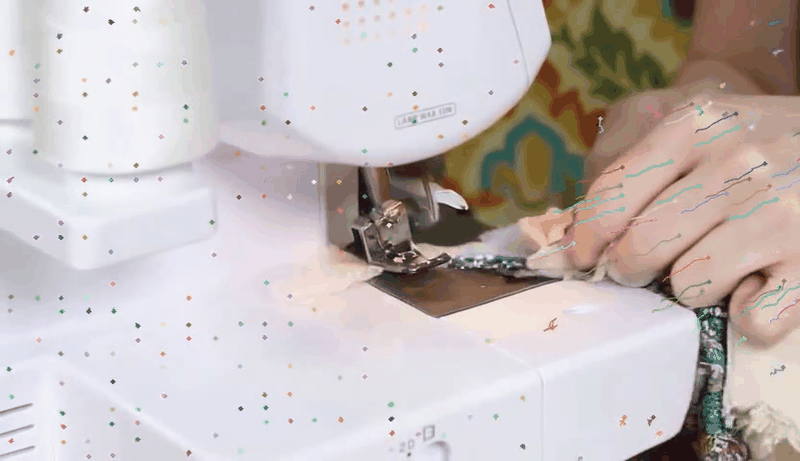} &
        \includegraphics[width=0.2\linewidth]{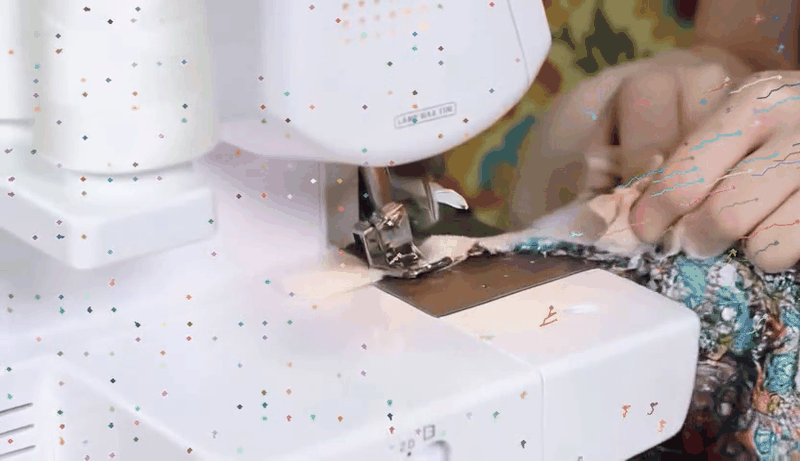} \\
        \includegraphics[width=0.2\linewidth]{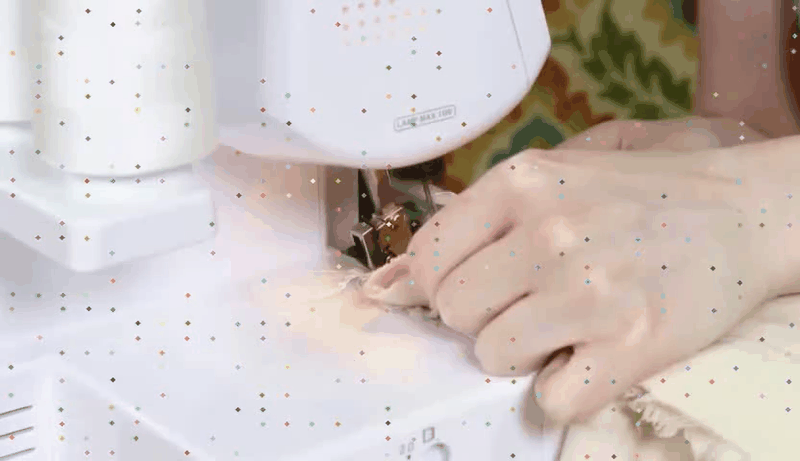} &
        \includegraphics[width=0.2\linewidth]{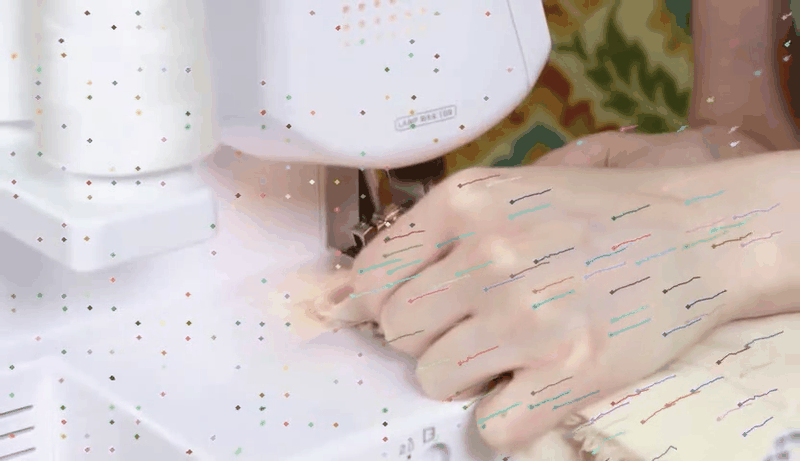} &
        \includegraphics[width=0.2\linewidth]{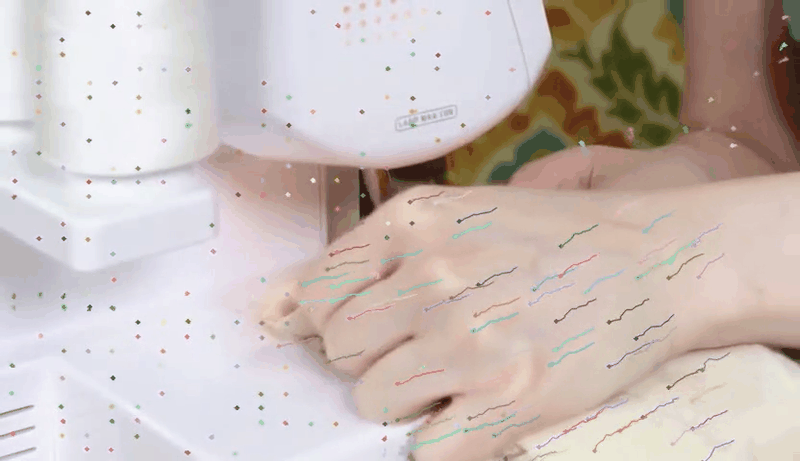} &
        \includegraphics[width=0.2\linewidth]{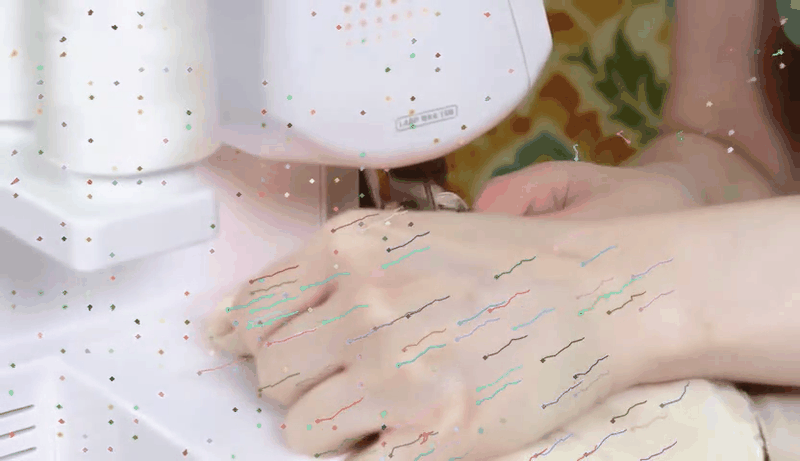} &
        \includegraphics[width=0.2\linewidth]{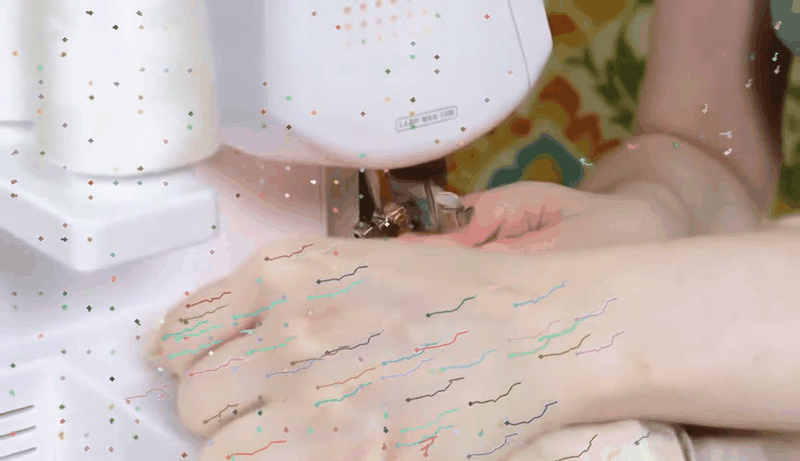} \\
        \multicolumn{5}{c}{\small\textbf{Camera control: pan and zoom}} \\
        \includegraphics[width=0.2\linewidth]{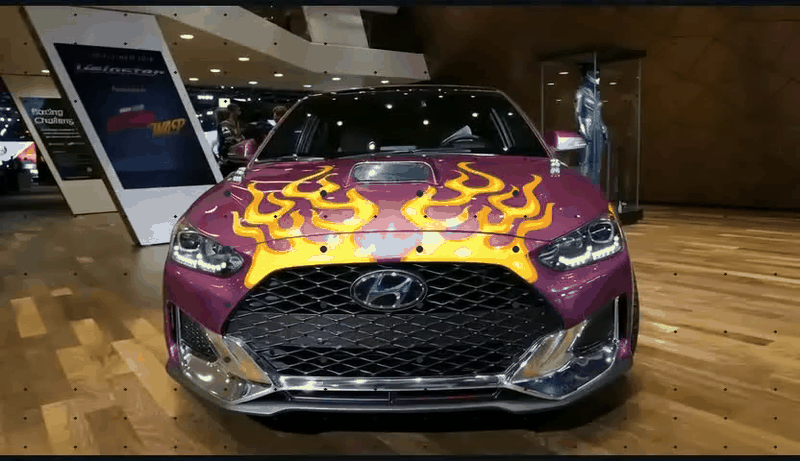} &
        \includegraphics[width=0.2\linewidth]{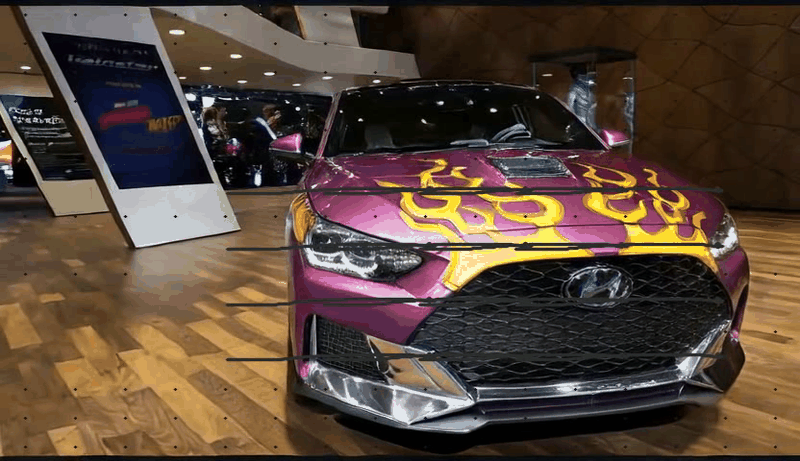} &
        \includegraphics[width=0.2\linewidth]{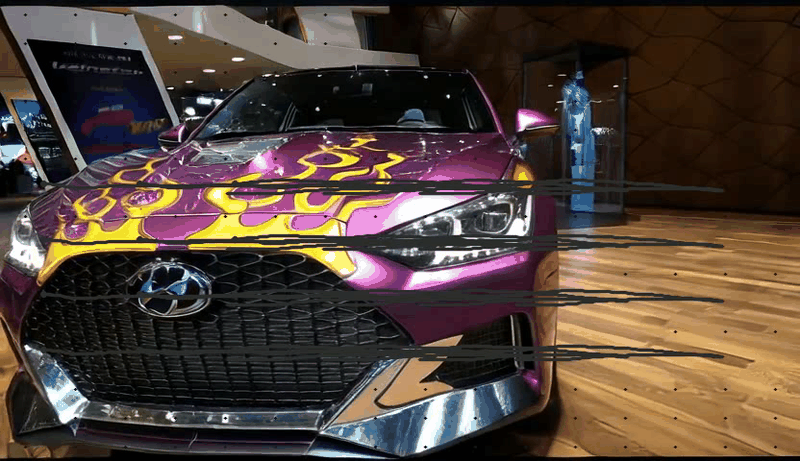} &
        \includegraphics[width=0.2\linewidth]{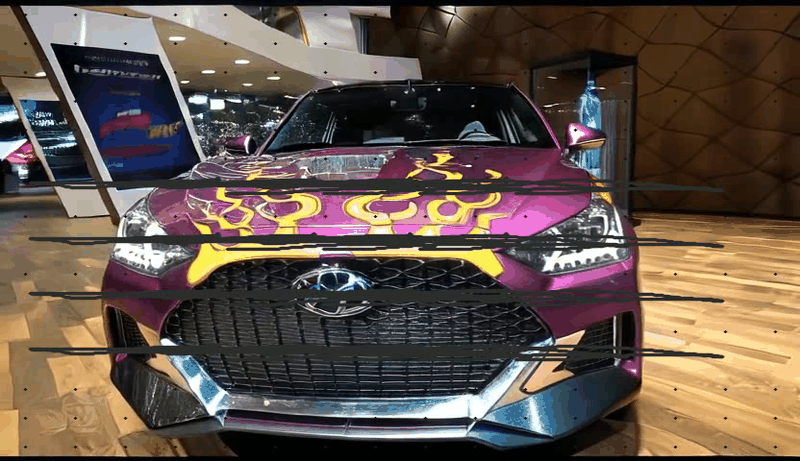} &
        \includegraphics[width=0.2\linewidth]{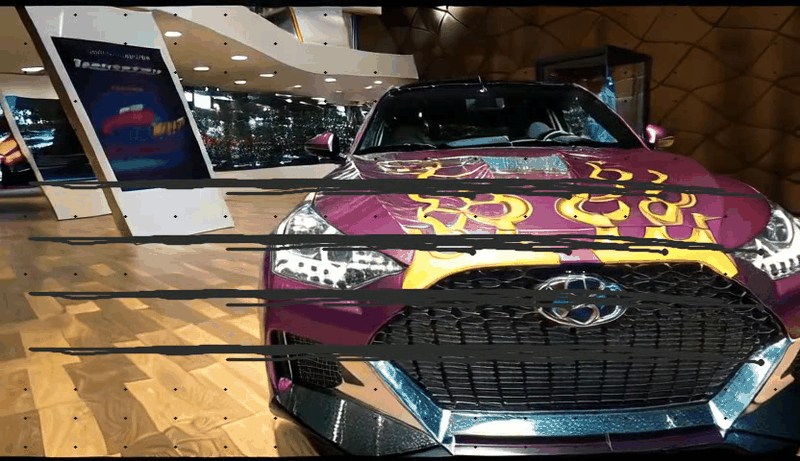} \\
        \includegraphics[width=0.2\linewidth]{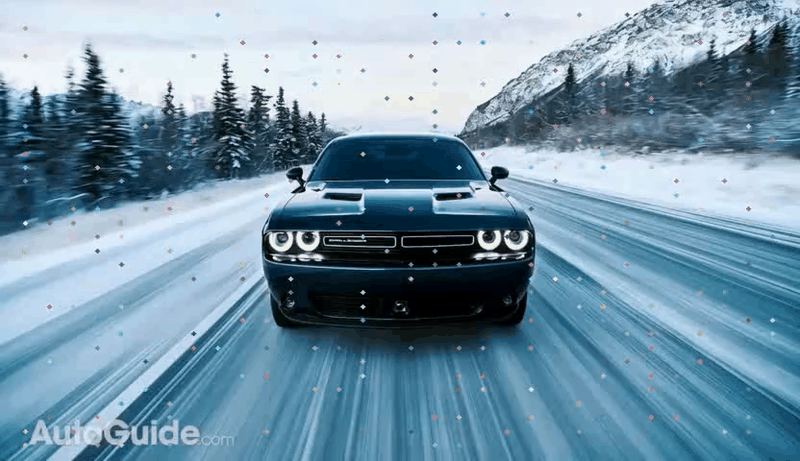} &
        \includegraphics[width=0.2\linewidth]{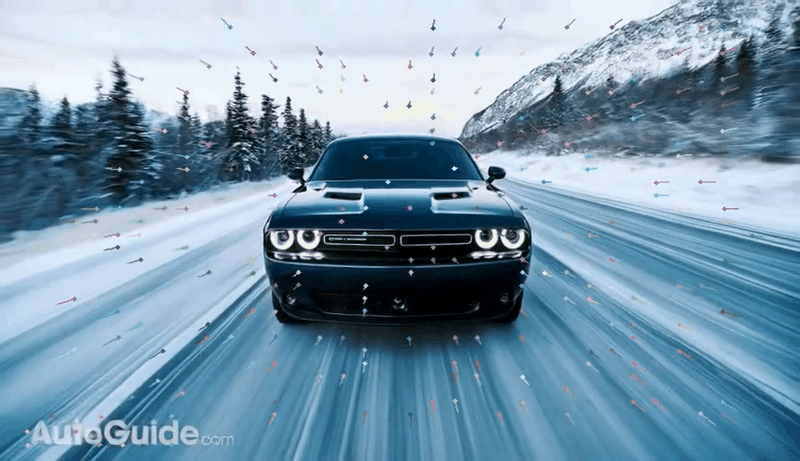} &
        \includegraphics[width=0.2\linewidth]{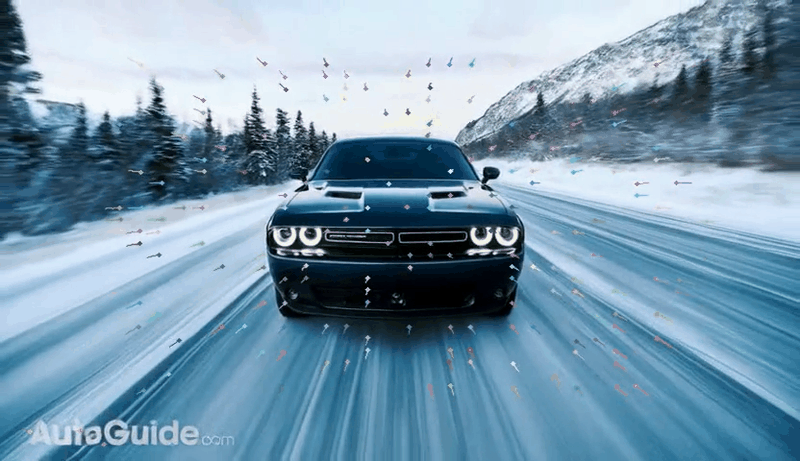} &
        \includegraphics[width=0.2\linewidth]{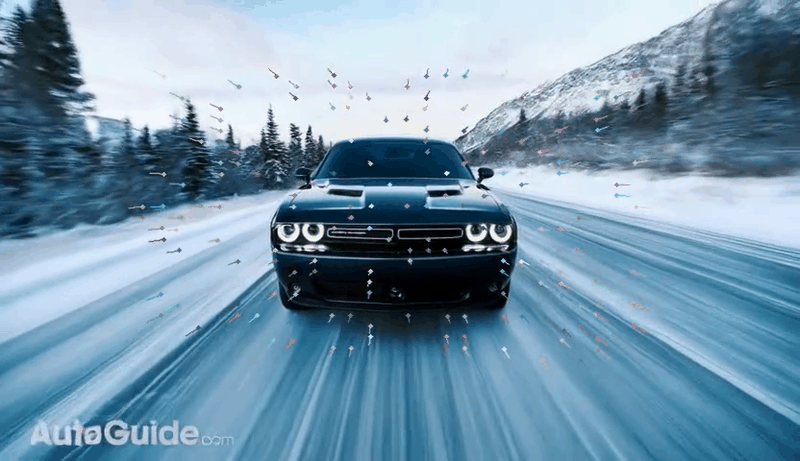} &
        \includegraphics[width=0.2\linewidth]{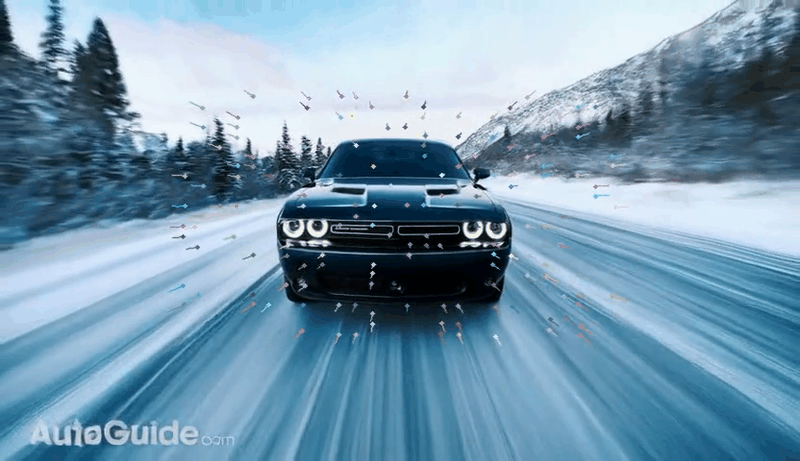} \\
        \multicolumn{5}{c}{\small\textbf{Object and camera joint control}} \\
        \includegraphics[width=0.2\linewidth]{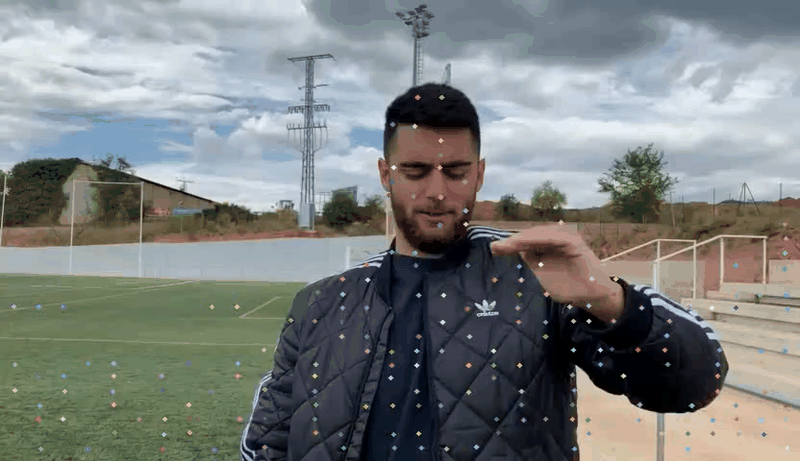} &
        \includegraphics[width=0.2\linewidth]{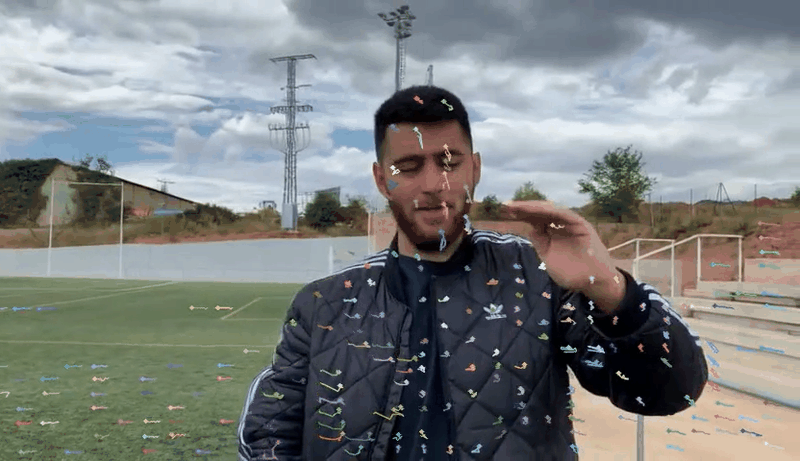} &
        \includegraphics[width=0.2\linewidth]{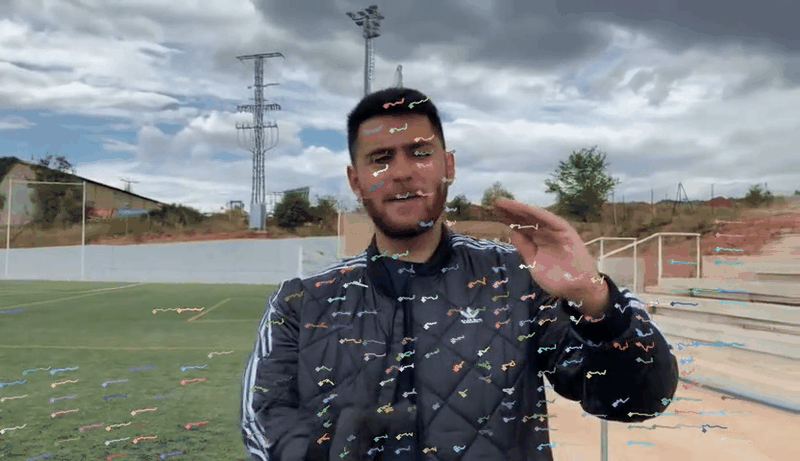} &
        \includegraphics[width=0.2\linewidth]{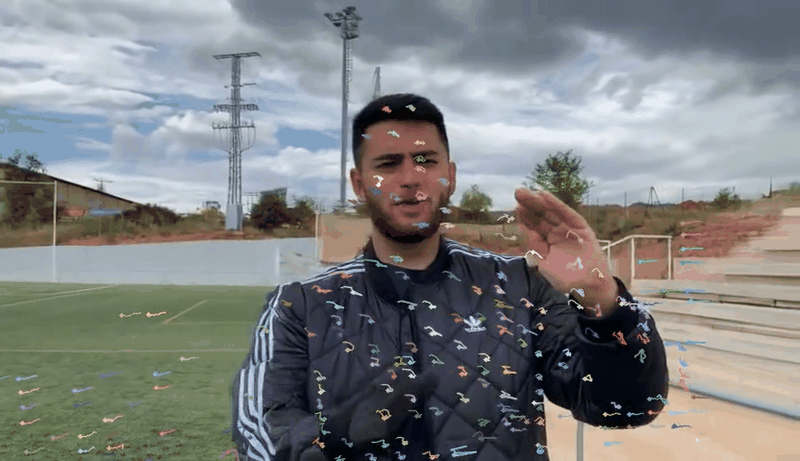} &
        \includegraphics[width=0.2\linewidth]{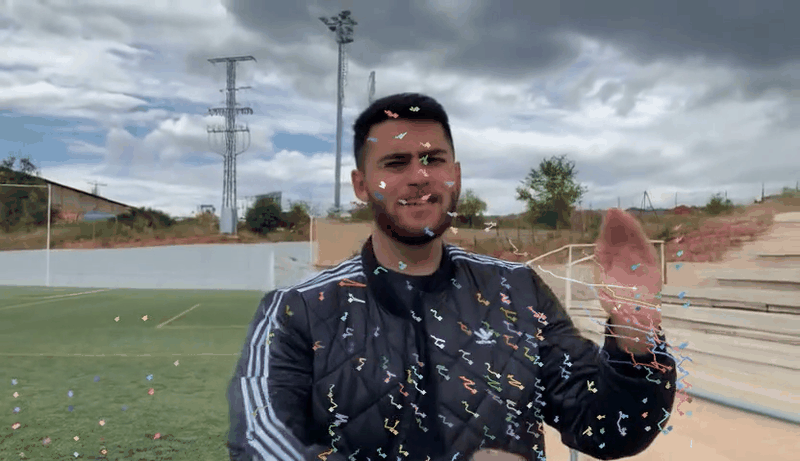} \\
        \multicolumn{5}{c}{\small\textbf{Depth control}} \\
        \includegraphics[width=0.2\linewidth]{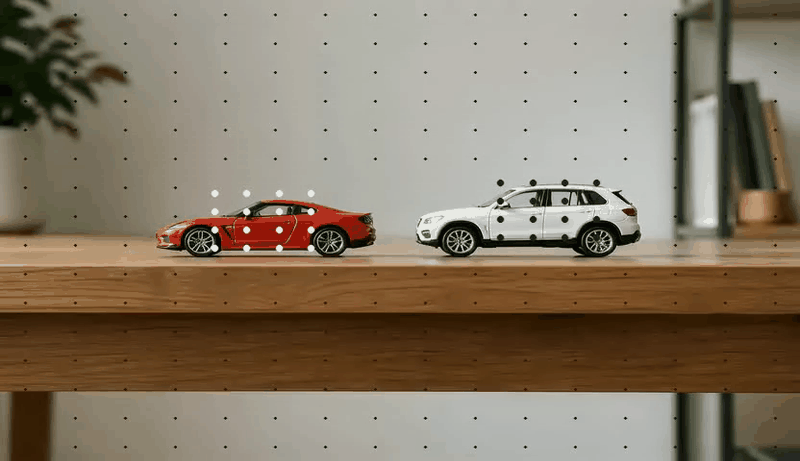} &
        \includegraphics[width=0.2\linewidth]{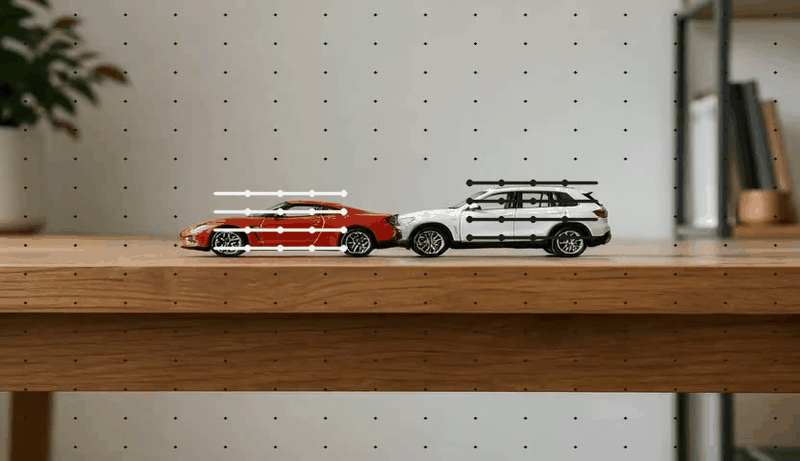} &
        \includegraphics[width=0.2\linewidth]{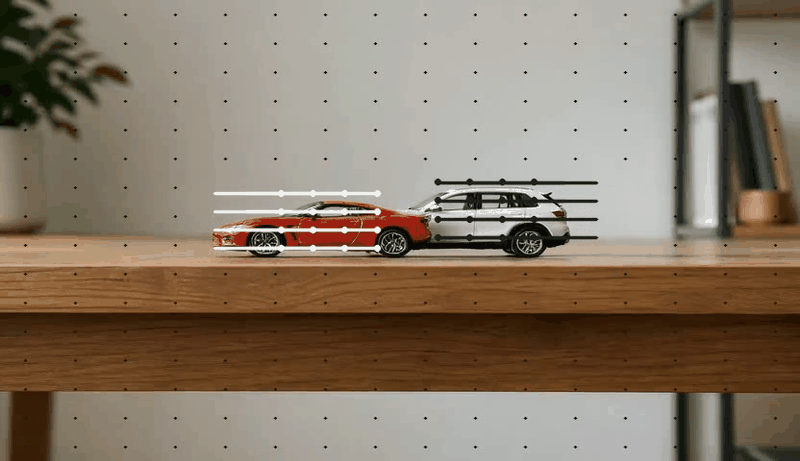} &
        \includegraphics[width=0.2\linewidth]{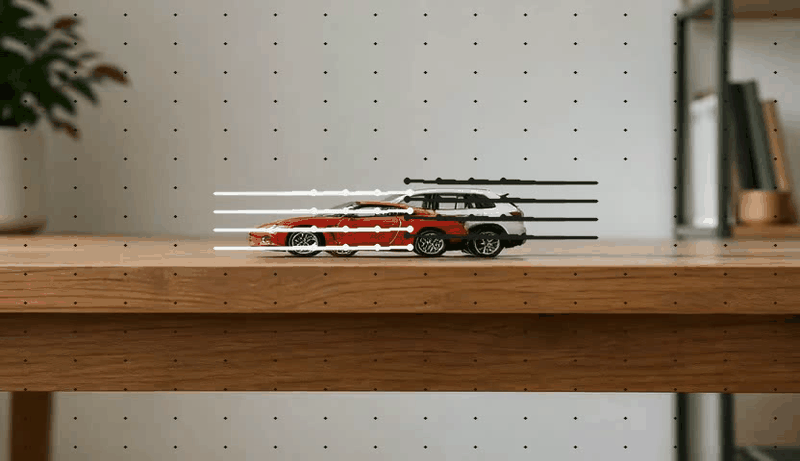} &
        \includegraphics[width=0.2\linewidth]{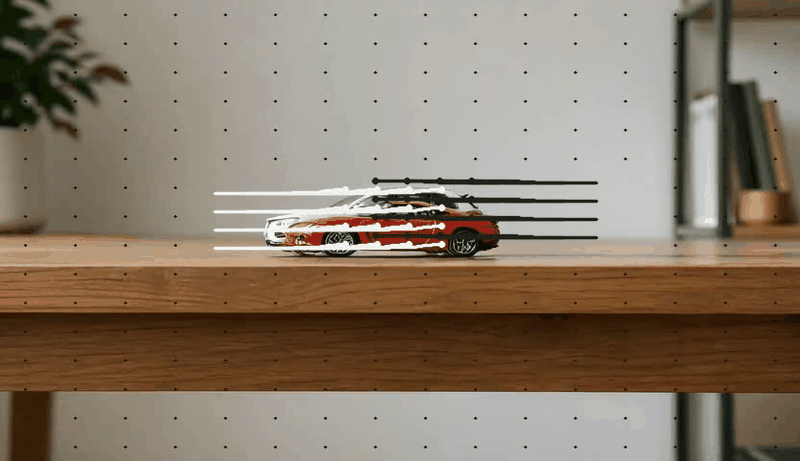} \\
        \includegraphics[width=0.2\linewidth]{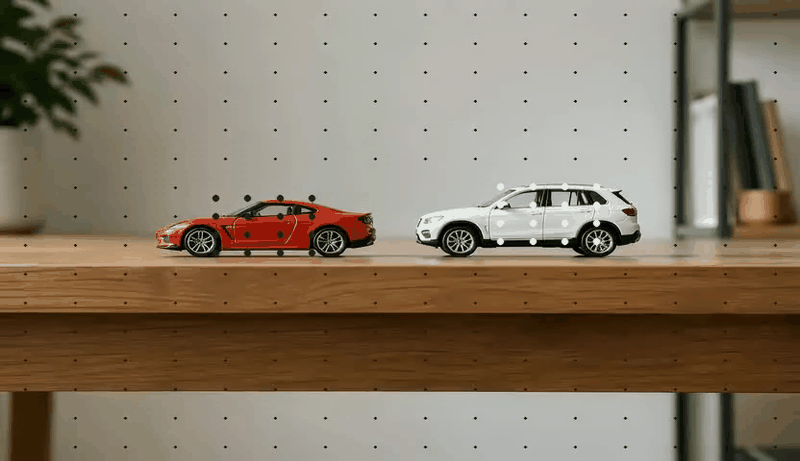} &
        \includegraphics[width=0.2\linewidth]{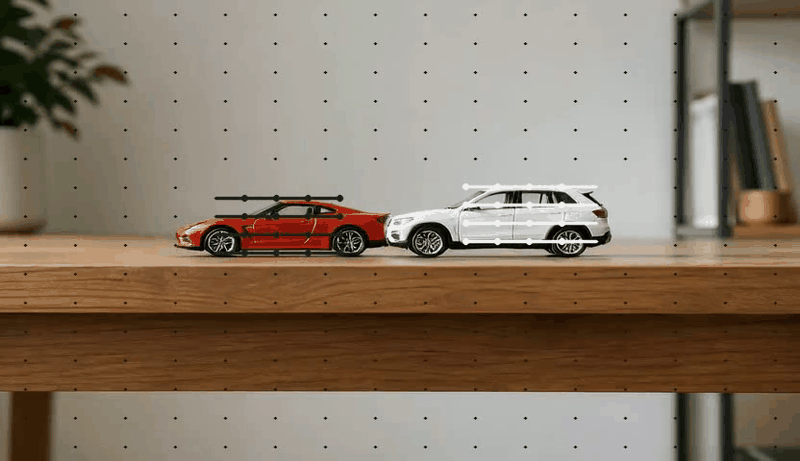} &
        \includegraphics[width=0.2\linewidth]{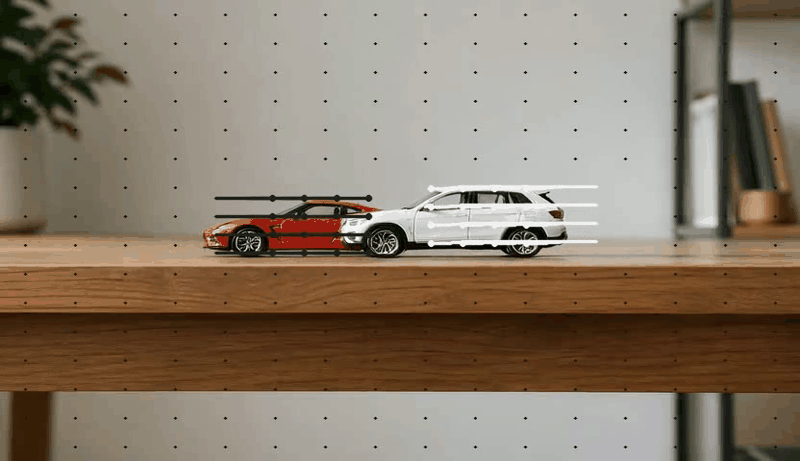} &
        \includegraphics[width=0.2\linewidth]{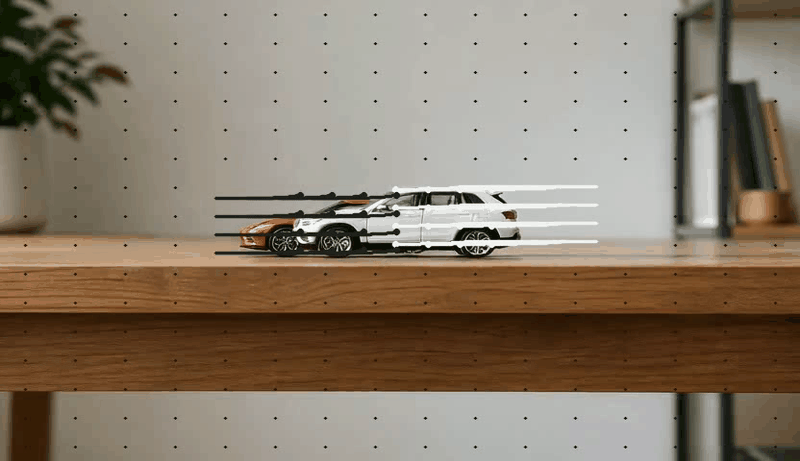} &
        \includegraphics[width=0.2\linewidth]{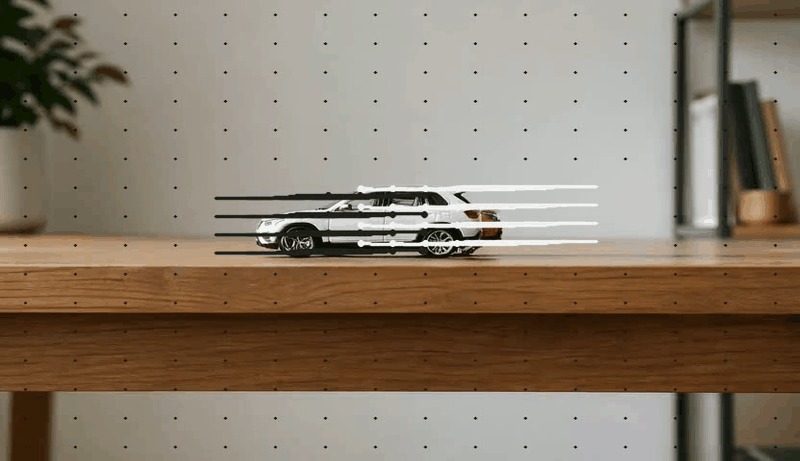} \\
        \multicolumn{5}{c}{\small\textbf{Motion transfer}} \\
        \includegraphics[width=0.2\linewidth]{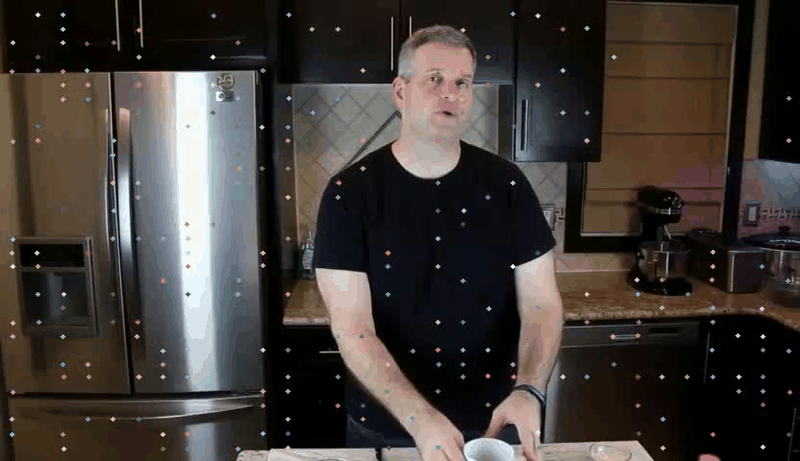} &
        \includegraphics[width=0.2\linewidth]{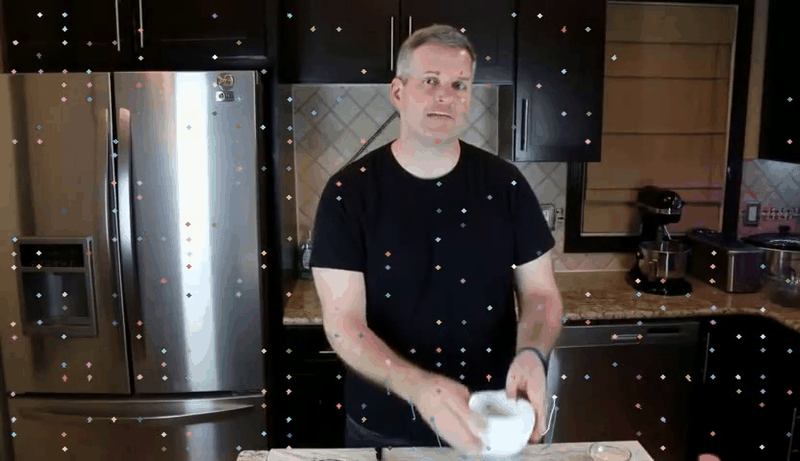} &
        \includegraphics[width=0.2\linewidth]{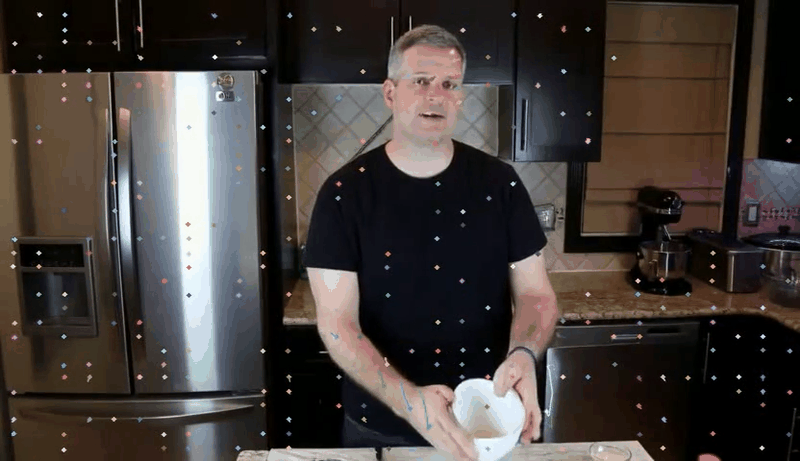} &
        \includegraphics[width=0.2\linewidth]{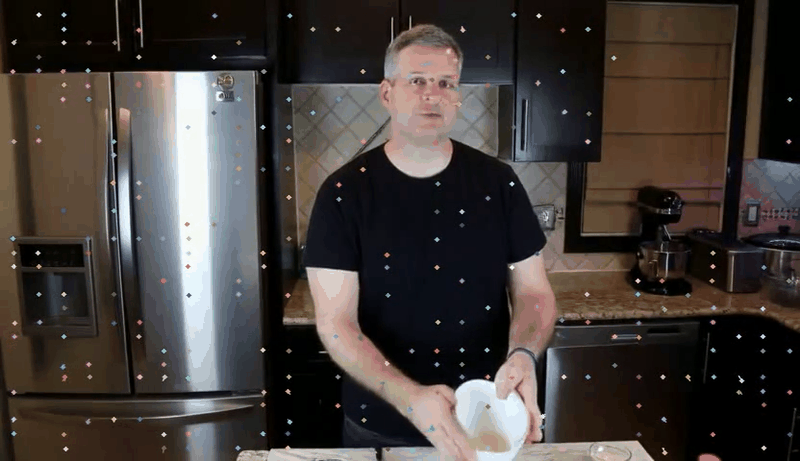} &
        \includegraphics[width=0.2\linewidth]{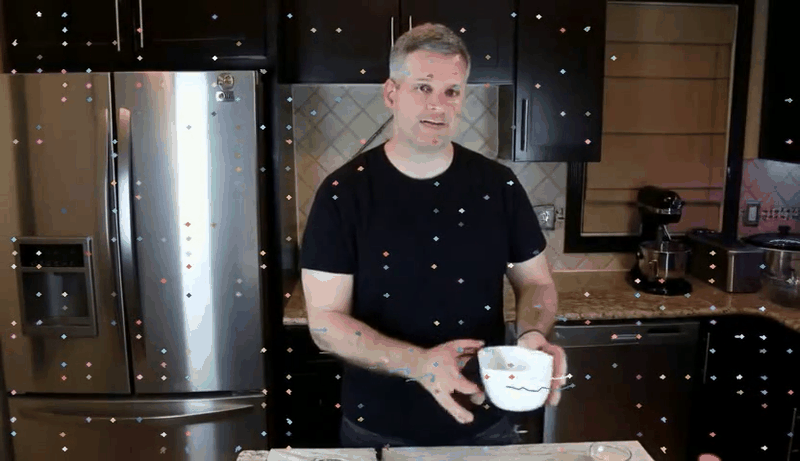} \\
        \includegraphics[width=0.2\linewidth]{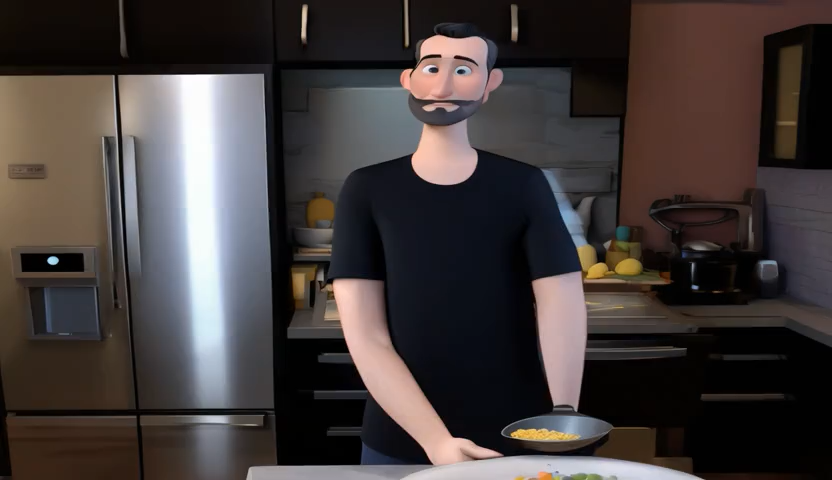} &
        \includegraphics[width=0.2\linewidth]{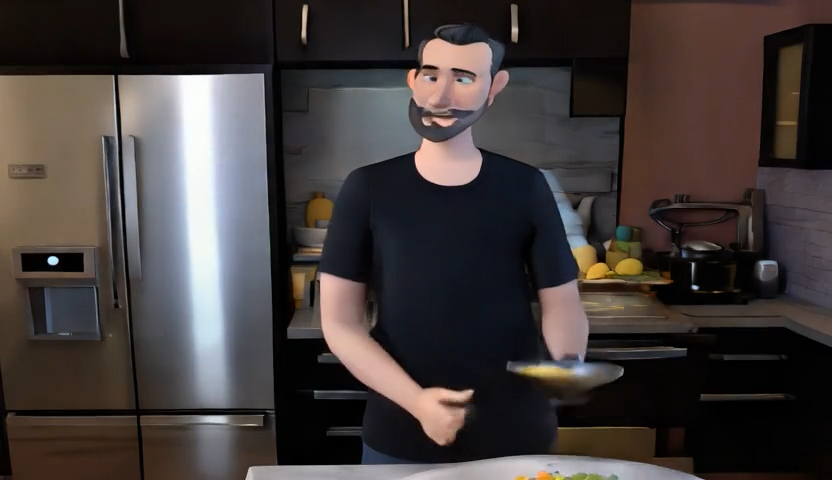} &
        \includegraphics[width=0.2\linewidth]{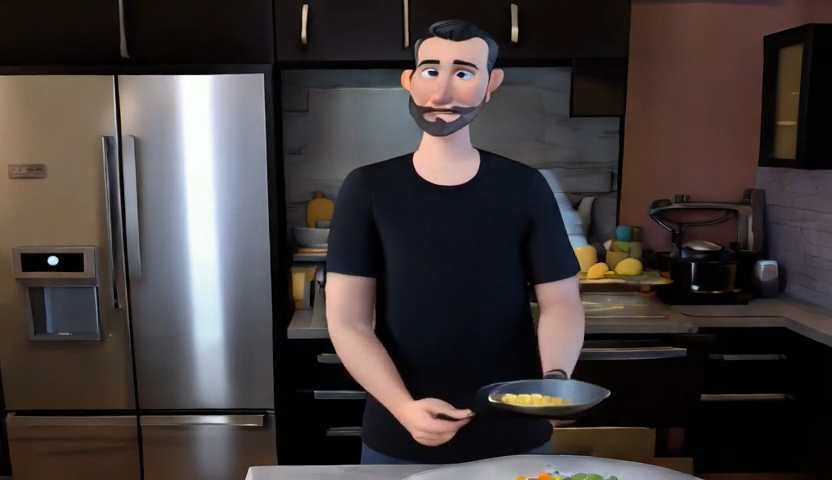} &
        \includegraphics[width=0.2\linewidth]{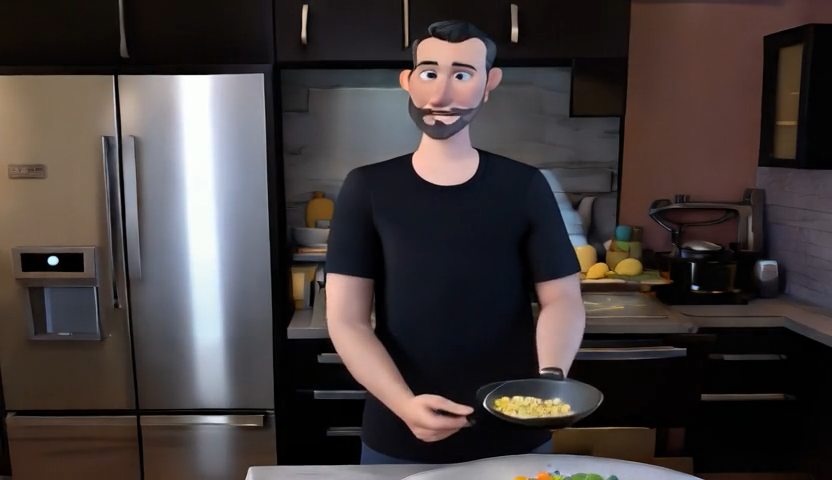} &
        \includegraphics[width=0.2\linewidth]{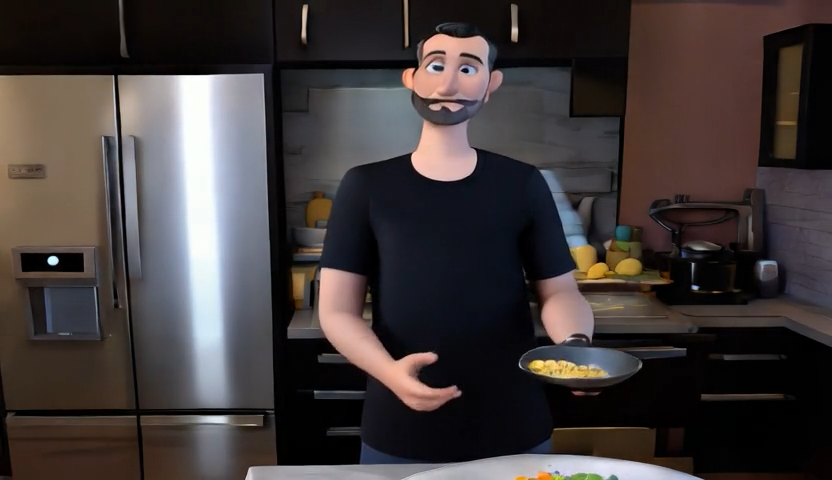} \\
    \end{tabular}
    \caption{\textbf{Qualitative results of \modelname across diverse control modes.} Each group demonstrates one control capability: \textbf{Object control} steers the trajectory of individual objects; \textbf{Camera control} adjusts viewpoint and camera movement; \textbf{Object and camera joint control} coordinates object dynamics with camera motion; \textbf{Depth control} conditions generation on per-point depth to resolve occlusion and depth ordering; and \textbf{Motion transfer} carries motion from a reference video onto new content. Frames run in temporal order from left to right. Point trajectories are overlaid for visualization; in the depth control group, white points mark trajectories at shallower depths and black points those at greater depths.}
    \label{fig:qualitative_results}
\end{figure*}

\section{Experiments}\label{sec:exp}

We evaluate the 3D track-conditioned teacher and its streaming distilled student on three regimes that reflect the paper's core claims: (i) joint object and camera control, (ii) motion transfer, and (iii) interactive streaming control. \cref{fig:qualitative_results} gives an overview of the qualitative results.

\subsection{Joint Object and Camera Control}\label{sec:exp_joint}

\paragraph{Setup.}
Given text, a first frame, and motion conditioning, a model should synthesize video whose object motion and camera motion follow the conditioning while remaining visually plausible. We evaluate on 30 in-the-wild videos from the DAVIS validation set \citep{perazzi2016davis}, selected for challenging coupled camera and object motion. Each method receives the same first frame and text prompt where applicable, with conditioning in its native interface (\eg, 2D tracks, 2D flow, or 3D tracks); our models take text, image, and 3D tracks, and we report both the full teacher and the causal streaming variant. We compare against recent motion-conditioned generators spanning 2D-track, 2D-flow, and 3D-track conditioning: Image Conductor \citep{li2025imageconductor}, Go-With-The-Flow \citep{burgert2025gowiththeflow}, Diffusion-As-Shader \citep{gu2025das}, ATI \citep{wang2025ati}, and MotionStream \citep{motionstream2025}, of which only Diffusion-As-Shader also consumes 3D tracks. Backbones and input modalities are listed in \cref{tab:davis}.

We report reconstruction fidelity against ground-truth DAVIS frames via PSNR and SSIM (higher is better), perceptual quality via LPIPS \citep{zhang2018perceptual} (lower is better), motion faithfulness via end-point error (EPE) between predicted and target motion (lower is better), and inference throughput (FPS) at each method's default resolution and step count.

\paragraph{Results.}
Our teacher attains the best motion alignment on DAVIS, with the lowest EPE (5.29) and LPIPS (0.404) and the highest SSIM (0.479). The gain from geometry is clearest under a matched backbone: on Wan~2.2-5B, MotionStream reaches EPE 7.86 and LPIPS 0.427, so replacing 2D with 3D tracks cuts EPE by roughly 33\% and LPIPS by 5\%. Weaker image-plane interfaces fall much further behind (ATI 17.41, Go-With-The-Flow 41.99, Image Conductor 91.64), while MotionStream on the smaller Wan~2.1-1.3B backbone is our closest competitor on EPE (5.35) and SSIM (0.477) yet still trails on perceptual quality. Diffusion-As-Shader also consumes 3D tracks but runs at 0.29\,FPS and remains far behind on EPE (40.23) and LPIPS (0.483), indicating that 3D conditioning alone is insufficient without our geometric motion head and training recipe.

PSNR is the one metric on which we do not lead (16.04 against 16.61). This follows from what pixel-wise PSNR rewards, namely exact reproduction of the reference, which penalizes the plausible variation a controllable generator introduces wherever the tracks leave content underspecified, such as disoccluded surfaces and objects entering the frame. We therefore treat EPE and LPIPS as the primary indicators of control quality and report PSNR for completeness.

Our causal student reaches 20.6\,FPS, ahead of MotionStream Causal on both Wan~2.1-1.3B (16.7) and Wan~2.2-5B (10.4), while its EPE (5.48) stays within 4\% of our own teacher and below every causal baseline. Distillation therefore preserves 3D-track control rather than trading it for speed: the teacher sets the best SSIM, LPIPS, and EPE trade-off, and the student gives the strongest combination of throughput and track accuracy among Wan~2.2 systems.

\begin{table}[t!]
    \centering
    \small
    \caption{\textbf{Results on joint object and camera control.} We evaluate motion-conditioned methods on 30 in-the-wild videos from the DAVIS validation set that contain challenging camera and object movement. Best results are in \textbf{bold}, second best are \underline{underlined}.}
    \label{tab:davis}
    \resizebox{\linewidth}{!}{%
        \begin{tabular}{llcccccc}
            \toprule
            \multirow{2}{*}{Method} & \multirow{2}{*}{Backbone} & \multirow{2}{*}{Input} & \multirow{2}{*}{FPS} & \multicolumn{4}{c}{DAVIS Validation Set} \\
            \cmidrule(lr){5-8}
            & & & & PSNR$\uparrow$ & SSIM$\uparrow$ & LPIPS$\downarrow$ & EPE$\downarrow$ \\
            \midrule
            Image Conductor \citep{li2025imageconductor} & AnimateDiff & T, I, 2D Track & 2.98 & 11.30 & 0.214 & 0.664 & 91.64 \\
            Go-With-The-Flow \citep{burgert2025gowiththeflow} & CogVideoX-5B & T, I, 2D Flow & 0.60 & 15.62 & 0.392 & 0.490 & 41.99 \\
            Diffusion-As-Shader \citep{gu2025das} & CogVideoX-5B & I, 3D Track & 0.29 & 15.80 & 0.372 & 0.483 & 40.23 \\
            ATI \citep{wang2025ati} & Wan 2.1-14B & I, 2D Track & 0.23 & 15.33 & 0.374 & 0.473 & 17.41 \\
            MotionStream Teacher \citep{motionstream2025} & Wan 2.1-1.3B & T, I, 2D Track & 0.79 & \textbf{16.61} & 0.477 & 0.427 & \underline{5.35} \\
            MotionStream Causal \citep{motionstream2025} & Wan 2.1-1.3B & T, I, 2D Track & \underline{16.7} & 16.20 & 0.447 & 0.443 & 7.80 \\
            MotionStream Teacher \citep{motionstream2025} & Wan 2.2-5B & T, I, 2D Track & 0.74 & 16.10 & 0.466 & 0.427 & 7.86 \\
            MotionStream Causal \citep{motionstream2025} & Wan 2.2-5B & T, I, 2D Track & 10.4 & \underline{16.30} & 0.456 & 0.438 & 11.18 \\
            \midrule
            Our Teacher & Wan 2.2-5B & T, I, 3D Track & 0.84 & 16.04 & \textbf{0.479} & \textbf{0.404} & \textbf{5.29} \\
            Our Causal & Wan 2.2-5B & T, I, 3D Track & \textbf{20.6} & 15.48 & \underline{0.457} & \underline{0.426} & 5.48 \\
            \bottomrule
        \end{tabular}%
    }%
\end{table}

\subsection{Motion Transfer}
\label{sec:exp_transfer}

Given a source video, we extract its 3D tracks and camera parameters with SpatialTrackerV2, restyle the first frame into a new target appearance with Stable Diffusion XL, and condition our model on the stylized frame together with the extracted motion. As shown in the bottom group of \cref{fig:qualitative_results}, the generated video reproduces the original 3D-consistent motion, including human body dynamics, camera motion, and background parallax, while adopting an entirely new visual style and scene content. The unified 3D track representation thus disentangles motion from appearance: the geometric motion head encodes dynamics independently of visual content, so one motion condition can drive diverse target appearances without retraining or per-instance optimization.

\subsection{Interactive Streaming Control}\label{sec:exp_stream}

\begin{figure}[t!]
    \centering
    \begin{subfigure}[b]{0.5\linewidth}
        \includegraphics[width=\linewidth, trim=0 30 0 0, clip]{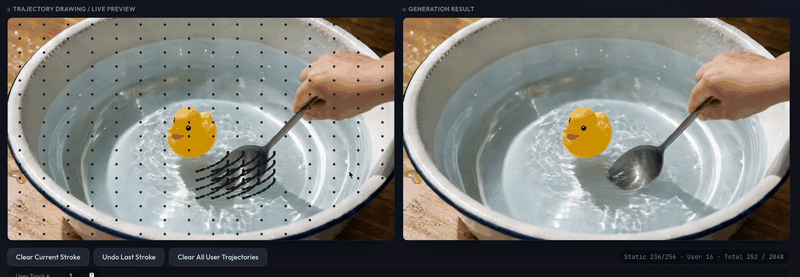}
    \end{subfigure}%
    \begin{subfigure}[b]{0.5\linewidth}
        \includegraphics[width=\linewidth, trim=0 30 0 0, clip]{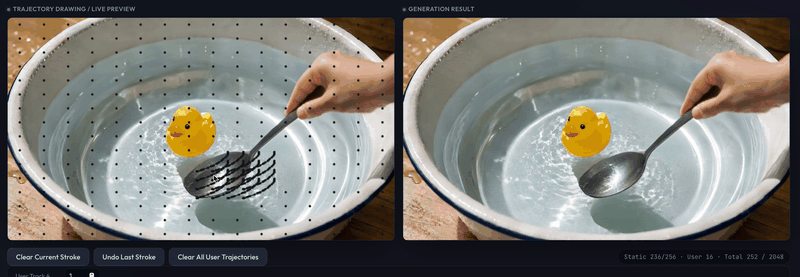}
    \end{subfigure}\\
    \begin{subfigure}[b]{0.5\linewidth}
        \includegraphics[width=\linewidth, trim=0 30 0 0, clip]{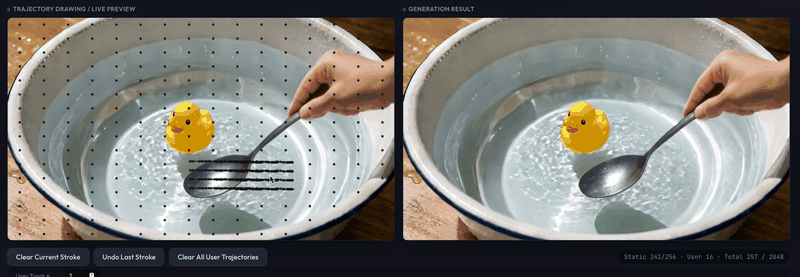}
    \end{subfigure}%
    \begin{subfigure}[b]{0.5\linewidth}
        \includegraphics[width=\linewidth, trim=0 30 0 0, clip]{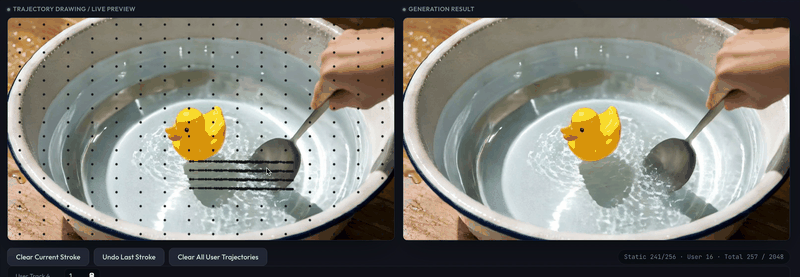}
    \end{subfigure}\\
    \begin{subfigure}[b]{0.5\linewidth}
        \includegraphics[width=\linewidth, trim=0 30 0 0, clip]{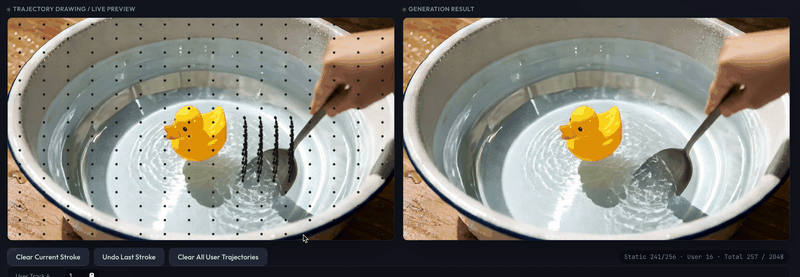}
    \end{subfigure}%
    \begin{subfigure}[b]{0.5\linewidth}
        \includegraphics[width=\linewidth, trim=0 30 0 0, clip]{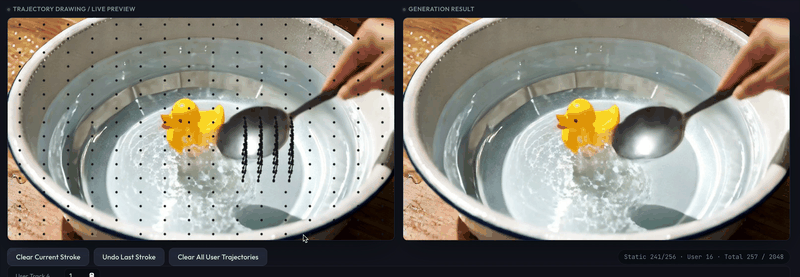}
    \end{subfigure}
    \caption{\textbf{Frames sampled from a real-time interactive streaming session.} In each panel, the left half shows the user's interaction trajectory and the right half the video frame generated from it. Over a 30-second session the user introduces a spoon to poke a rubber duck, revising the trajectory as the model streams coherent frames that follow the evolving input.}
    \label{fig:stream_duck_spoon}
\end{figure}

Our causal student conditions online: track signals arrive sequentially during generation and the model responds within the stream, closing the control loop while the video plays. \cref{fig:stream_duck_spoon} samples such a session. Over 30 seconds the user introduces a spoon to interact with a rubber duck, progressively updating the trajectory while the model generates temporally coherent frames that follow the revised input. This confirms that the segment-compatible motion head, which produces bit-identical output whether tracks are encoded per chunk or over the full sequence, supports seamless online conditioning. Extended results up to 350 frames are provided in \cref{sec:supp:longgen}.

The student sustains over 20\,FPS at 480p on a single high-end GPU at a memory cost independent of the generated length. \cref{tab:efficiency} summarizes the efficiency gains over the teacher: chunk-wise generation with 4-step denoising yields far lower latency and higher throughput than the teacher's full-sequence 50-step sampling. Peak memory is modestly higher because of the rolling KV cache that preserves cross-chunk consistency, but this footprint is bounded and stays flat as the video lengthens, whereas full-sequence attention grows with clip length. Constant memory \emph{with respect to length}, rather than a lower absolute footprint, is what makes unbounded streaming feasible on one GPU.

\subsection{Ablation Study}\label{sec:exp_ablation}

\begin{table}[t!]
    \centering
    \small
    \caption{\textbf{Streaming efficiency comparison.} Latency, peak memory, and throughput measured on 20 validation videos. $w$: window size; $c$: chunk size. Best results are in \textbf{bold}, second best are \underline{underlined}.}
    \label{tab:efficiency}
    \setlength{\tabcolsep}{5pt}
    \begin{tabular}{@{}lccc@{}}
        \toprule
        Model & Latency (s) $\downarrow$ & Memory (GB) $\downarrow$ & Throughput (FPS) $\uparrow$ \\
        \midrule
        Teacher (Full Attention) & $34.92 \pm 0.351$ & $\textbf{23.9} \pm \textbf{0.82}$ & $0.84 \pm 0.001$ \\
        \midrule
        Causal ($w{=}21,\; c{=}4$) & $\underline{0.756} \pm \underline{0.006}$ & $31.8 \pm 0.72$ & $\underline{20.6} \pm \underline{0.03}$ \\
        Causal ($w{=}9,\; c{=}4$) & $\underline{0.756} \pm \underline{0.006}$ & $\underline{30.1} \pm \underline{0.65}$ & $\textbf{21.3} \pm \textbf{0.01}$ \\
        Causal ($w{=}21,\; c{=}1$) & $\textbf{0.411} \pm \textbf{0.005}$ & $31.4 \pm 0.68$ & $10.5 \pm 0.02$ \\
        \bottomrule
    \end{tabular}
\end{table}

\begin{table}[t!]
    \centering
    \small
    \caption{\textbf{Ablations of core designs.} ``Full'' is our complete teacher pipeline. Metrics are reported on 20 clean validation videos with 81 frames from \dataset.}
    \label{tab:ablation}
    \setlength{\tabcolsep}{3pt}
    \begin{tabular}{lcccc}
        \toprule
        Variant & PSNR$\uparrow$ & SSIM$\uparrow$ & LPIPS$\downarrow$ & EPE$\downarrow$ \\
        \midrule
        Full (3D + ID + 2-stage) & $\textbf{19.13} \pm \textbf{3.16}$ & $\textbf{0.661} \pm \textbf{0.174}$ & $\textbf{0.200} \pm \textbf{0.086}$ & $\textbf{0.985} \pm \textbf{1.095}$ \\
        \midrule
        2D tracks only & $18.95 \pm 3.09$ & $0.649 \pm 0.172$ & $0.225 \pm 0.097$ & $1.523 \pm 1.224$ \\
        w/o track-ID embedding & $16.55 \pm 3.32$ & $0.592 \pm 0.180$ & $0.283 \pm 0.090$ & $2.455 \pm 1.148$ \\
        Single-stage training (no curriculum) & $17.62 \pm 3.41$ & $0.613 \pm 0.185$ & $0.248 \pm 0.093$ & $3.217 \pm 1.562$ \\
        \bottomrule
    \end{tabular}
\end{table}

\begin{wrapfigure}{r}{0.5\linewidth}
    \centering
    \vspace{-30pt}
    \includegraphics[width=\linewidth]{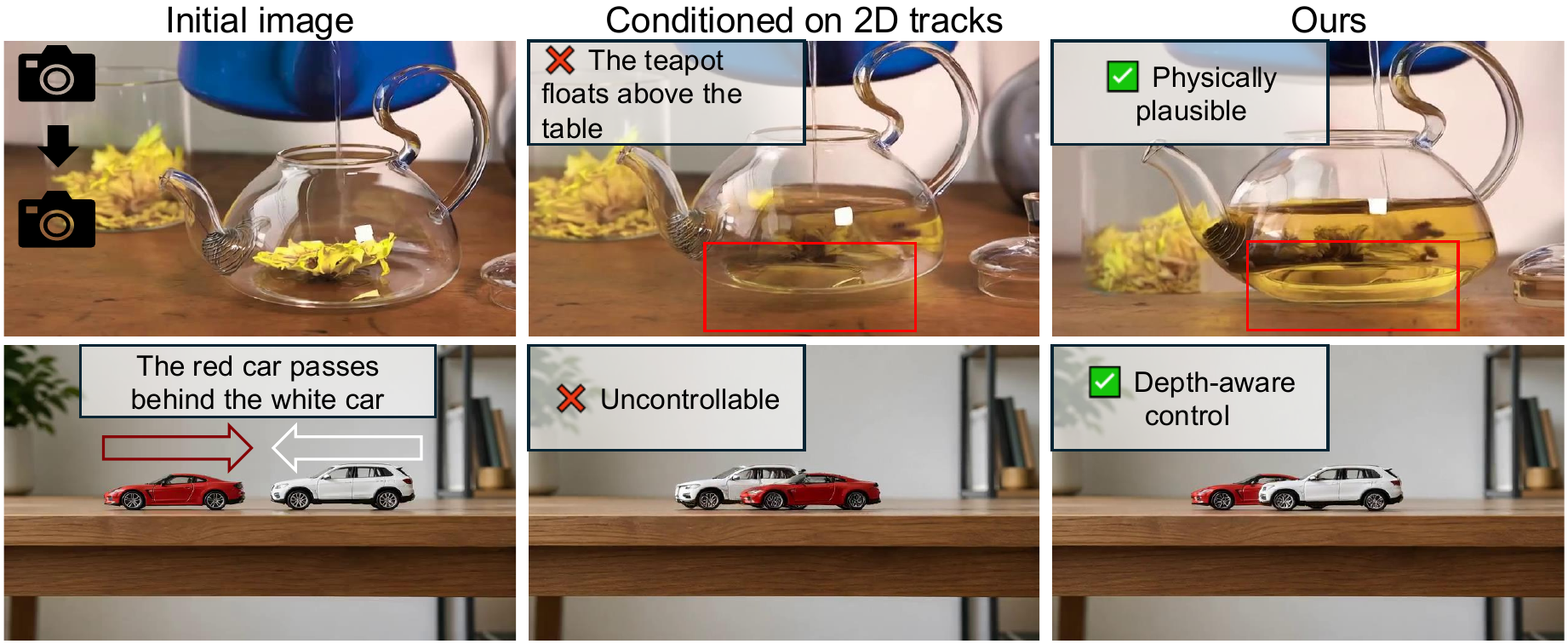}
    \caption{\textbf{Advantage of 3D over 2D track conditioning.} \textit{Top:} under a camera move-down, 2D tracks leave the teapot floating above the table through depth ambiguity, whereas our 3D-aware conditioning keeps it grounded. \textit{Bottom:} instructed that the red car should pass behind the white car, 2D conditioning resolves the depth ordering incorrectly, while our depth-aware control respects the intended occlusion.}
    \label{fig:depth}
\end{wrapfigure}

\paragraph{3D versus 2D motion conditioning.}
We compare the full model against a 2D-only variant that drops the depth branch and camera-aware components, conditioning on projected 2D tracks alone under the same motion-head capacity budget. Quantitatively it costs 0.5 EPE and 0.025 LPIPS (\cref{tab:ablation}); qualitatively the failure is more telling. As \cref{fig:depth} shows, image-plane conditioning suffers from depth ambiguity and yields physically implausible configurations, while 3D tracks with explicit depth produce motion that respects scene geometry and supports fine-grained control over depth-dependent interactions. This confirms the value of disentangling object motion from camera-induced image motion. Additional comparisons with and without track conditioning appear in \cref{sec:supp:track_effect}.

\paragraph{Motion head.}
Removing the track-ID embedding causes the largest single-component drop, with EPE rising from 0.985 to 2.455 and LPIPS from 0.200 to 0.283, indicating that sinusoidal identity encoding is what lets the model tell individual trajectories apart and follow each faithfully.

\paragraph{Training curriculum.}
Single-stage training at full resolution degrades every metric (EPE 3.217 versus 0.985, SSIM 0.613 versus 0.661), confirming that the low-resolution warm-up stabilizes motion alignment before scaling to higher resolution and longer horizons. Further ablations on distillation hyperparameters, random seeds, and the number of control points are provided in \cref{sec:supp:ablation_distill,sec:supp:seeds,sec:supp:pointnum}.

\section{Conclusion}
\label{sec:conclusion}

We presented \modelname, a controllable video generation framework that unifies camera motion, object trajectories, and per-point depth into a single 3D-track interface for 4D control. Combining large-scale in-the-wild 3D motion supervision, a temporally separable Geometric Motion Head for joint camera, object, and depth control, and a causal streaming distillation that reduces 50-step inference to 4 steps at memory independent of length, our approach achieves state-of-the-art motion-control precision while enabling, to our knowledge, the first real-time 4D-controllable streaming video synthesis on a single GPU. These results suggest that grounding generative video in explicit 3D geometry, coupled with efficient causal inference, is a viable path toward interactive world models with closed-loop spatiotemporal control. We discuss remaining failure modes in \cref{sec:supp:failure} and outline limitations and future directions in \cref{sec:limitations}.

\paragraph{Acknowledgment.}

S. Li and Y. Zhu are supported in part by the National Natural Science Foundation of China (62376009) and the Beijing Nova program.

\bibliography{reference_header,reference}

\begin{thebibliography}{43}
\providecommand{\natexlab}[1]{#1}
\providecommand{\url}[1]{\texttt{#1}}
\expandafter\ifx\csname urlstyle\endcsname\relax
  \providecommand{\doi}[1]{doi: #1}\else
  \providecommand{\doi}{doi: \begingroup \urlstyle{rm}\Url}\fi

\bibitem[Black et~al.(2024)Black, Brown, Driess, Esmail, Equi, Finn, Fusai, Groom, Hausman, Ichter, Jakubczak, Jones, Ke, Levine, Li-Bell, Mothukuri, Nair, Pertsch, Shi, Tanner, Vuong, Walling, Wang, and Zhilinsky]{black2024pi0}
Kevin Black, Noah Brown, Danny Driess, Adnan Esmail, Michael Equi, Chelsea Finn, Niccolo Fusai, Lachy Groom, Karol Hausman, Brian Ichter, Szymon Jakubczak, Tim Jones, Liyiming Ke, Sergey Levine, Adrian Li-Bell, Mohith Mothukuri, Suraj Nair, Karl Pertsch, Lucy~Xiaoyang Shi, James Tanner, Quan Vuong, Anna Walling, Haohuan Wang, and Ury Zhilinsky.
\newblock $\pi_0$: A vision-language-action flow model for general robot control.
\newblock \emph{arXiv preprint arXiv:2410.24164}, 2024.

\bibitem[Bruce et~al.(2024)Bruce, Dennis, Edwards, Parker-Holder, Shi, Hughes, Lai, Mavalankar, Steigerwald, Apps, Aytar, Bechtle, Behbahani, Chan, Heess, Gonzalez, Osindero, Ozair, Reed, Zhang, Zolna, Clune, Freitas, Singh, and Rockt{\"a}schel]{pmlr-v235-bruce24a}
Jake Bruce, Michael~D. Dennis, Ashley Edwards, Jack Parker-Holder, Yuge Shi, Edward Hughes, Matthew Lai, Aditi Mavalankar, Richie Steigerwald, Chris Apps, Yusuf Aytar, Sarah Maria~Elisabeth Bechtle, Feryal Behbahani, Stephanie~C.Y. Chan, Nicolas Heess, Lucy Gonzalez, Simon Osindero, Sherjil Ozair, Scott Reed, Jingwei Zhang, Konrad Zolna, Jeff Clune, Nando~De Freitas, Satinder Singh, and Tim Rockt{\"a}schel.
\newblock Genie: Generative interactive environments.
\newblock In \emph{International Conference on Machine Learning (ICML)}, 2024.

\bibitem[Burgert et~al.(2025)Burgert, Xu, Xian, Pilarski, Clausen, He, Ma, Deng, Li, Mousavi, Ryoo, Debevec, and Yu]{burgert2025gowiththeflow}
Ryan Burgert, Yuancheng Xu, Wenqi Xian, Oliver Pilarski, Pascal Clausen, Mingming He, Li~Ma, Yitong Deng, Lingxiao Li, Mohsen Mousavi, Michael Ryoo, Paul Debevec, and Ning Yu.
\newblock Go-with-the-flow: Motion-controllable video diffusion models using real-time warped noise.
\newblock In \emph{Conference on Computer Vision and Pattern Recognition (CVPR)}, 2025.

\bibitem[Chen et~al.(2024)Chen, Mart{\'\i}~Mons{\'o}, Du, Simchowitz, Tedrake, and Sitzmann]{chen2024diffusion}
Boyuan Chen, Diego Mart{\'\i}~Mons{\'o}, Yilun Du, Max Simchowitz, Russ Tedrake, and Vincent Sitzmann.
\newblock Diffusion forcing: Next-token prediction meets full-sequence diffusion.
\newblock In \emph{Advances in Neural Information Processing Systems (NeurIPS)}, 2024.

\bibitem[Du et~al.(2023)Du, Yang, Dai, Dai, Nachum, Tenenbaum, Schuurmans, and Abbeel]{du2024learning}
Yilun Du, Sherry Yang, Bo~Dai, Hanjun Dai, Ofir Nachum, Joshua~B Tenenbaum, Dale Schuurmans, and Pieter Abbeel.
\newblock Learning universal policies via text-guided video generation.
\newblock In \emph{Advances in Neural Information Processing Systems (NeurIPS)}, 2023.

\bibitem[Geng et~al.(2025)Geng, Herrmann, Hur, Cole, Zhang, Pfaff, Lopez-Guevara, Aytar, Rubinstein, Sun, et~al.]{geng2025motion}
Daniel Geng, Charles Herrmann, Junhwa Hur, Forrester Cole, Serena Zhang, Tobias Pfaff, Tatiana Lopez-Guevara, Yusuf Aytar, Michael Rubinstein, Chen Sun, et~al.
\newblock Motion prompting: Controlling video generation with motion trajectories.
\newblock In \emph{Conference on Computer Vision and Pattern Recognition (CVPR)}, 2025.

\bibitem[Gu et~al.(2025)Gu, Yan, Lu, Li, Dou, Si, Dong, Liu, Lin, Liu, Wang, and Liu]{gu2025das}
Zekai Gu, Rui Yan, Jiahao Lu, Peng Li, Zhiyang Dou, Chenyang Si, Zhen Dong, Qifeng Liu, Cheng Lin, Ziwei Liu, Wenping Wang, and Yuan Liu.
\newblock Diffusion as shader: 3d-aware video diffusion for versatile video generation control.
\newblock In \emph{ACM SIGGRAPH Conference Proceedings}, 2025.

\bibitem[Ha \& Schmidhuber(2018)Ha and Schmidhuber]{ha2018worldmodels}
David Ha and J{\"u}rgen Schmidhuber.
\newblock Recurrent world models facilitate policy evolution.
\newblock In \emph{Advances in Neural Information Processing Systems (NeurIPS)}, 2018.

\bibitem[He et~al.(2024)He, Xu, Guo, Wetzstein, Dai, Li, and Yang]{he2024cameractrl}
Hao He, Yinghao Xu, Yuwei Guo, Gordon Wetzstein, Bo~Dai, Hongsheng Li, and Ceyuan Yang.
\newblock Cameractrl: Enabling camera control for text-to-video generation.
\newblock \emph{arXiv preprint arXiv:2404.02101}, 2024.

\bibitem[Ho et~al.(2022)Ho, Salimans, Gritsenko, Chan, Norouzi, and Fleet]{ho2022video}
Jonathan Ho, Tim Salimans, Alexey Gritsenko, William Chan, Mohammad Norouzi, and David~J. Fleet.
\newblock Video diffusion models.
\newblock In \emph{Advances in Neural Information Processing Systems (NeurIPS)}, 2022.

\bibitem[Hu et~al.(2022)Hu, Shen, Wallis, Allen-Zhu, Li, Wang, Wang, and Chen]{hu2022lora}
Edward~J Hu, Yelong Shen, Phillip Wallis, Zeyuan Allen-Zhu, Yuanzhi Li, Shean Wang, Liang Wang, and Weizhu Chen.
\newblock Lora: Low-rank adaptation of large language models.
\newblock In \emph{International Conference on Learning Representations (ICLR)}, 2022.

\bibitem[Hu et~al.(2020)Hu, Anderson, Li, Sun, Carr, Ragan-Kelley, and Durand]{hu2019difftaichi}
Yuanming Hu, Luke Anderson, Tzu-Mao Li, Qi~Sun, Nathan Carr, Jonathan Ragan-Kelley, and Fr{\'e}do Durand.
\newblock Difftaichi: Differentiable programming for physical simulation.
\newblock In \emph{International Conference on Learning Representations (ICLR)}, 2020.

\bibitem[Huang et~al.(2025)Huang, Li, He, Zhou, and Shechtman]{huang2025selfforcing}
Xun Huang, Zhengqi Li, Guande He, Mingyuan Zhou, and Eli Shechtman.
\newblock Self forcing: Bridging the train-test gap in autoregressive video diffusion.
\newblock In \emph{Advances in Neural Information Processing Systems (NeurIPS)}, 2025.

\bibitem[Karaev et~al.(2025)Karaev, Makarov, Wang, Neverova, Vedaldi, and Rupprecht]{karaev2024cotracker3}
Nikita Karaev, Yuri Makarov, Jianyuan Wang, Natalia Neverova, Andrea Vedaldi, and Christian Rupprecht.
\newblock Cotracker3: Simpler and better point tracking by pseudo-labelling real videos.
\newblock In \emph{International Conference on Computer Vision (ICCV)}, 2025.

\bibitem[Kondratyuk et~al.(2024)Kondratyuk, Yu, Gu, Lezama, Huang, Schindler, Hornung, Birodkar, Yan, Chiu, Somandepalli, Akbari, Alon, Cheng, Dillon, Gupta, Hahn, Hauth, Hendon, Martinez, Minnen, Sirotenko, Sohn, Yang, Adam, Yang, Essa, Wang, Ross, Seybold, and Jiang]{kondratyuk2024videopoet}
Dan Kondratyuk, Lijun Yu, Xiuye Gu, Jos{\'e} Lezama, Jonathan Huang, Grant Schindler, Rachel Hornung, Vighnesh Birodkar, Jimmy Yan, Ming-Chang Chiu, Krishna Somandepalli, Hassan Akbari, Yair Alon, Yong Cheng, Joshua~V. Dillon, Agrim Gupta, Meera Hahn, Anja Hauth, David Hendon, Alonso Martinez, David Minnen, Mikhail Sirotenko, Kihyuk Sohn, Xuan Yang, Hartwig Adam, Ming-Hsuan Yang, Irfan Essa, Huisheng Wang, David~A. Ross, Bryan Seybold, and Lu~Jiang.
\newblock Videopoet: A large language model for zero-shot video generation.
\newblock In \emph{International Conference on Machine Learning (ICML)}, 2024.

\bibitem[Kong et~al.(2024)Kong, Tian, Zhang, Min, Dai, Zhou, Xiong, Li, Wu, Zhang, et~al.]{kong2024hunyuanvideo}
Weijie Kong, Qi~Tian, Zijian Zhang, Rox Min, Zuozhuo Dai, Jin Zhou, Jiangfeng Xiong, Xin Li, Bo~Wu, Jianwei Zhang, et~al.
\newblock Hunyuanvideo: A systematic framework for large video generative models.
\newblock \emph{arXiv preprint arXiv:2412.03603}, 2024.

\bibitem[Lee et~al.(2026)Lee, Zhang, Huang, Wang, Lee, Huang, Shechtman, and Li]{lee2026generative}
Yao-Chih Lee, Zhoutong Zhang, Jiahui Huang, Jui-Hsien Wang, Joon-Young Lee, Jia-Bin Huang, Eli Shechtman, and Zhengqi Li.
\newblock Generative video motion editing with 3d point tracks.
\newblock In \emph{Conference on Computer Vision and Pattern Recognition (CVPR)}, 2026.

\bibitem[Li et~al.(2025)Li, Wang, Zhang, Wang, Yuan, Xie, Shan, and Zou]{li2025imageconductor}
Yaowei Li, Xintao Wang, Zhaoyang Zhang, Zhouxia Wang, Ziyang Yuan, Liangbin Xie, Ying Shan, and Yuexian Zou.
\newblock Image conductor: Precision control for interactive video synthesis.
\newblock In \emph{AAAI Conference on Artificial Intelligence (AAAI)}, 2025.

\bibitem[Li et~al.(2019)Li, Wu, Tedrake, Tenenbaum, and Torralba]{li2019learning}
Yunzhu Li, Jiajun Wu, Russ Tedrake, Joshua~B Tenenbaum, and Antonio Torralba.
\newblock Learning particle dynamics for manipulating rigid bodies, deformable objects, and fluids.
\newblock In \emph{International Conference on Learning Representations (ICLR)}, 2019.

\bibitem[Nan et~al.(2025)Nan, Xie, Zhou, Fan, Yang, Chen, Li, Yang, and Tai]{nan2025openvid}
Kepan Nan, Rui Xie, Penghao Zhou, Tiehan Fan, Zhenheng Yang, Zhijie Chen, Xiang Li, Jian Yang, and Ying Tai.
\newblock Openvid-1m: A large-scale high-quality dataset for text-to-video generation.
\newblock In \emph{International Conference on Learning Representations (ICLR)}, 2025.

\bibitem[Peebles \& Xie(2023)Peebles and Xie]{peebles2023scalable}
William Peebles and Saining Xie.
\newblock Scalable diffusion models with transformers.
\newblock In \emph{International Conference on Computer Vision (ICCV)}, 2023.

\bibitem[Perazzi et~al.(2016)Perazzi, Pont-Tuset, McWilliams, {Van Gool}, Gross, and Sorkine-Hornung]{perazzi2016davis}
Federico Perazzi, Jordi Pont-Tuset, Brian McWilliams, Luc {Van Gool}, Markus Gross, and Alexander Sorkine-Hornung.
\newblock A benchmark dataset and evaluation methodology for video object segmentation.
\newblock In \emph{Conference on Computer Vision and Pattern Recognition (CVPR)}, 2016.

\bibitem[Salimans \& Ho(2022)Salimans and Ho]{salimans2022progressive}
Tim Salimans and Jonathan Ho.
\newblock Progressive distillation for fast sampling of diffusion models.
\newblock In \emph{International Conference on Learning Representations (ICLR)}, 2022.

\bibitem[Sch{\"o}nberger \& Frahm(2016)Sch{\"o}nberger and Frahm]{schonberger2016colmap}
Johannes~L. Sch{\"o}nberger and Jan-Michael Frahm.
\newblock Structure-from-motion revisited.
\newblock In \emph{Conference on Computer Vision and Pattern Recognition (CVPR)}, 2016.

\bibitem[Shi et~al.(2024)Shi, Wang, Ye, Long, Li, and Yang]{shi2024mvdream}
Yichun Shi, Peng Wang, Jianglong Ye, Mai Long, Kejie Li, and Xiao Yang.
\newblock Mvdream: Multi-view diffusion for 3d generation.
\newblock In \emph{International Conference on Learning Representations (ICLR)}, 2024.

\bibitem[Shin et~al.(2026)Shin, Li, Zhang, Zhu, Park, Shechtman, and Huang]{motionstream2025}
Joonghyuk Shin, Zhengqi Li, Richard Zhang, Jun-Yan Zhu, Jaesik Park, Eli Shechtman, and Xun Huang.
\newblock Motionstream: Real-time video generation with interactive motion controls.
\newblock In \emph{International Conference on Learning Representations (ICLR)}, 2026.

\bibitem[Teng et~al.(2023)Teng, Xie, Wu, Han, Li, and Liu]{teng2023dragavideo}
Yao Teng, Enze Xie, Yue Wu, Haoyu Han, Zhenguo Li, and Xihui Liu.
\newblock Drag-a-video: Non-rigid video editing with point-based interaction.
\newblock \emph{arXiv preprint arXiv:2312.02936}, 2023.

\bibitem[Wan et~al.(2025)Wan, Wang, Ai, Wen, Mao, Xie, Chen, Yu, Zhao, Yang, et~al.]{wan2025wan}
Team Wan, Ang Wang, Baole Ai, Bin Wen, Chaojie Mao, Chen-Wei Xie, Di~Chen, Feiwu Yu, Haiming Zhao, Jianxiao Yang, et~al.
\newblock Wan: Open and advanced large-scale video generative models.
\newblock \emph{arXiv preprint arXiv:2503.20314}, 2025.

\bibitem[Wang et~al.(2025)Wang, Huang, Fang, Yang, and Ma]{wang2025ati}
Angtian Wang, Haibin Huang, Zhiyuan Fang, Yiding Yang, and Chongyang Ma.
\newblock Ati: Any trajectory instruction for controllable video generation.
\newblock \emph{arXiv preprint arXiv:2505.22944}, 2025.

\bibitem[Wang et~al.(2024{\natexlab{a}})Wang, Leroy, Cabon, Chidlovskii, and Revaud]{wang2024dust3r}
Shuzhe Wang, Vincent Leroy, Yohann Cabon, Boris Chidlovskii, and J{\'e}r{\^o}me Revaud.
\newblock Dust3r: Geometric 3d vision made easy.
\newblock In \emph{Conference on Computer Vision and Pattern Recognition (CVPR)}, 2024{\natexlab{a}}.

\bibitem[Wang et~al.(2024{\natexlab{b}})Wang, Yuan, Wang, Li, Chen, Xia, Luo, and Shan]{wang2024motionctrl}
Zhouxia Wang, Ziyang Yuan, Xintao Wang, Yaowei Li, Tianshui Chen, Menghan Xia, Ping Luo, and Ying Shan.
\newblock Motionctrl: A unified and flexible motion controller for video generation.
\newblock In \emph{ACM SIGGRAPH Conference Proceedings}, 2024{\natexlab{b}}.

\bibitem[Wu et~al.(2024{\natexlab{a}})Wu, Yi, Fang, Xie, Zhang, Wei, Liu, Tian, and Wang]{wu20244dgaussians}
Guanjun Wu, Taoran Yi, Jiemin Fang, Lingxi Xie, Xiaopeng Zhang, Wei Wei, Wenyu Liu, Qi~Tian, and Xinggang Wang.
\newblock 4d gaussian splatting for real-time dynamic scene rendering.
\newblock In \emph{Conference on Computer Vision and Pattern Recognition (CVPR)}, 2024{\natexlab{a}}.

\bibitem[Wu et~al.(2024{\natexlab{b}})Wu, Li, Gu, Zhao, He, Zhang, Shou, Li, Gao, and Zhang]{wu2024draganything}
Weijia Wu, Zhuang Li, Yuchao Gu, Rui Zhao, Yefei He, David~Junhao Zhang, Mike~Zheng Shou, Yan Li, Tingting Gao, and Di~Zhang.
\newblock Draganything: Motion control for anything using entity representation.
\newblock In \emph{European Conference on Computer Vision (ECCV)}, 2024{\natexlab{b}}.

\bibitem[Xiao et~al.(2024)Xiao, Tian, Chen, Han, and Lewis]{xiao2023streamingllm}
Guangxuan Xiao, Yuandong Tian, Beidi Chen, Song Han, and Mike Lewis.
\newblock Efficient streaming language models with attention sinks.
\newblock In \emph{International Conference on Learning Representations (ICLR)}, 2024.

\bibitem[Xiao et~al.(2025)Xiao, Wang, Xue, Karaev, Makarov, Kang, Zhu, Bao, Shen, and Zhou]{xiao2025spatialtrackerv2}
Yuxi Xiao, Jianyuan Wang, Nan Xue, Nikita Karaev, Yuri Makarov, Bingyi Kang, Xing Zhu, Hujun Bao, Yujun Shen, and Xiaowei Zhou.
\newblock Spatialtrackerv2: 3d point tracking made easy.
\newblock In \emph{International Conference on Computer Vision (ICCV)}, 2025.

\bibitem[Yang et~al.(2024)Yang, Du, Ghasemipour, Tompson, Kaelbling, Schuurmans, and Abbeel]{yang2024unisim}
Sherry Yang, Yilun Du, Kamyar Ghasemipour, Jonathan Tompson, Leslie Kaelbling, Dale Schuurmans, and Pieter Abbeel.
\newblock Learning interactive real-world simulators.
\newblock In \emph{International Conference on Learning Representations (ICLR)}, 2024.

\bibitem[Yin et~al.(2024{\natexlab{a}})Yin, Gharbi, Park, Zhang, Shechtman, Durand, and Freeman]{yin2024improved}
Tianwei Yin, Micha{\"e}l Gharbi, Taesung Park, Richard Zhang, Eli Shechtman, Fredo Durand, and William~T Freeman.
\newblock Improved distribution matching distillation for fast image synthesis.
\newblock In \emph{Advances in Neural Information Processing Systems (NeurIPS)}, 2024{\natexlab{a}}.

\bibitem[Yin et~al.(2024{\natexlab{b}})Yin, Gharbi, Zhang, Shechtman, Durand, Freeman, and Park]{yin2024dmd}
Tianwei Yin, Micha{\"e}l Gharbi, Richard Zhang, Eli Shechtman, Fr{\'e}do Durand, William~T Freeman, and Taesung Park.
\newblock One-step diffusion with distribution matching distillation.
\newblock In \emph{Conference on Computer Vision and Pattern Recognition (CVPR)}, 2024{\natexlab{b}}.

\bibitem[Yin et~al.(2025)Yin, Zhang, Zhang, Freeman, Durand, Shechtman, and Huang]{yin2025causvid}
Tianwei Yin, Qiang Zhang, Richard Zhang, William~T Freeman, Fredo Durand, Eli Shechtman, and Xun Huang.
\newblock From slow bidirectional to fast autoregressive video diffusion models.
\newblock In \emph{Conference on Computer Vision and Pattern Recognition (CVPR)}, 2025.

\bibitem[Yin et~al.(2023)Yin, Zhang, Chen, Cai, Yu, Wang, Chen, and Shen]{yin2023metric3d}
Wei Yin, Chi Zhang, Hao Chen, Zhipeng Cai, Gang Yu, Kaixuan Wang, Xiaozhi Chen, and Chunhua Shen.
\newblock Metric3d: Towards zero-shot metric 3d prediction from a single image.
\newblock In \emph{International Conference on Computer Vision (ICCV)}, 2023.

\bibitem[Yu et~al.(2023)Yu, Cheng, Sohn, Lezama, Zhang, Chang, Hauptmann, Yang, Hao, Essa, and Jiang]{yu2023magvit}
Lijun Yu, Yong Cheng, Kihyuk Sohn, Jos{\'e} Lezama, Han Zhang, Huiwen Chang, Alexander~G. Hauptmann, Ming-Hsuan Yang, Yuan Hao, Irfan Essa, and Lu~Jiang.
\newblock Magvit: Masked generative video transformer.
\newblock In \emph{Conference on Computer Vision and Pattern Recognition (CVPR)}, 2023.

\bibitem[Zhang et~al.(2018)Zhang, Isola, Efros, Shechtman, and Wang]{zhang2018perceptual}
Richard Zhang, Phillip Isola, Alexei~A. Efros, Eli Shechtman, and Oliver Wang.
\newblock The unreasonable effectiveness of deep features as a perceptual metric.
\newblock In \emph{Conference on Computer Vision and Pattern Recognition (CVPR)}, 2018.

\bibitem[Zhang et~al.(2023)Zhang, Sheng, Zhou, Chen, Zheng, Cai, Song, Tian, Re, Barrett, Wang, and Chen]{zhang2024h2o}
Zhenyu Zhang, Ying Sheng, Tianyi Zhou, Tianlong Chen, Lianmin Zheng, Ruisi Cai, Zhao Song, Yuandong Tian, Christopher Re, Clark Barrett, Zhangyang Wang, and Beidi Chen.
\newblock H2o: Heavy-hitter oracle for efficient generative inference of large language models.
\newblock In \emph{Advances in Neural Information Processing Systems (NeurIPS)}, 2023.

\end{thebibliography}
\bibliographystyle{iclr2027_conference}

\clearpage
\appendix
\renewcommand\thefigure{A\arabic{figure}}
\setcounter{figure}{0}
\renewcommand\thetable{A\arabic{table}}
\setcounter{table}{0}
\renewcommand\theequation{A\arabic{equation}}
\setcounter{equation}{0}
\pagenumbering{arabic}
\renewcommand*{\thepage}{A\arabic{page}}
\setcounter{footnote}{0}

\section*{Appendix}
\startcontents[appendices]
\printcontents[appendices]{l}{1}{\setcounter{tocdepth}{2}}

\section{More Visualization}

\subsection{Long Video Generation}\label{sec:supp:longgen}

\begin{figure}[t!]
    \centering
    \begin{minipage}[c]{0.77\linewidth}
        \centering
        \begin{tabular}{@{}c@{}c@{}c@{}}
            \includegraphics[width=0.33\linewidth]{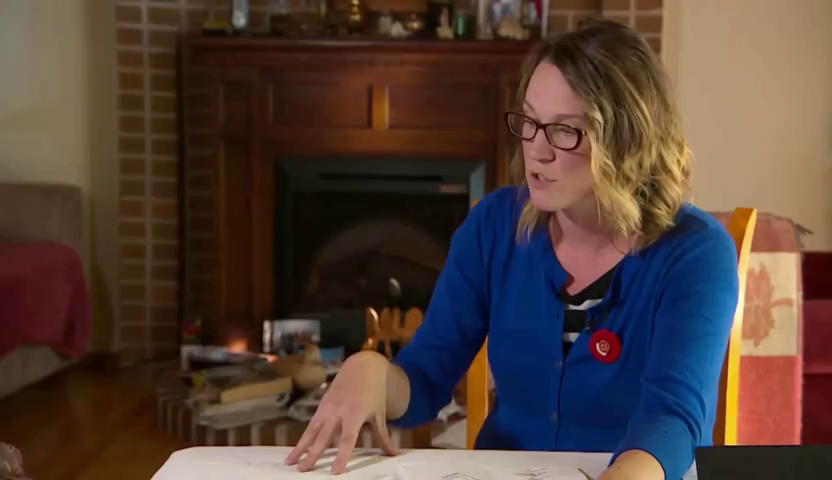} &
            \includegraphics[width=0.33\linewidth]{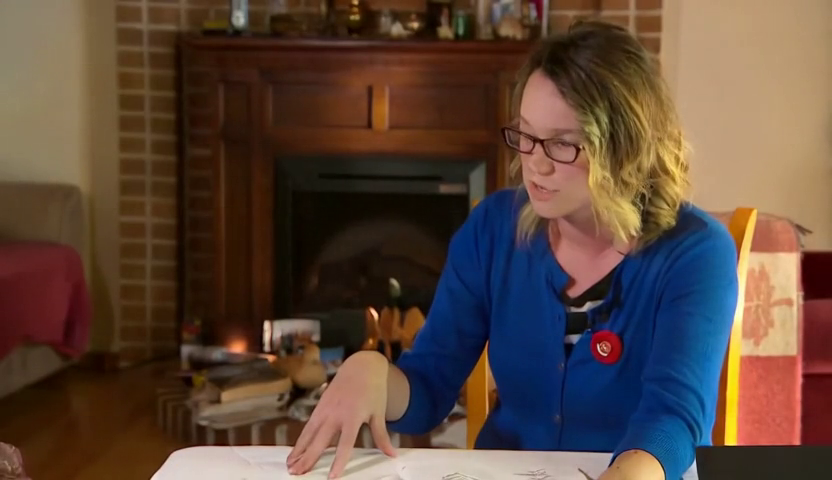} &
            \includegraphics[width=0.33\linewidth]{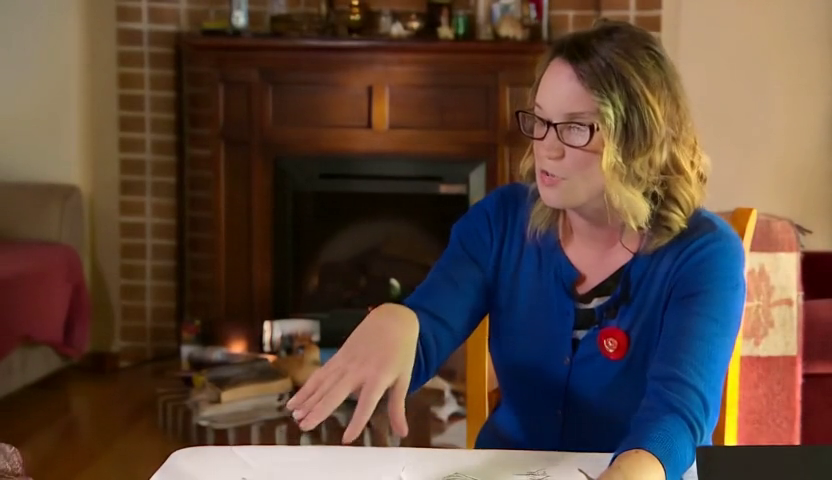} \\
            \includegraphics[width=0.33\linewidth]{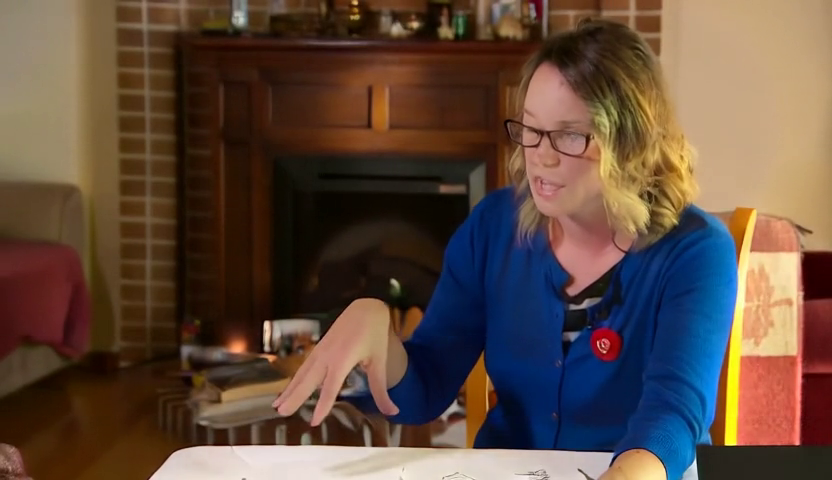} &
            \includegraphics[width=0.33\linewidth]{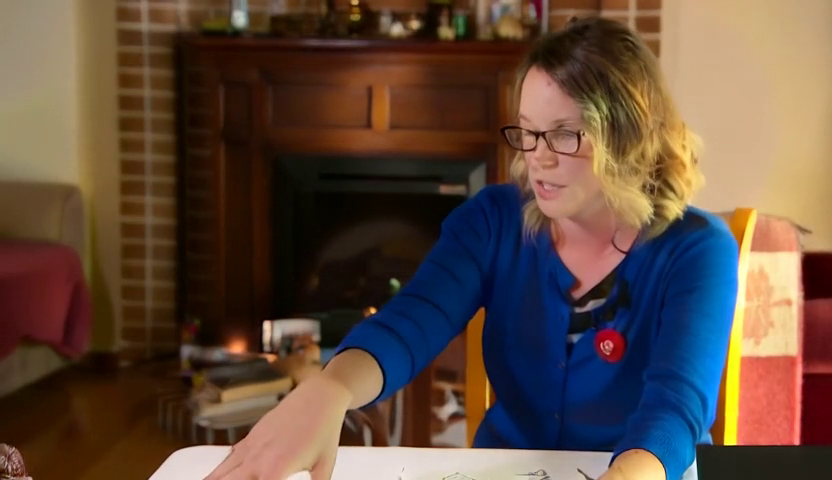} &
            \includegraphics[width=0.33\linewidth]{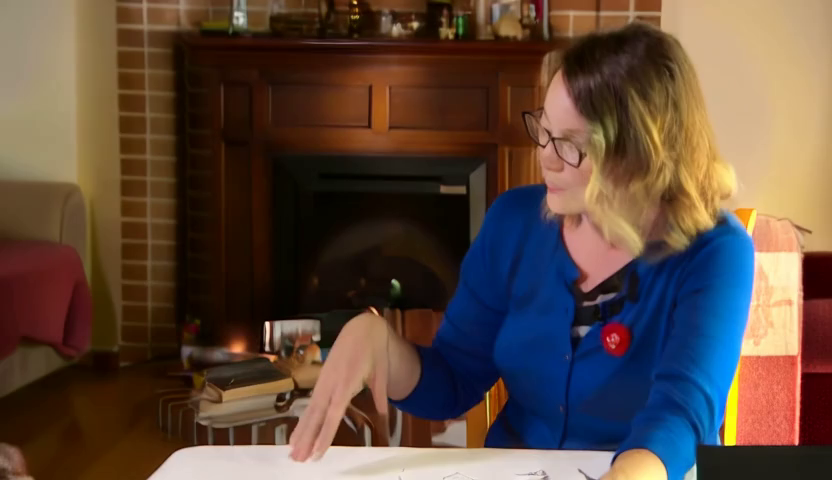} \\
        \end{tabular}
    \end{minipage}%
    \begin{minipage}[c]{0.23\linewidth}
        \centering
        \includegraphics[width=\linewidth, trim={17.7cm 0 0 0}, clip]{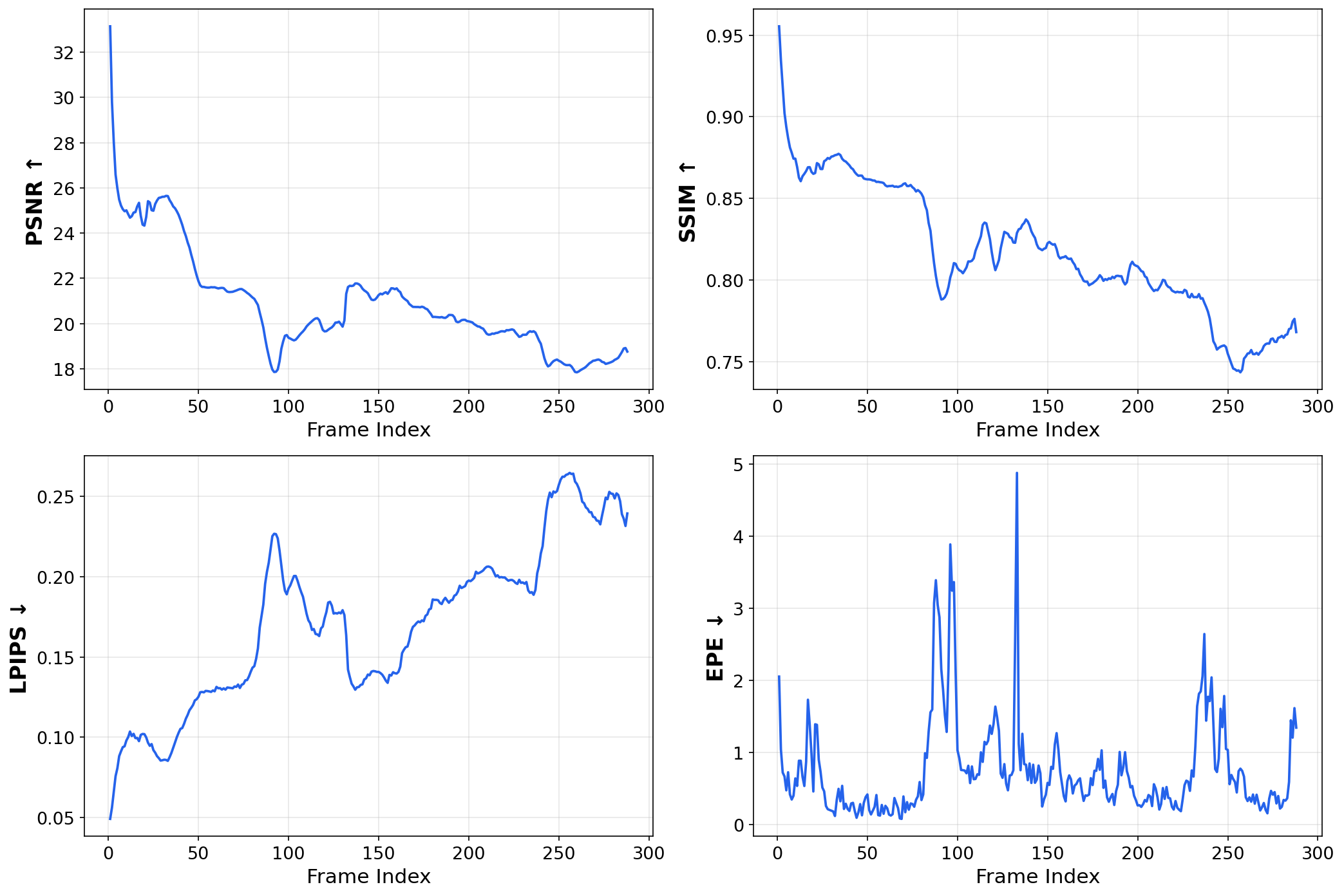}
    \end{minipage}
    \noindent\dashrule{\linewidth}{1pt}{3pt}{2pt}
    \begin{minipage}[c]{0.77\linewidth}
        \centering
        \begin{tabular}{@{}c@{}c@{}c@{}}
            \includegraphics[width=0.33\linewidth]{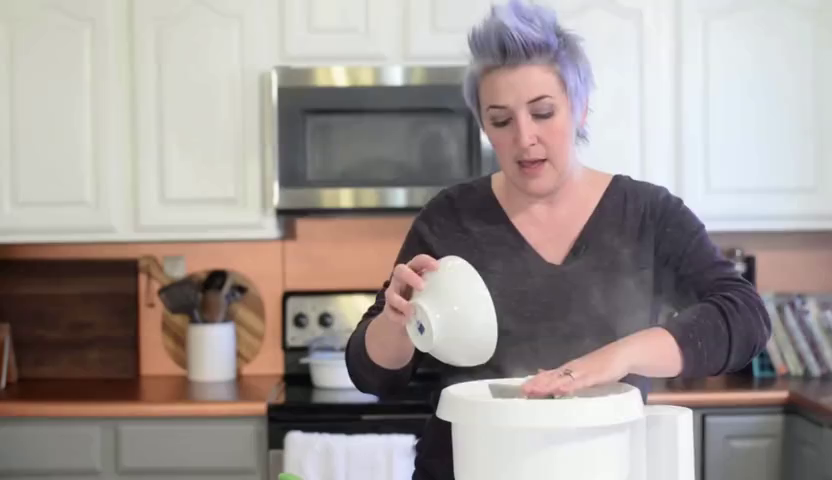} &
            \includegraphics[width=0.33\linewidth]{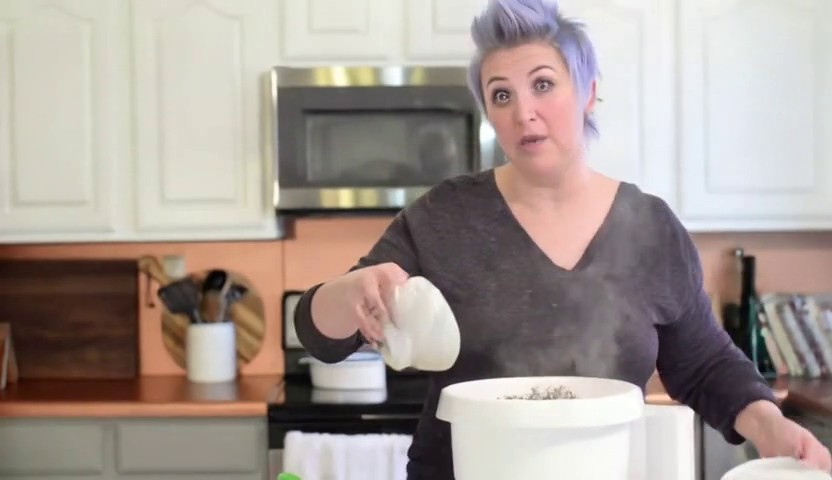} &
            \includegraphics[width=0.33\linewidth]{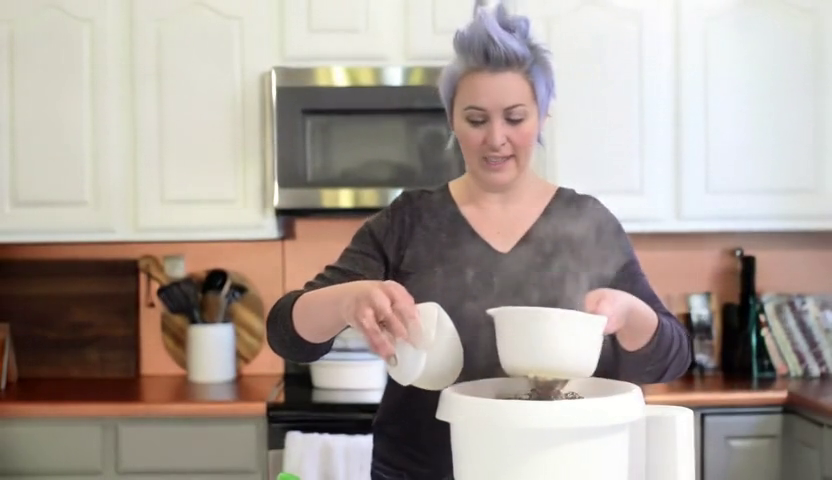} \\
            \includegraphics[width=0.33\linewidth]{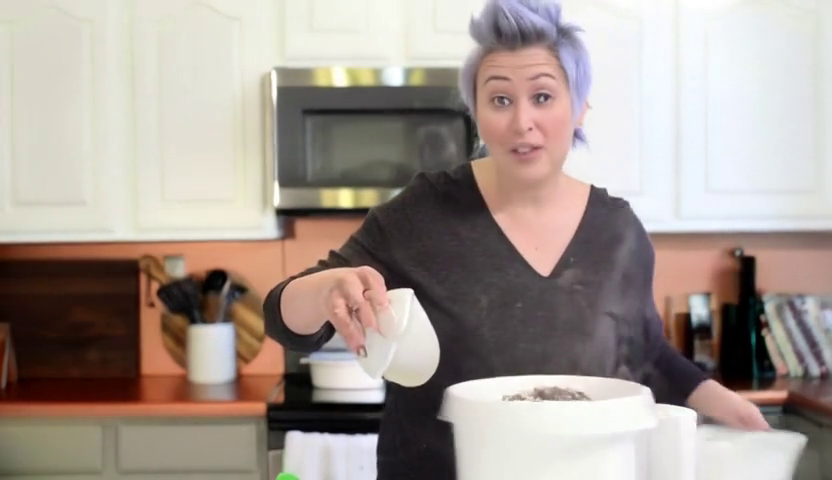} &
            \includegraphics[width=0.33\linewidth]{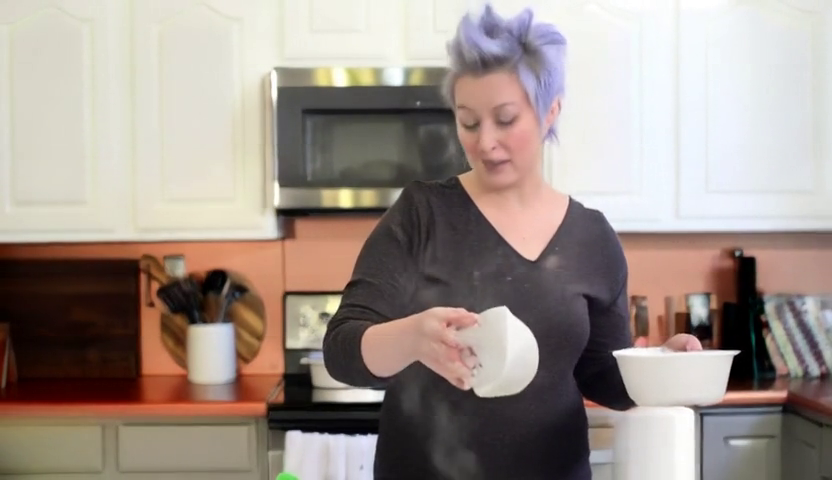} &
            \includegraphics[width=0.33\linewidth]{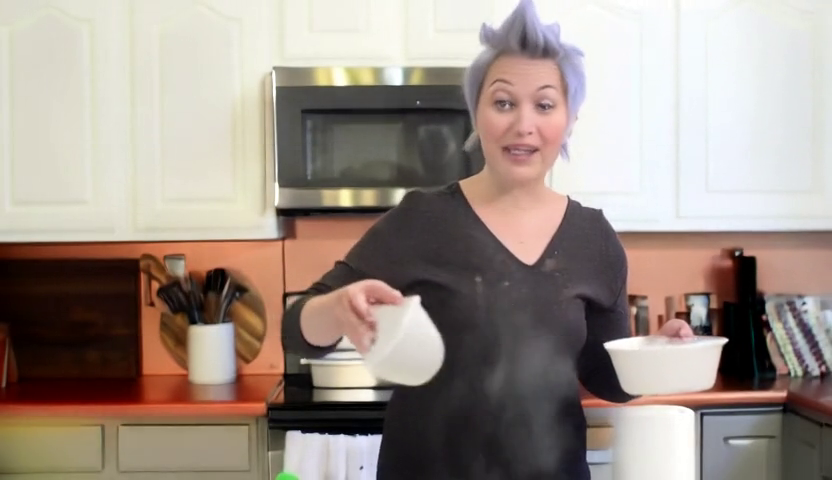} \\
        \end{tabular}
    \end{minipage}%
    \begin{minipage}[c]{0.23\linewidth}
        \centering
        \includegraphics[width=\linewidth, trim={17.7cm 0 0 0}, clip]{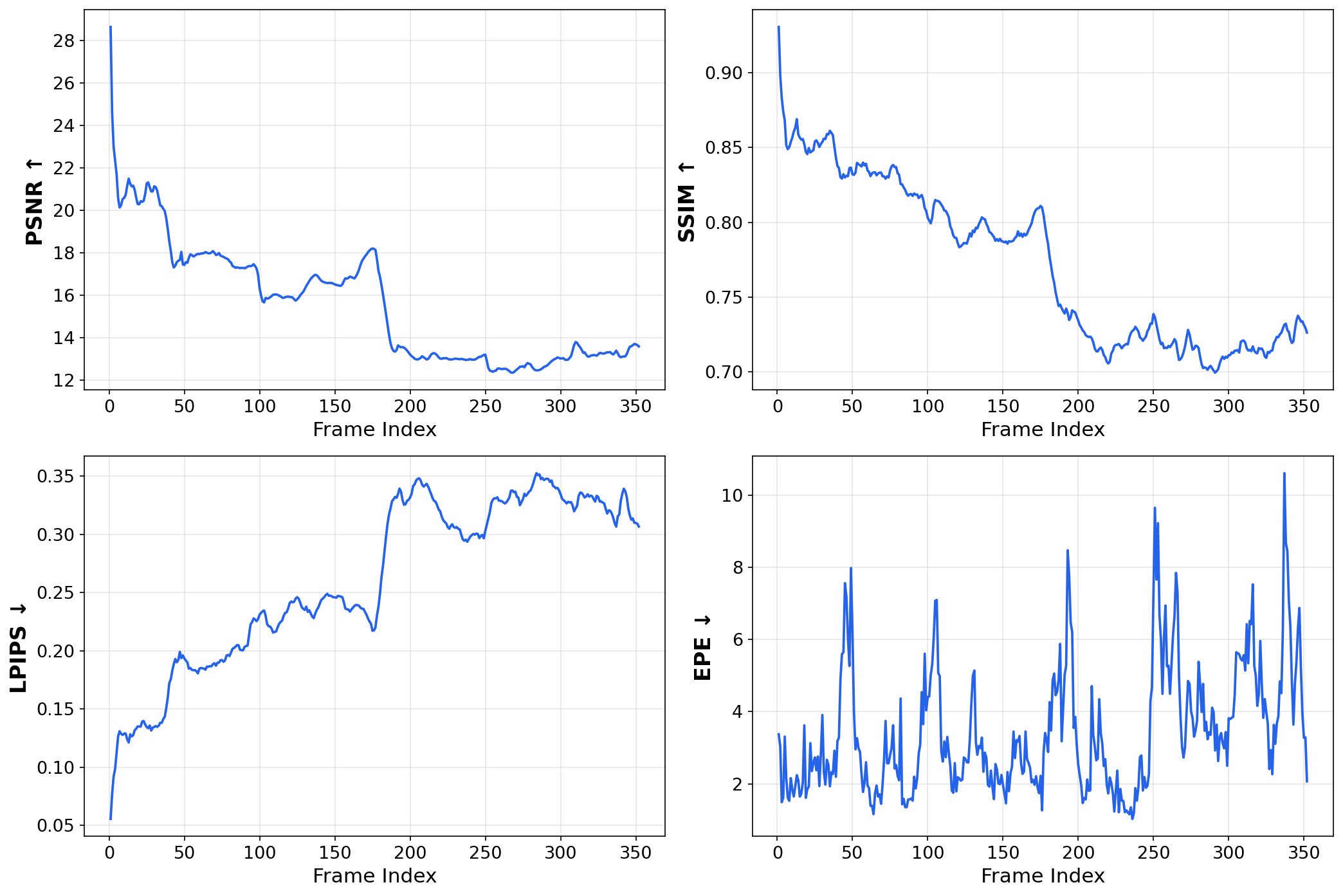}
    \end{minipage}
    \noindent\dashrule{\linewidth}{1pt}{3pt}{2pt}
    \begin{minipage}[c]{0.77\linewidth}
        \centering
        \begin{tabular}{@{}c@{}c@{}c@{}}
            \includegraphics[width=0.33\linewidth]{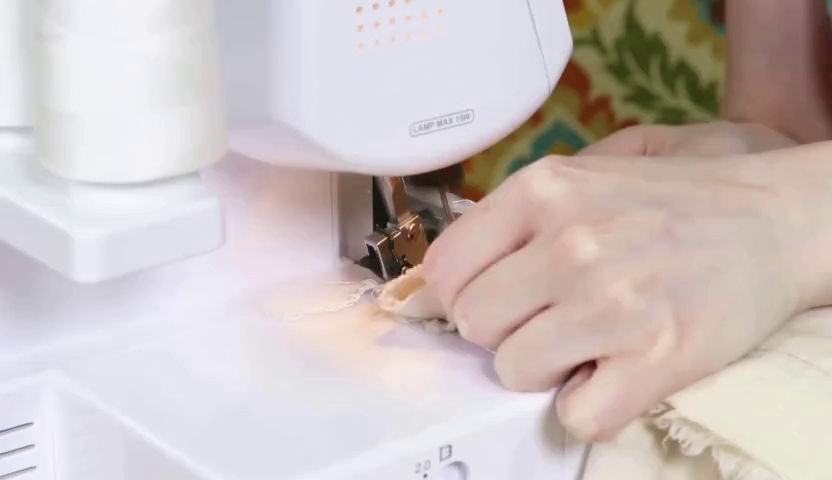} &
            \includegraphics[width=0.33\linewidth]{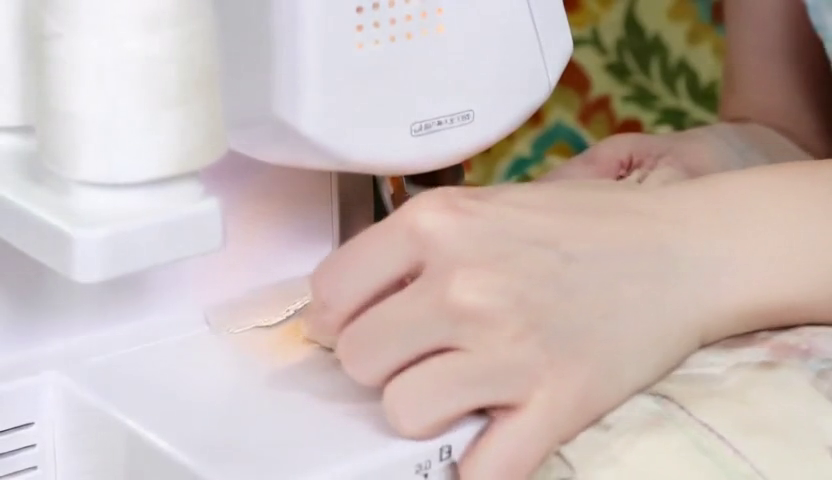} &
            \includegraphics[width=0.33\linewidth]{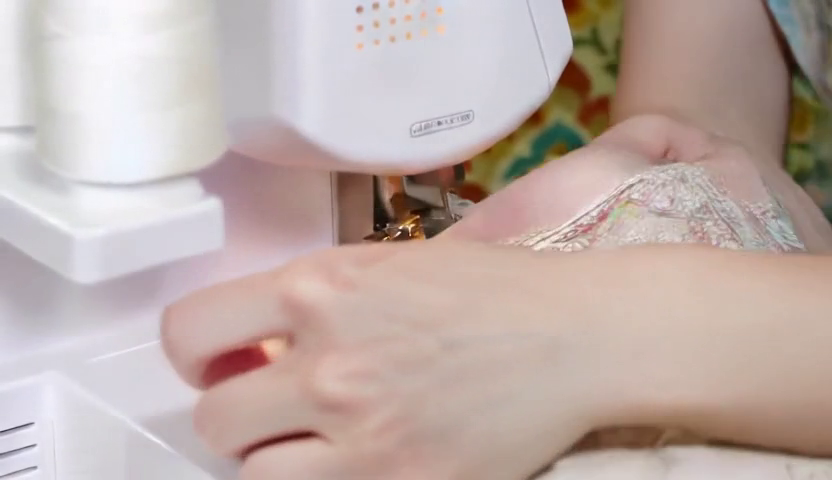} \\
            \includegraphics[width=0.33\linewidth]{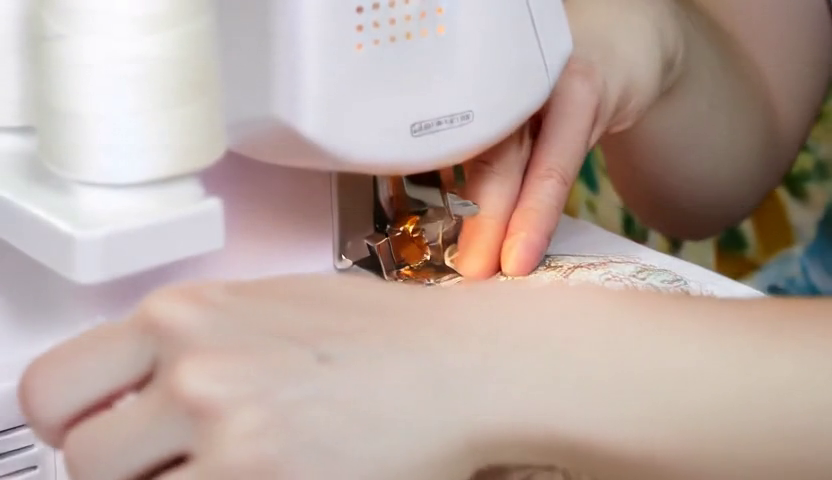} &
            \includegraphics[width=0.33\linewidth]{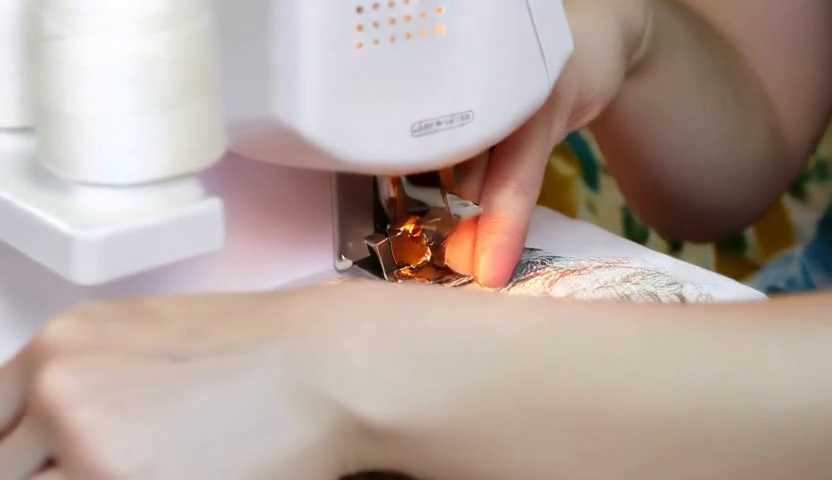} &
            \includegraphics[width=0.33\linewidth]{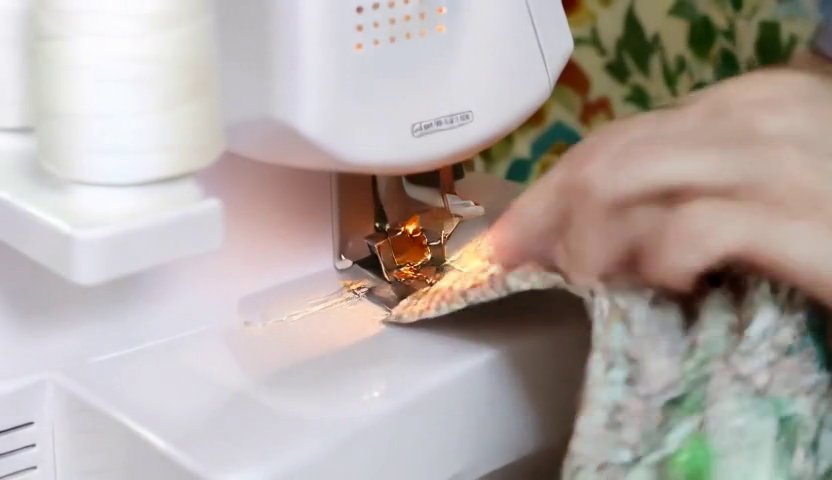} \\
        \end{tabular}
    \end{minipage}%
    \begin{minipage}[c]{0.23\linewidth}
        \centering
        \includegraphics[width=\linewidth, trim={17.7cm 0 0 0}, clip]{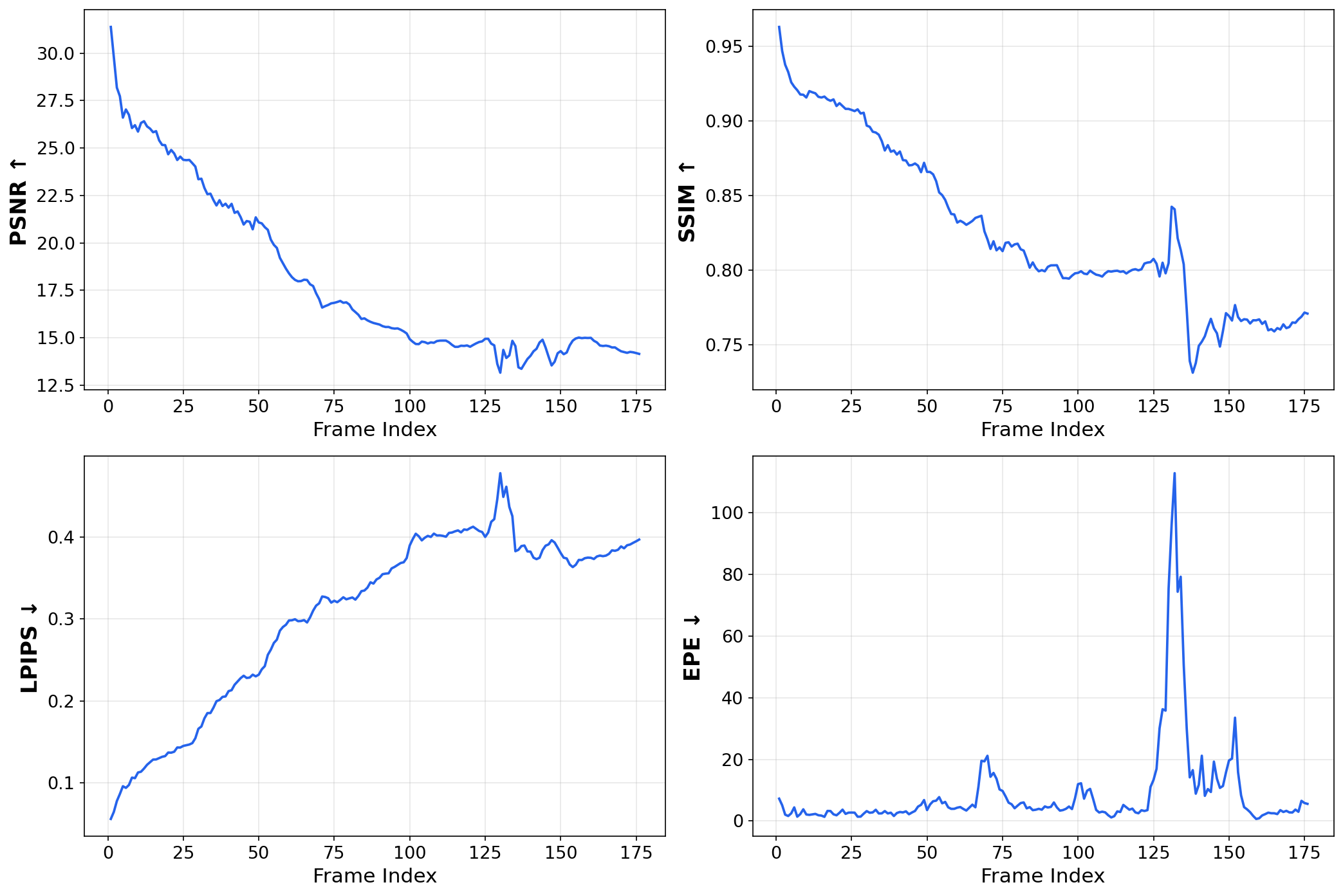}
    \end{minipage}
    \caption{\textbf{Long video generation.} Three sequences generated by our method, each shown as six uniformly sampled frames (left) with per-frame SSIM~$\uparrow$ and EPE~$\downarrow$ over the full sequence (right).}
    \label{fig:longgen}
\end{figure}

We demonstrate generation well beyond the training clip length. \cref{fig:longgen} shows three sequences spanning up to 350 frames: a seated woman gesturing at a table (top), a woman cooking in a kitchen (middle), and a close-up of hands performing delicate craft work (bottom).

Across all three, appearance stays plausible and motion temporally coherent for the full duration, with fine-grained detail such as facial features, hand movement, and object interaction remaining consistent into the later frames. Per-frame SSIM starts high (\eg, above 0.90) and declines gradually as the sequence progresses, which is expected given the growing temporal distance from the conditioning frame, yet stays above 0.75 throughout, indicating that visual fidelity is well preserved.

Track accuracy stays low for most frames, with occasional spikes at rapid motion events. The third sequence shows a pronounced EPE peak around frame 125, coinciding with a fast hand movement near the sewing machine. These spikes are transient and the metric recovers quickly, so abrupt motion does not cascade into persistent drift. Together these results confirm that our approach extends to several hundred frames while maintaining quality in both appearance and motion fidelity.

\subsection{With or Without Track Conditioning}\label{sec:supp:track_effect}

\begin{figure}[t!]
    \centering
    \setlength{\tabcolsep}{0pt}
    \renewcommand{\arraystretch}{0.8}
    \begin{tabular}{cccc}
        \multicolumn{4}{c}{\small\textbf{Rolling ball without track}} \\
        \includegraphics[width=0.25\linewidth]{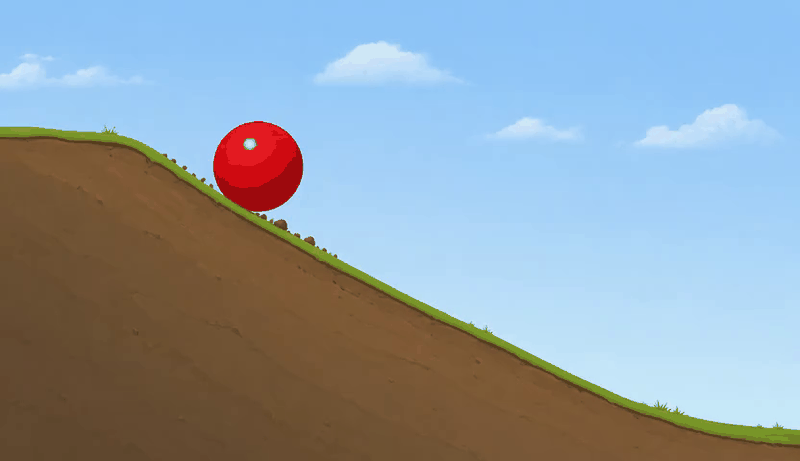} &
        \includegraphics[width=0.25\linewidth]{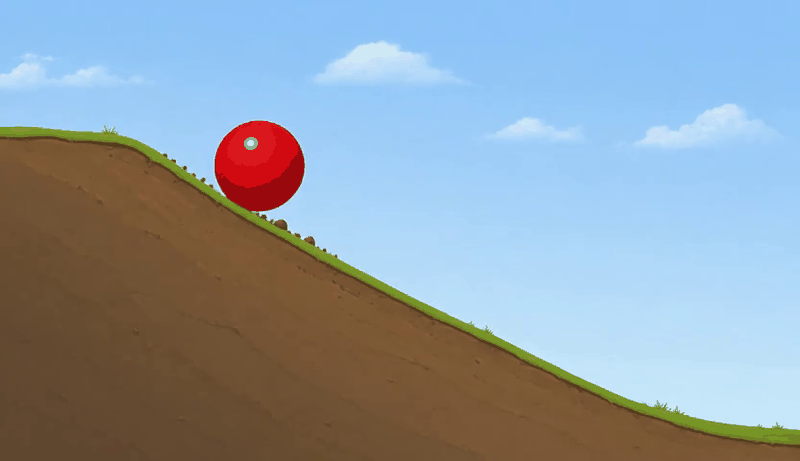} &
        \includegraphics[width=0.25\linewidth]{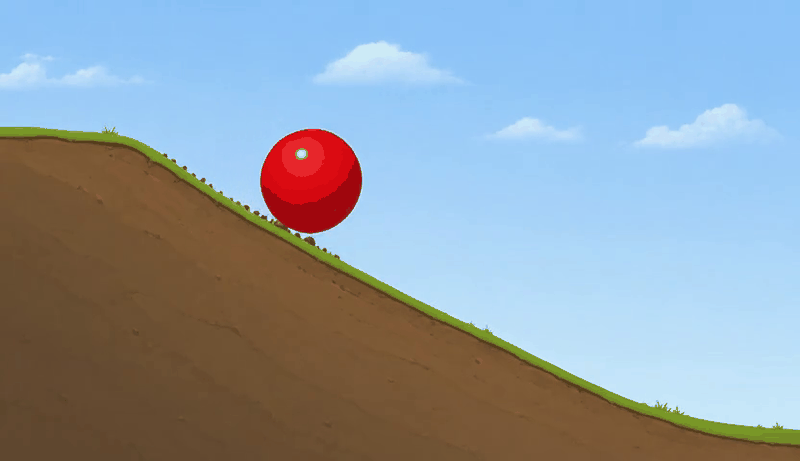} &
        \includegraphics[width=0.25\linewidth]{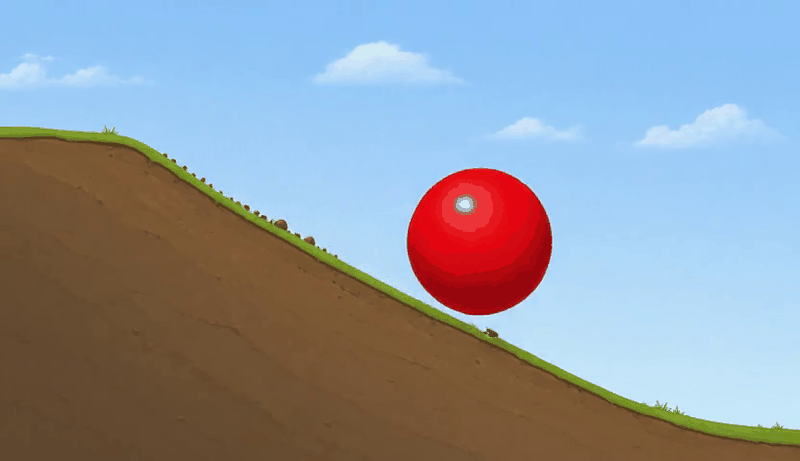} \\
        \multicolumn{4}{c}{\small\textbf{Rolling ball with track}} \\
        \includegraphics[width=0.25\linewidth]{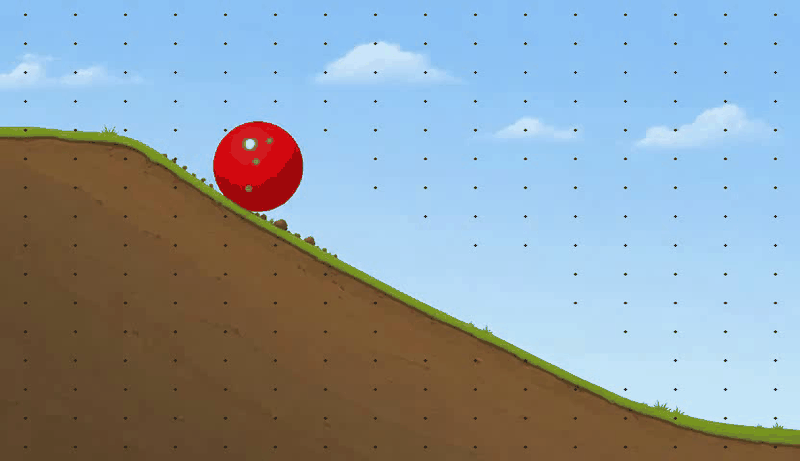} &
        \includegraphics[width=0.25\linewidth]{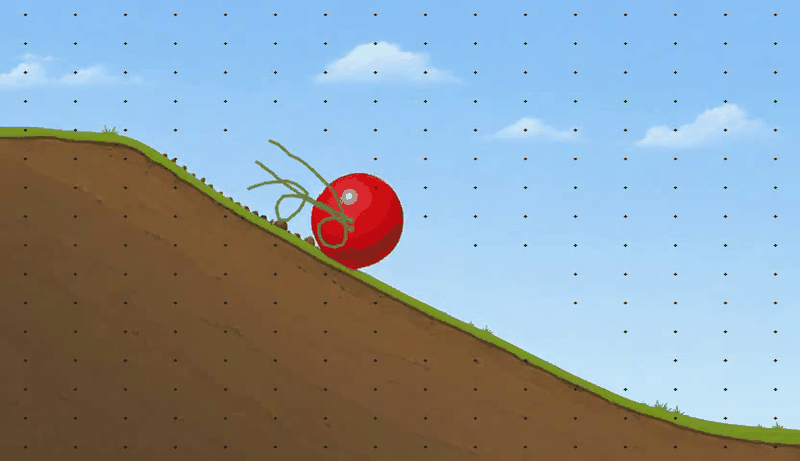} &
        \includegraphics[width=0.25\linewidth]{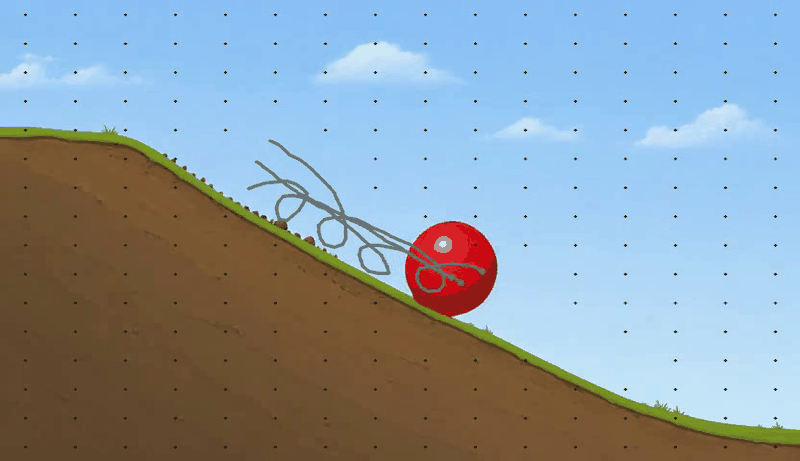} &
        \includegraphics[width=0.25\linewidth]{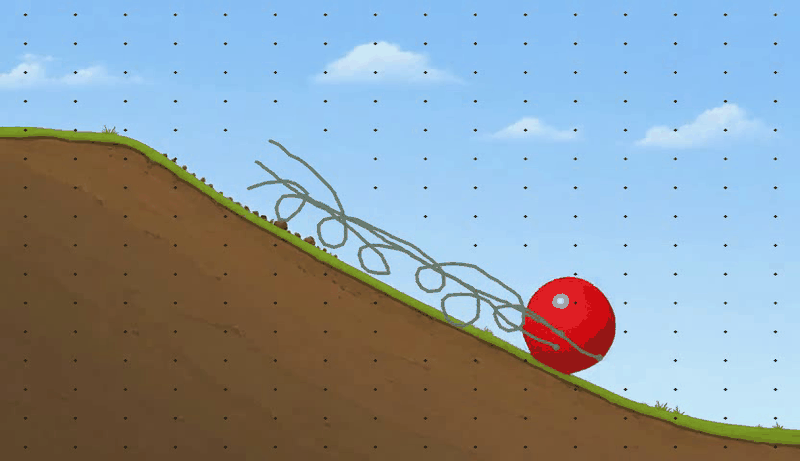} \\
        \multicolumn{4}{c}{\small\textbf{Push can without track}} \\
        \includegraphics[width=0.25\linewidth]{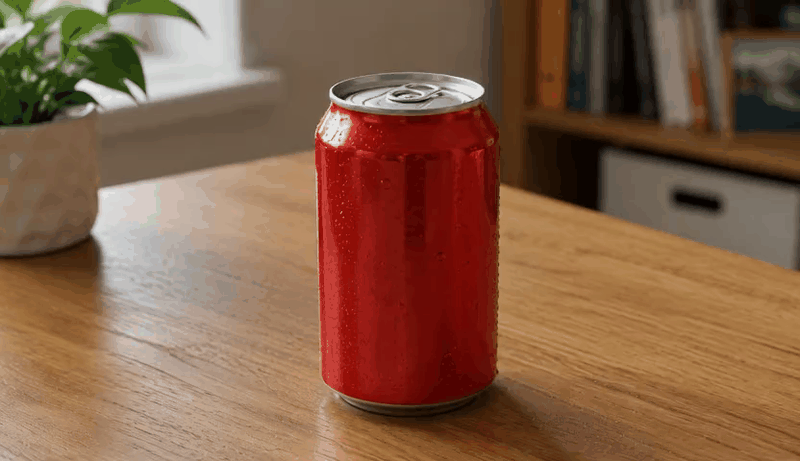} &
        \includegraphics[width=0.25\linewidth]{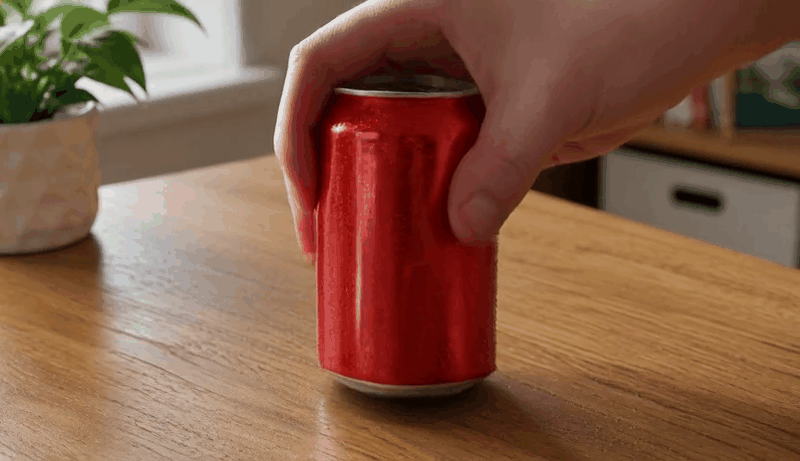} &
        \includegraphics[width=0.25\linewidth]{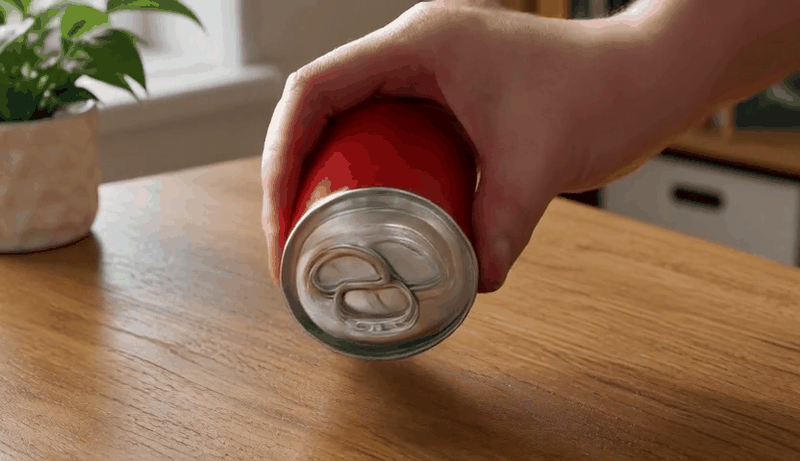} &
        \includegraphics[width=0.25\linewidth]{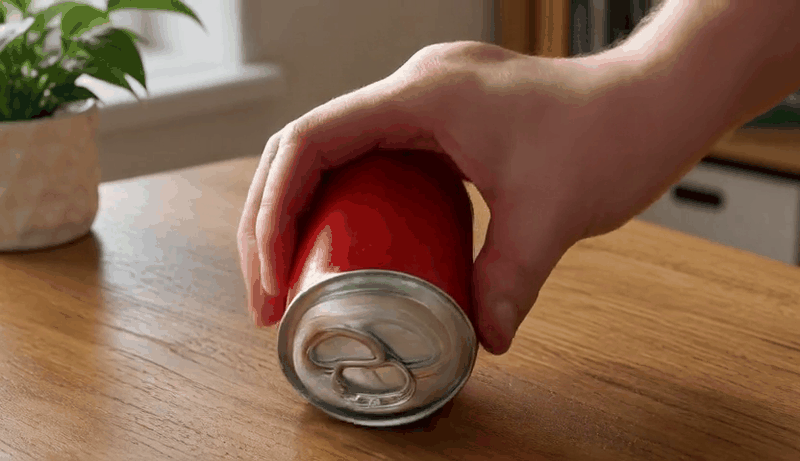} \\
        \multicolumn{4}{c}{\small\textbf{Push can with track}} \\
        \includegraphics[width=0.25\linewidth]{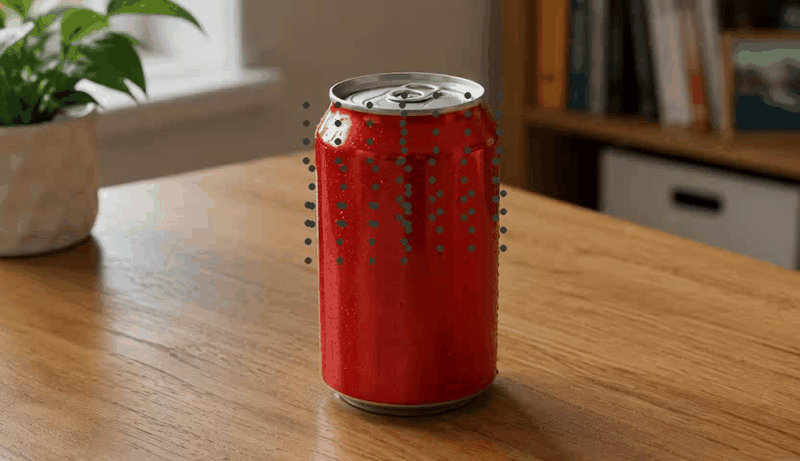} &
        \includegraphics[width=0.25\linewidth]{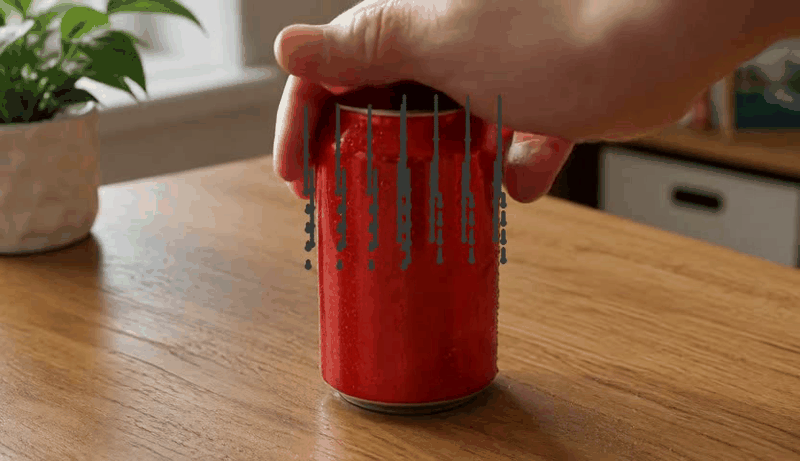} &
        \includegraphics[width=0.25\linewidth]{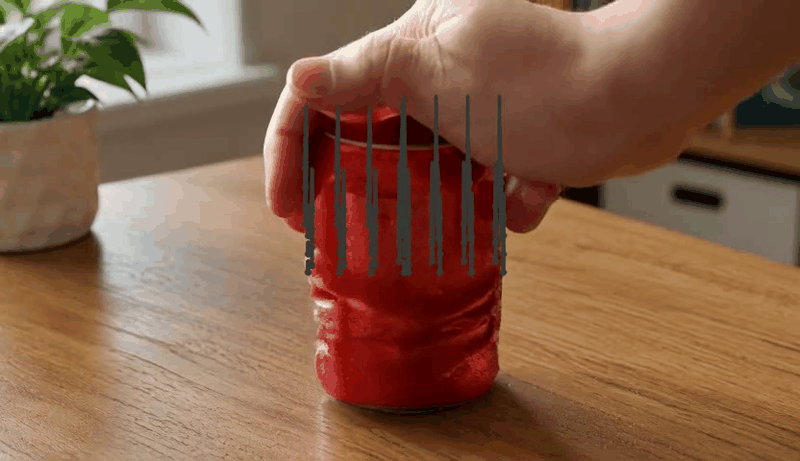} &
        \includegraphics[width=0.25\linewidth]{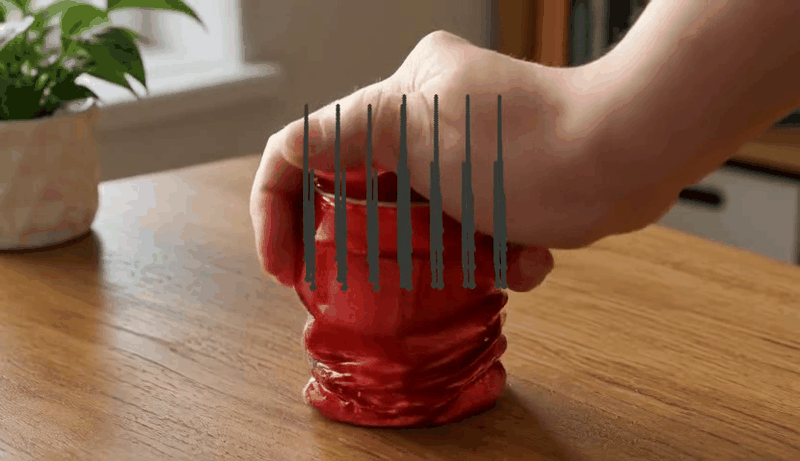} \\
    \end{tabular}
    \caption{\textbf{Effect of point track conditioning.} Without track guidance the rolling ball grows unnaturally over time and the pushed can rotates implausibly. Track conditioning enforces spatial consistency, preserving object scale and yielding more physically coherent interaction.}
    \label{fig:rolling_push}
\end{figure}

We qualitatively examine how 3D point track conditioning affects generation on two representative scenarios, shown in \cref{fig:rolling_push}.

\paragraph{Rolling ball.}
Without track conditioning (first group), the model produces plausible initial frames but progressively enlarges the ball, violating size constancy. With track conditioning (second group), the trajectories anchor the ball's spatial extent across frames, holding scale constant and producing a natural rolling motion.

\paragraph{Pushing a can.}
Without track guidance (third group), the pushed can undergoes implausible rotation and deformation as the sequence progresses, hallucinating motion incoherent with the initial contact. Conditioning on point tracks (fourth group) constrains the can to a physically plausible rigid-body trajectory, preserving shape and orientation while reflecting the applied force direction.

\paragraph{Discussion.}
Text prompts alone provide insufficient geometric constraint for fine-grained object dynamics. Point track conditioning acts as a spatial prior that fixes \emph{where} objects should be at each timestep, complementing the semantic guidance from text that specifies \emph{what} should happen. The combination resolves common failure modes of unconditioned generation, including object drift, scale inconsistency, and implausible deformation, without explicit physics simulation.

\subsection{Failure Cases}\label{sec:supp:failure}

\begin{figure}[t!]
    \centering
    \setlength{\tabcolsep}{0pt}
    \renewcommand{\arraystretch}{0.8}
    \begin{tabular}{cccc}
        \multicolumn{4}{c}{\small\textbf{Visual blur}} \\
        \includegraphics[width=0.25\linewidth]{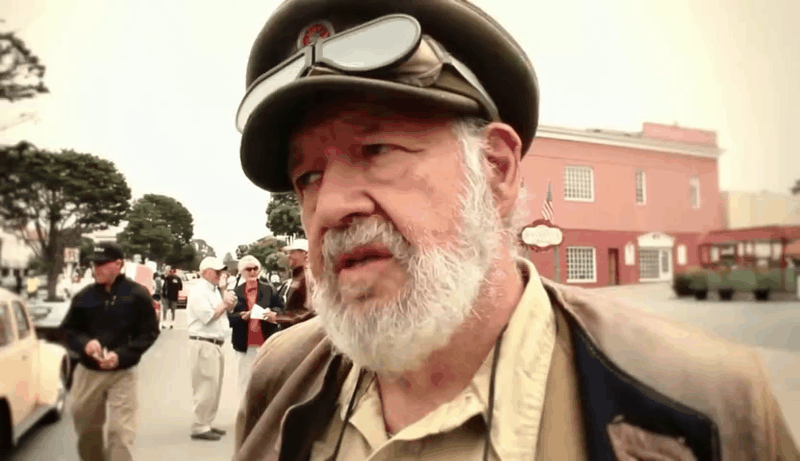} &
        \includegraphics[width=0.25\linewidth]{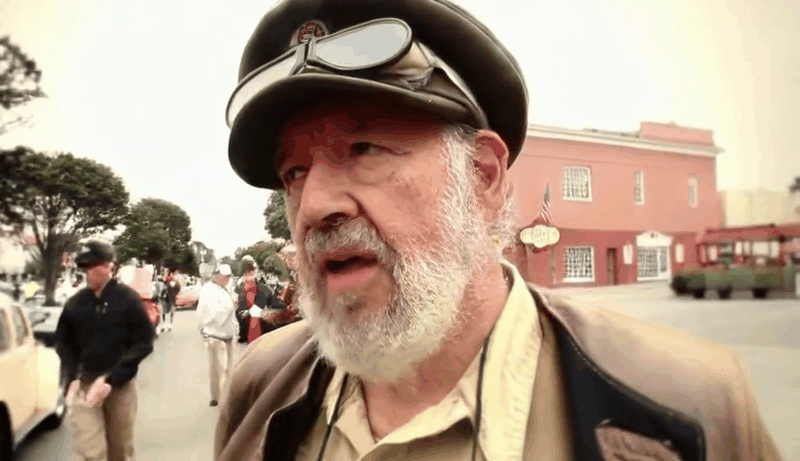} &
        \includegraphics[width=0.25\linewidth]{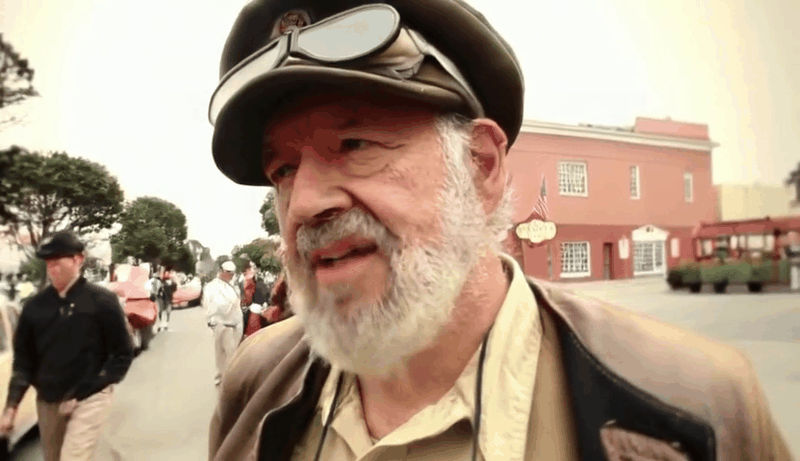} &
        \includegraphics[width=0.25\linewidth]{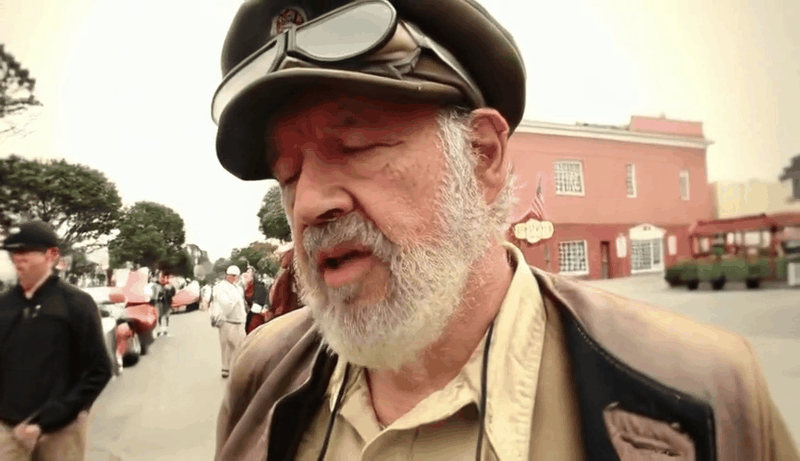} \\
        \multicolumn{4}{c}{\small\textbf{Broken causality}} \\
        \includegraphics[width=0.25\linewidth]{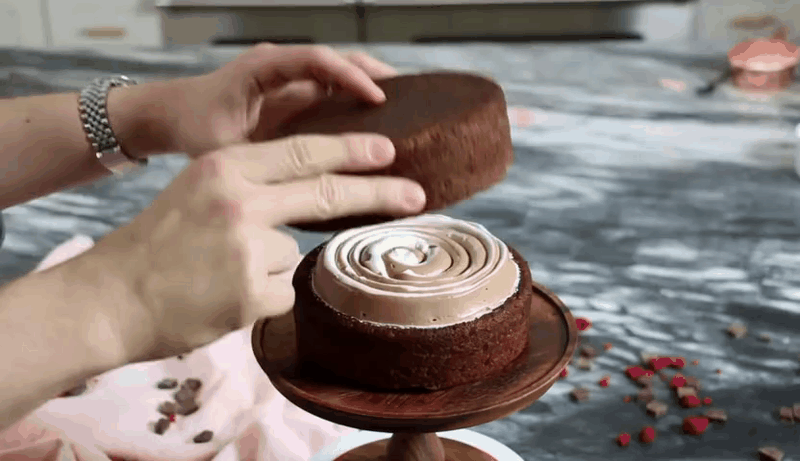} &
        \includegraphics[width=0.25\linewidth]{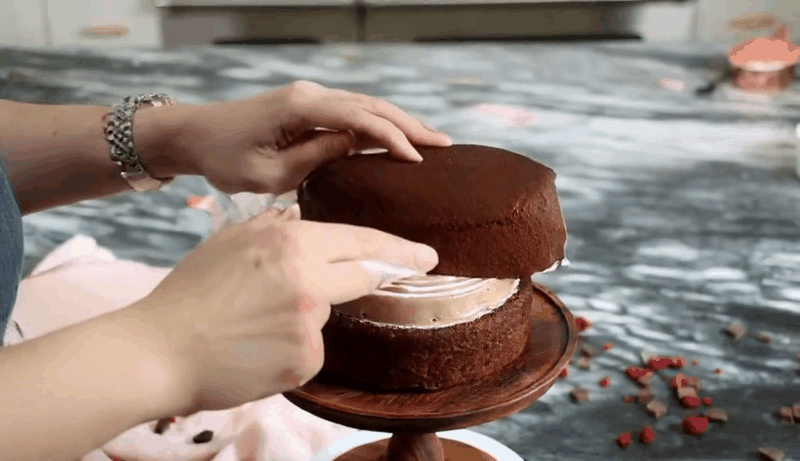} &
        \includegraphics[width=0.25\linewidth]{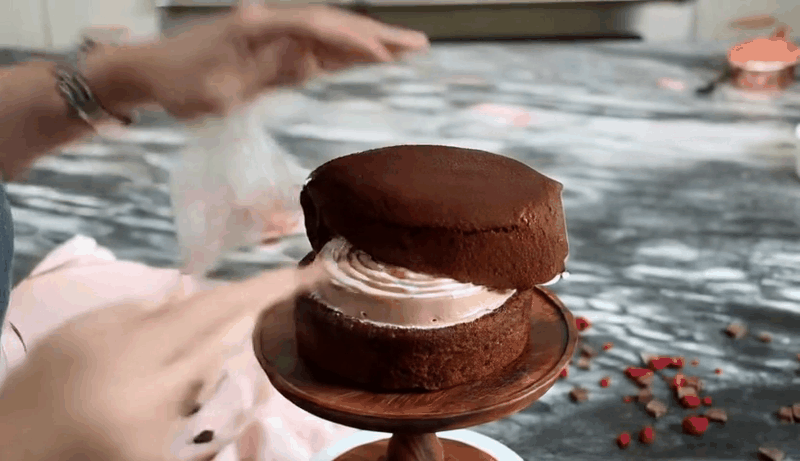} &
        \includegraphics[width=0.25\linewidth]{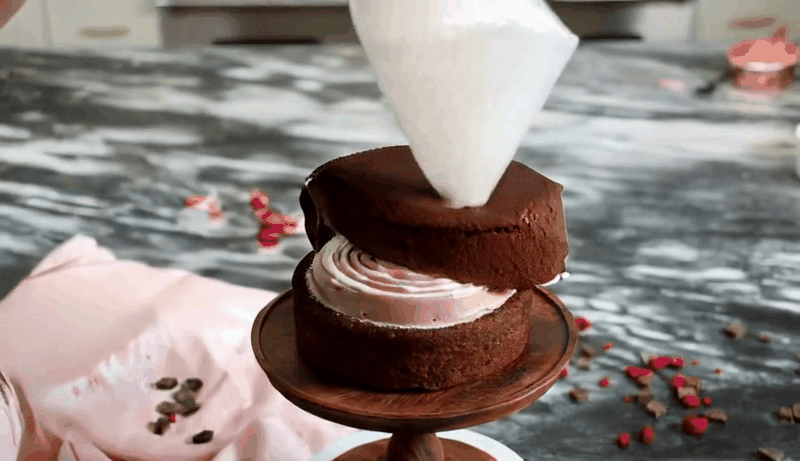} \\
    \end{tabular}
    \caption{\textbf{Failure cases.} Top: small faces and background objects blur progressively as the video extends. Bottom: causally inconsistent interaction, where a cake develops an unexplained missing section after being topped and a piping bag materializes from thin air.}
    \label{fig:failure}
\end{figure}

We identify two representative failure modes. First, when subjects or objects occupy a small region of the frame, results tend to blur, losing fine detail such as facial features and object texture (\cref{fig:failure}, top). This likely follows from the limited spatial resolution allocated to small regions, which makes high-frequency detail hard to sustain over time. Second, our method can produce sequences that violate causality and object permanence (\cref{fig:failure}, bottom): after a topping is placed on a cake, the cake inexplicably develops a missing chunk, and a piping bag appears from nowhere in later frames. These artifacts suggest the model lacks a coherent account of causal object interaction and struggles to hold object state consistent across frames. Incorporating structured physical or causal reasoning is a promising direction for future work.

\section{Training Details}

\subsection{Teacher Training}\label{sec:supp:teacher}

\paragraph{Base model and fine-tuning strategy.}
The teacher is the Wan2.2 text-and-image-to-video DiT with 5B parameters \citep{wan2025wan}, operating on a latent space from a causal 3D VAE with spatial compression $16\times$ and temporal compression $4\times$. It takes 48-channel video latents and generates 81-frame clips at $480\times832$. We freeze the pretrained DiT and inject LoRA adapters \citep{hu2022lora} of rank $r{=}32$ into the query, key, value, and output projections (\texttt{q, k, v, o}) and both FFN linear layers (\texttt{ffn.0, ffn.2}). Alongside the LoRA parameters we jointly train:
\begin{itemize}[nosep,leftmargin=*]
    \item the motion track encoder;
    \item the expanded patch embedding, whose input channels grow from $C_{\mathrm{lat}}{=}48$ to $C_{\mathrm{lat}}+C_{\mathrm{trk}}+C_{\mathrm{dep}}=48+64+16=128$ to accept the concatenated motion features.
\end{itemize}
The original patch embedding weights are preserved and the new channels zero-filled, giving a stable initialization from pretrained knowledge. The conditioning tensor $\mathbf{F}^{\mathrm{cond}}=[\mathbf{F}^{\mathrm{trk}};\,\mathbf{F}^{\mathrm{dep}}]\in\mathbb{R}^{80\times F\times H_\ell\times W_\ell}$ is concatenated with the noisy latents along the channel dimension before patch embedding, as defined in \cref{sec:model}.

\paragraph{3D track data preparation.}
We extract per-video 3D tracks with SpatialTrackerV2 \citep{xiao2025spatialtrackerv2} and store them in a single HDF5 file with per-video groups, each containing:
\begin{itemize}[nosep,leftmargin=*]
    \item \texttt{tracks}: $[N, T, 3]$ tensor of $(x, y, z)$ in camera space;
    \item \texttt{intrinsics}: $[T, 3, 3]$ camera intrinsic matrices;
    \item \texttt{visibility}: $[N, T]$ binary occlusion mask, not used in training.
\end{itemize}
At load time, tracks are projected to image coordinates with the intrinsics, giving normalized positions in $[0,1]^2$. Depth is converted to inverse depth and min--max normalized to $[0,1]$ per sample. Each training sample draws a fixed $N{=}256$ tracks by random permutation of the tracked points in that video.

\paragraph{Training objective.}
We adopt the rectified flow-matching objective of Wan2.2. Given clean latents $\mathbf{x}_0$ encoded from the training video, Gaussian noise $\boldsymbol{\epsilon}\sim\mathcal{N}(\mathbf{0},\mathbf{I})$, and a flow time $\tau\sim\mathcal{U}(0,1000)$ warped by the shifted-sigmoid schedule (shift $s{=}5$, $\sigma(\tau)=s\tau/(1+(s-1)\tau)$), the noisy latent is
\begin{equation}
    \mathbf{x}_\tau = (1 - \sigma_\tau)\,\mathbf{x}_0 + \sigma_\tau\,\boldsymbol{\epsilon}.
\end{equation}
The velocity target is $\mathbf{v}=\boldsymbol{\epsilon}-\mathbf{x}_0$, and we minimize
\begin{equation}
    \mathcal{L} = \bigl\|\mathbf{v}_\theta(\mathbf{x}_\tau, \tau, \mathbf{c}, \mathbf{F}^{\mathrm{cond}}) - \mathbf{v}\bigr\|_2^2,
\end{equation}
where $\mathbf{c}$ denotes text and image conditioning. The first-frame latent is injected as ground truth rather than noised, and the loss is evaluated only on the remaining latent frames.

\paragraph{Training configuration.}
\cref{tab:teacher_config} summarizes the key hyperparameters. Training uses the Accelerate framework with DDP across multiple nodes, and gradient checkpointing throughout the DiT backbone to fit 81-frame clips within per-GPU memory. Trainable parameters comprise LoRA adapters in all 40 transformer blocks (about 200M at rank 32), the motion track encoder (about 0.3M), the depth branch (about 0.1M), and the expanded patch embedding (about 2.6M for the 80 new input channels).

\begin{table}[t!]
    \centering
    \small
    \caption{\textbf{Teacher model training configuration.}}
    \label{tab:teacher_config}
    \begin{tabular}{ll}
        \toprule
        \textbf{Hyperparameter} & \textbf{Value} \\
        \midrule
        Base model & Wan2.2-TI2V-5B \\
        Resolution (stage 1) & $256 \times 256$, 49 frames \\
        Resolution (stage 2) & $480 \times 832$, 81 frames (resumed from stage 1) \\
        LoRA rank & 32 \\
        LoRA target modules & \texttt{q, k, v, o, ffn.0, ffn.2} \\
        Learning rate (stage 1) & $1 \times 10^{-4}$ \\
        Learning rate (stage 2) & $5 \times 10^{-5}$ \\
        Epochs & 3 \\
        GPUs & 80 (multi-node, NCCL) \\
        Precision & BF16 (mixed precision) \\
        Gradient checkpointing & Enabled \\
        \midrule
        Track embed dim $D$ & 64 \\
        Track head output channels $C_{\mathrm{trk}}$ & 64 \\
        Depth embed dim $D_d$ & 16 \\
        Depth head output channels $C_{\mathrm{dep}}$ & 16 \\
        Tracks per sample $N$ & 256 (fixed) \\
        Max track ID & 256 \\
        \midrule
        Dataset & \dataset \\
        Data workers & 8 per process \\
        Distributed framework & Accelerate (DDP) \\
        \bottomrule
    \end{tabular}
\end{table}

\subsection{Streaming Distillation}\label{sec:supp:distill}

We distill the bidirectional teacher into a causal 4-step student following the self-forcing and DMD paradigm \citep{huang2025selfforcing,yin2024dmd,yin2025causvid}, in three sequential stages: ODE pair generation, ODE initialization, and adversarial DMD training.

\paragraph{Stage 0: ODE pair generation.}
We run the frozen teacher for 48 denoising steps (shifted-sigmoid schedule, shift $s{=}5$, $\sigma_{\min}{=}0$) with classifier-free guidance at scale 1 on each training video to produce full ODE trajectories. We use 48 rather than the default 50 steps so the trajectory divides evenly into the student's 4-step schedule, giving clean intermediate snapshots at uniform intervals. The teacher receives the first-frame latent, held fixed at frame 0 throughout, together with 3D motion track conditioning. We save 5 evenly spaced snapshots per trajectory (steps 0, 12, 24, 36, and the final clean latent), yielding tensors of shape $[5, F, C_{\mathrm{in}}, H_\ell, W_\ell]$ with $F{=}21$, $C_{\mathrm{in}}{=}48{+}80$, $H_\ell{=}30$, and $W_\ell{=}52$. Snapshots are aggregated into an LMDB database for fast random access during training.

\paragraph{Stage 1: ODE initialization.}
We initialize the causal student by supervising it on the pre-computed ODE pairs. The student shares the teacher's architecture but replaces bidirectional self-attention with \textbf{block-wise causal attention}: latent frames are grouped into chunks of $c{=}4$ frames, and when denoising chunk $i$ the model attends only to its own tokens and a causally accumulated KV cache from preceding chunks. At each step we sample an intermediate denoising index per block, extract the corresponding noisy latent from the saved trajectory, and regress the clean latent,
\begin{equation}
    \mathcal{L}_{\mathrm{ode}} = \bigl\|\hat{\mathbf{x}}_{0,\theta}(\mathbf{x}_{\tau_k}, \tau_k, \mathbf{c}) - \mathbf{x}_0\bigr\|_2^2,
\end{equation}
with $\tau_k\in\{1000, 750, 500, 250\}$ after warping by the shifted schedule. The first-frame latent is injected as ground truth and excluded from the loss. \cref{tab:ode_config} lists the hyperparameters.

\begin{table}[t!]
    \centering
    \small
    \caption{\textbf{ODE initialization training configuration.}}
    \label{tab:ode_config}
    \begin{tabular}{ll}
        \toprule
        \textbf{Hyperparameter} & \textbf{Value} \\
        \midrule
        Student architecture & Wan2.2-TI2V-5B (causal, block-wise attention) \\
        Frames per chunk $c$ & 4 \\
        Denoising steps (warped) & $\{1000, 750, 500, 250\}$ \\
        Independent first frame & Yes \\
        \midrule
        Learning rate & $2 \times 10^{-5}$ \\
        Optimizer & AdamW ($\beta_1{=}0.9$, $\beta_2{=}0.999$, wd${=}0.01$) \\
        Global batch size & 64 \\
        Per-GPU batch size & 1 \\
        Precision & BF16 (mixed) \\
        Gradient checkpointing & Enabled \\
        FSDP strategy & Hybrid full shard \\
        \midrule
        Dataset & Pre-computed ODE pairs (LMDB) \\
        Motion track channels & $C_{\mathrm{trk}}{=}64$, $C_{\mathrm{dep}}{=}16$ \\
        Tracks per sample & 256 \\
        \bottomrule
    \end{tabular}
\end{table}

\paragraph{Stage 2: DMD adversarial distillation.}
We then switch to distribution matching distillation \citep{yin2024dmd} with temporal self-rollout, using the objective given in \cref{sec:model}. Training involves three models:
\begin{itemize}[nosep,leftmargin=*]
    \item \textbf{Generator} $G_\theta$, the causal student, initialized from the ODE stage checkpoint.
    \item \textbf{Real score} $s_{\mathrm{real}}$, the frozen LoRA-finetuned bidirectional teacher loaded via DiffSynth with motion track conditioning.
    \item \textbf{Fake score} $s_{\mathrm{fake}}$, a learnable critic initialized from the same teacher weights and trained online.
\end{itemize}
At each generator step the student produces a full clip by temporal self-rollout: every chunk of $c{=}4$ frames is denoised through the 4-step schedule $\{1000, 750, 500, 250\}$ conditioned on its own previously generated latents via KV cache, matching streaming deployment, with the same denoising step used across all chunks within an iteration. The real score applies classifier-free guidance at scale 1 while the fake score uses the conditional prediction alone. The critic is updated at every step and the generator once every 5 critic steps, with an exponential moving average of the generator weights at decay 0.99 from step 200 onward. \cref{tab:dmd_config} lists the full configuration.

\begin{table}[t!]
    \centering
    \small
    \caption{\textbf{DMD distillation training configuration.}}
    \label{tab:dmd_config}
    \begin{tabular}{ll}
        \toprule
        \textbf{Hyperparameter} & \textbf{Value} \\
        \midrule
        Generator init & ODE stage checkpoint \\
        Real score (teacher) & Frozen LoRA-finetuned Wan2.2-TI2V-5B \\
        Fake score (critic) init & Same teacher weights (trainable) \\
        LoRA alpha & 1.0 \\
        \midrule
        CFG scale (teacher) & 1.0 \\
        Denoising steps & $\{1000, 750, 500, 250\}$ (4 steps) \\
        Frames per chunk $c$ & 4 \\
        Timestep shift $s$ & 5.0 \\
        Timestep range & $[20, 980]$ \\
        Context noise & 0 (clean KV cache) \\
        \midrule
        Generator lr & $2 \times 10^{-6}$ \\
        Critic lr & $4 \times 10^{-6}$ \\
        Optimizer & AdamW ($\beta_1{=}0$, $\beta_2{=}0.999$) \\
        Global batch size & 64 \\
        Per-GPU batch size & 1 \\
        Critic-to-generator update ratio $R$ & 5 \\
        EMA decay & 0.99 (start step 200) \\
        Max gradient norm & 10.0 (both) \\
        \midrule
        Precision & BF16 (mixed) \\
        Gradient checkpointing & Enabled \\
        FSDP strategy & Hybrid full shard \\
        \midrule
        Dataset & \dataset \\
        Resolution & $480 \times 832$, 81 frames \\
        Latent shape & $[1, 21, 48, 30, 52]$ \\
        Tracks per sample & 256 \\
        Denoising loss type & Flow matching \\
        \bottomrule
    \end{tabular}
\end{table}

\paragraph{Distributed training.}
Both stages use multi-node training with fully sharded data parallelism (FSDP, hybrid full-shard). The generator, real score, fake score, and text encoder are independently FSDP-wrapped with size-based auto-wrapping, while the VAE remains unsharded on each device for online latent encoding during DMD training. Distillation runs on 64 high-end GPUs.

\paragraph{Online 3D control at inference.}
Because the motion encoder is temporally separable (\cref{sec:model}), switching between offline full-sequence and online per-chunk conditioning requires no architectural change. Generation proceeds chunk by chunk: once a chunk is committed and decoded, the encoder recomputes $\mathbf{F}^{\mathrm{cond}}$ from the latest 3D tracks and depth over the upcoming segment. Users may revise trajectories, supply new tracker outputs, or adjust depth at any time, since only the current control window is required. This closes the loop between interactive 4D control and real-time video output at a memory cost independent of length.

\section{More Ablations}

\subsection{Ablations on Distillation}\label{sec:supp:ablation_distill}

\begin{figure}[t!]
    \centering
    \includegraphics[width=0.8\linewidth]{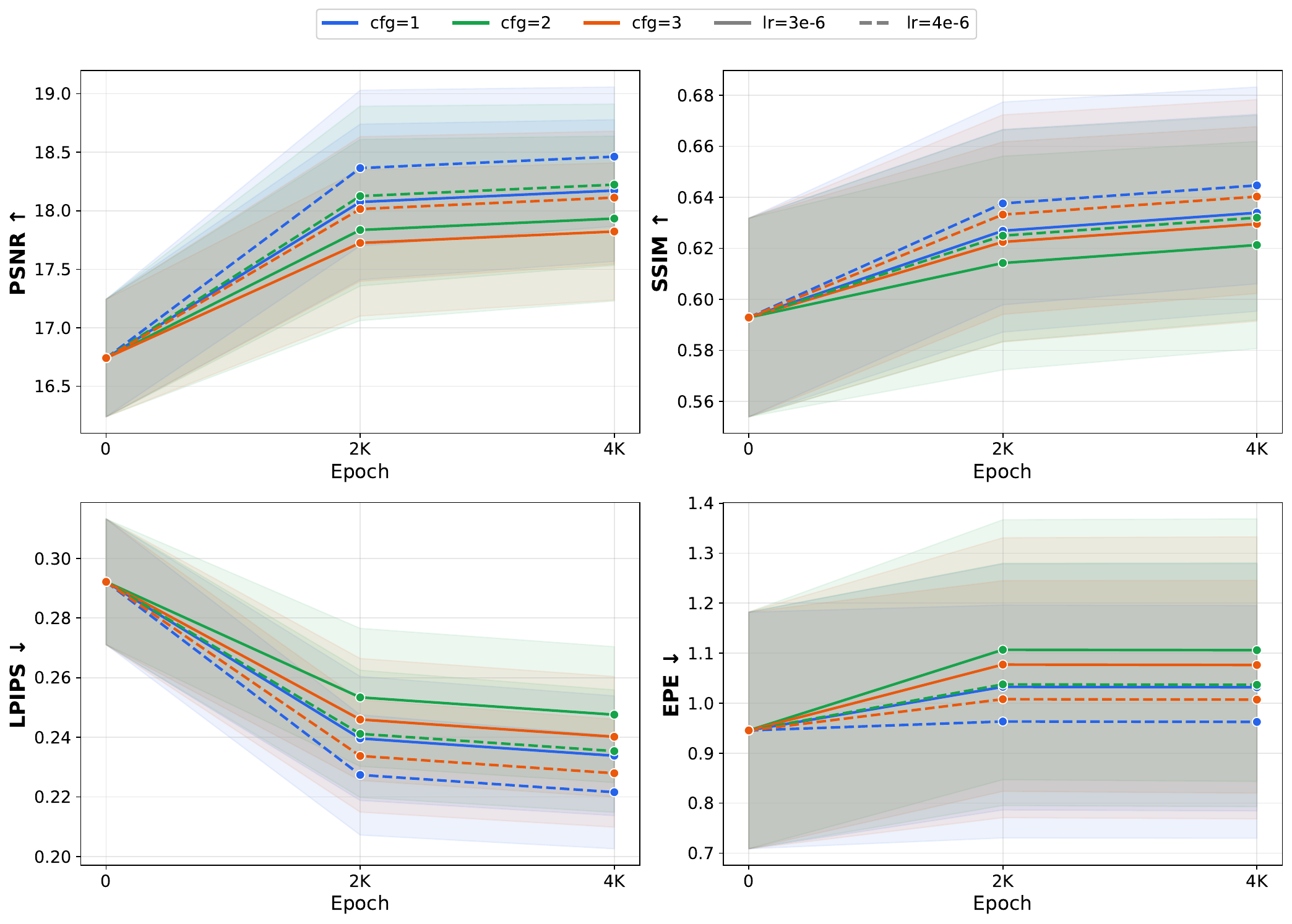}
    \caption{\textbf{Distillation ablation.} PSNR, SSIM, LPIPS, and EPE over 4K distillation steps for three guidance scales and two learning rates. Solid curves are lr\,=\,3e-6 and dashed curves lr\,=\,4e-6.}
    \label{fig:ablation_distill}
\end{figure}

We ablate two distillation hyperparameters: the classifier-free guidance scale (cfg $\in\{1,2,3\}$) and the learning rate (lr $\in\{3\text{e-}6, 4\text{e-}6\}$), tracking PSNR~$\uparrow$, SSIM~$\uparrow$, LPIPS~$\downarrow$, and EPE~$\downarrow$ over training up to 4K steps (\cref{fig:ablation_distill}).

All configurations improve steadily on the appearance metrics as training progresses, confirming that distillation works as intended. Among the guidance scales, cfg\,=\,1 is best on all four metrics, reaching the highest PSNR (about 18.5) and SSIM (about 0.65) and the lowest LPIPS (about 0.22) and EPE (about 0.95) at 4K steps. Raising guidance to 2 or 3 degrades results noticeably, suggesting that stronger guidance introduces artifacts during distillation that harm both visual quality and motion accuracy.

For the learning rate, lr\,=\,4e-6 (dashed) consistently beats lr\,=\,3e-6 (solid) across every guidance scale and metric, indicating that a moderately higher rate accelerates convergence without instability. The margin is widest on PSNR and LPIPS, where the dashed curves stay clearly ahead throughout.

EPE behaves differently from the appearance metrics: cfg\,=\,1 keeps it low and stable, whereas cfg\,=\,2 and cfg\,=\,3 drift slightly upward, so higher guidance appears to introduce temporal inconsistencies that accumulate over longer training. We therefore adopt cfg\,=\,1 with lr\,=\,4e-6 as the default.

\subsection{Effect of Random Seeds}\label{sec:supp:seeds}

\begin{figure}[t!]
    \centering
    \setlength{\tabcolsep}{0pt}
    \renewcommand{\arraystretch}{0.8}
    \begin{tabular}{ccccc}
        \small Frame 0 & \small Frame 20 & \small Frame 40 & \small Frame 60 & \small Frame 80 \\
        \multicolumn{5}{c}{\small\textbf{Seed 0}} \\
        \includegraphics[width=0.2\linewidth]{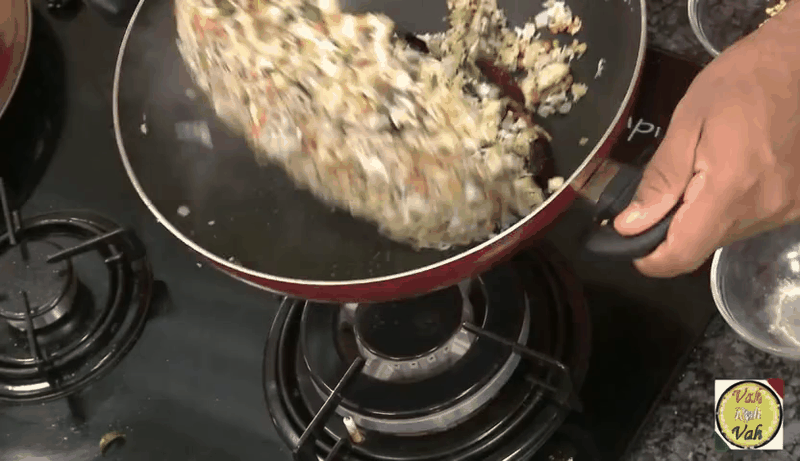} &
        \includegraphics[width=0.2\linewidth]{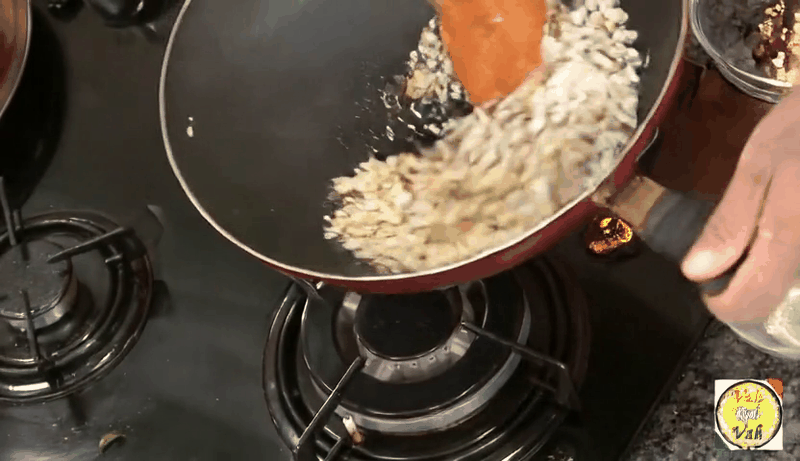} &
        \includegraphics[width=0.2\linewidth]{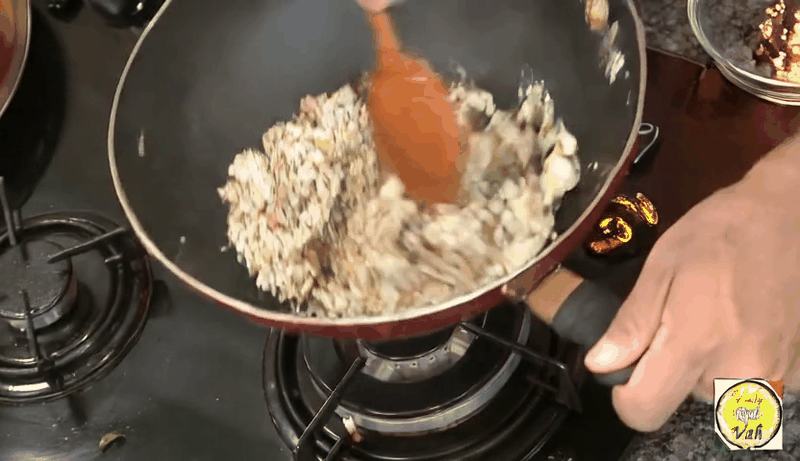} &
        \includegraphics[width=0.2\linewidth]{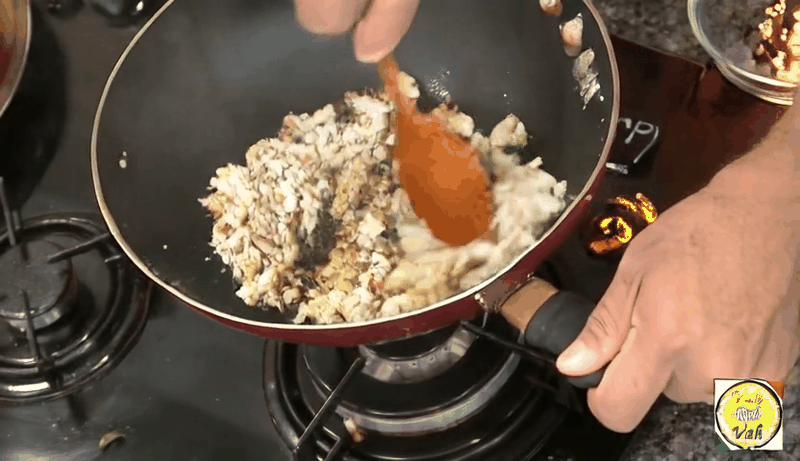} &
        \includegraphics[width=0.2\linewidth]{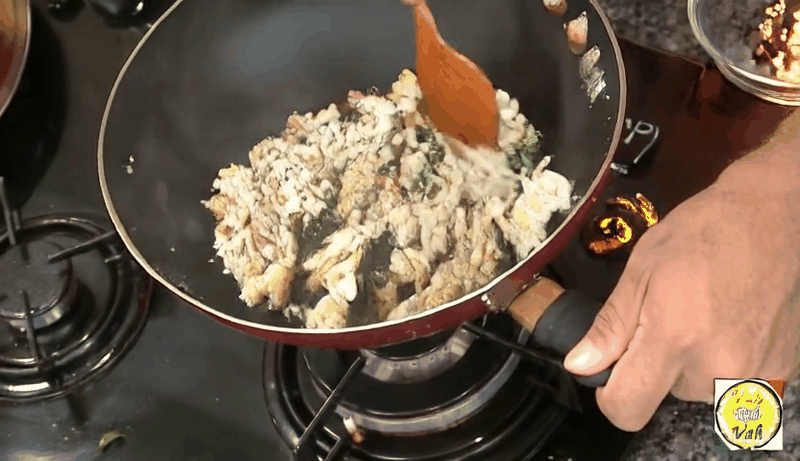} \\
        \multicolumn{5}{c}{\small\textbf{Seed 1}} \\
        \includegraphics[width=0.2\linewidth]{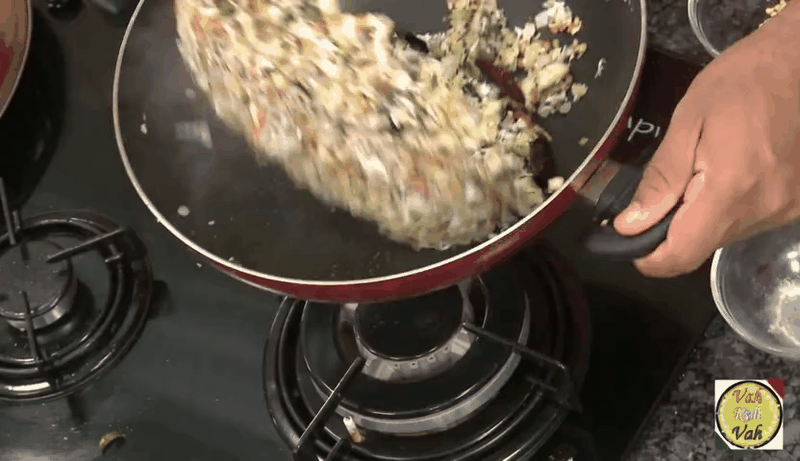} &
        \includegraphics[width=0.2\linewidth]{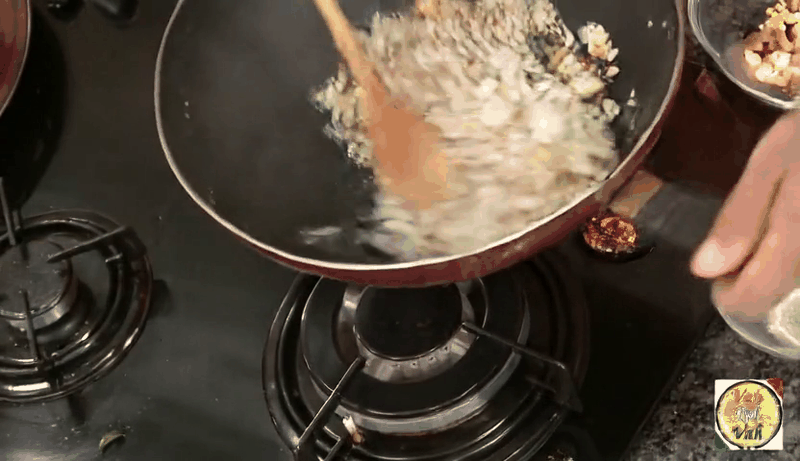} &
        \includegraphics[width=0.2\linewidth]{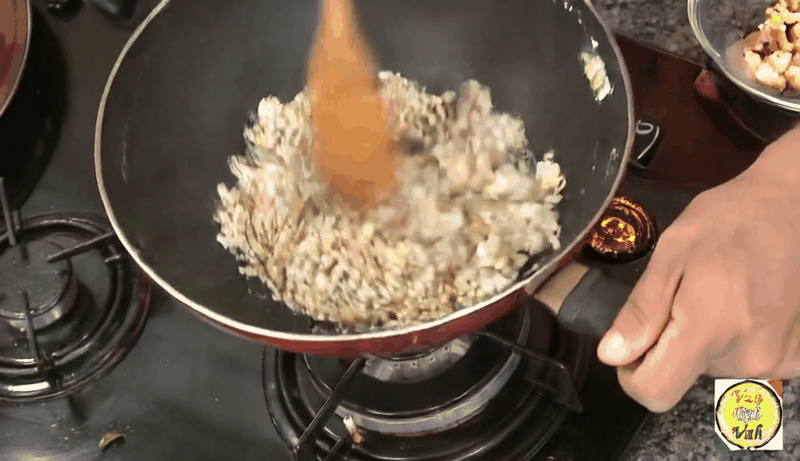} &
        \includegraphics[width=0.2\linewidth]{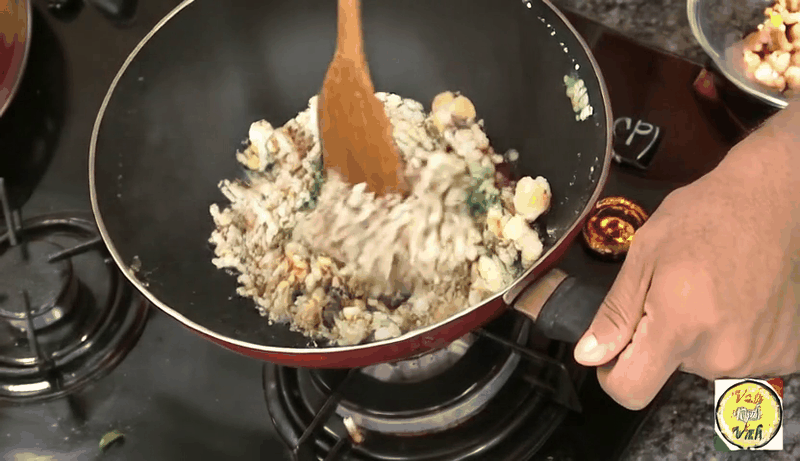} &
        \includegraphics[width=0.2\linewidth]{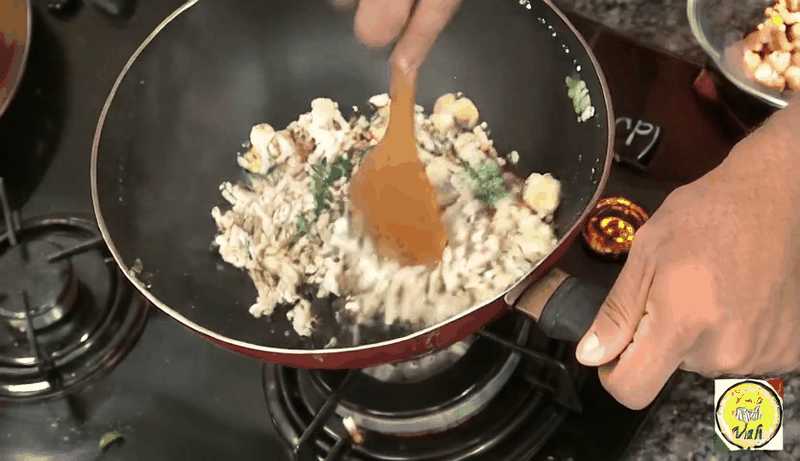} \\
        \multicolumn{5}{c}{\small\textbf{Seed 2}} \\
        \includegraphics[width=0.2\linewidth]{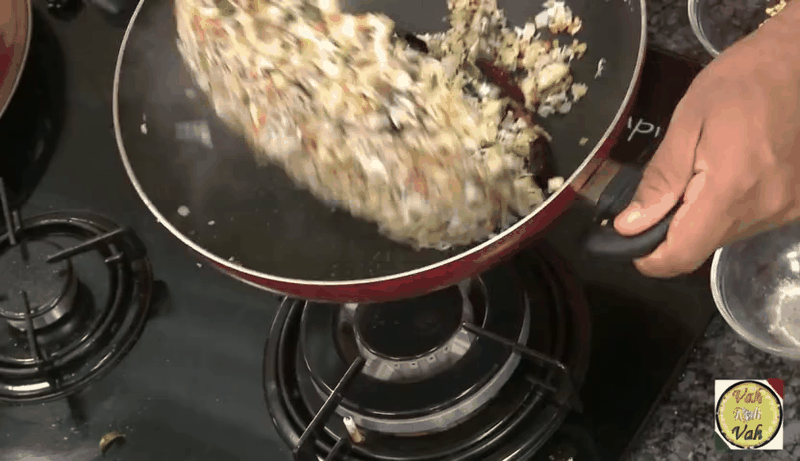} &
        \includegraphics[width=0.2\linewidth]{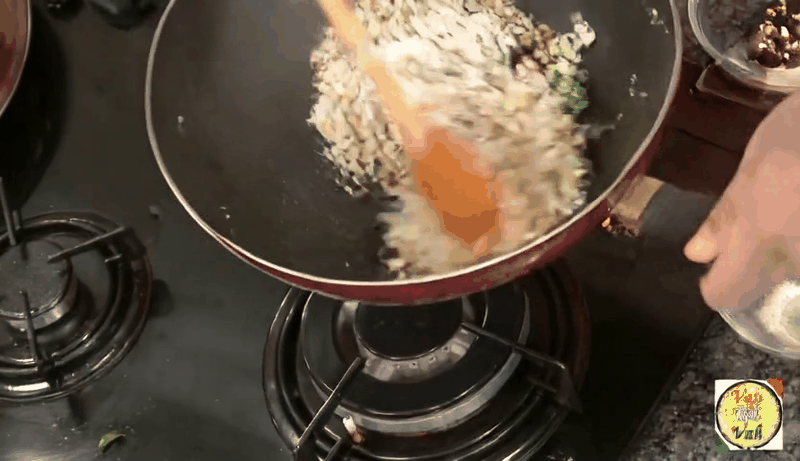} &
        \includegraphics[width=0.2\linewidth]{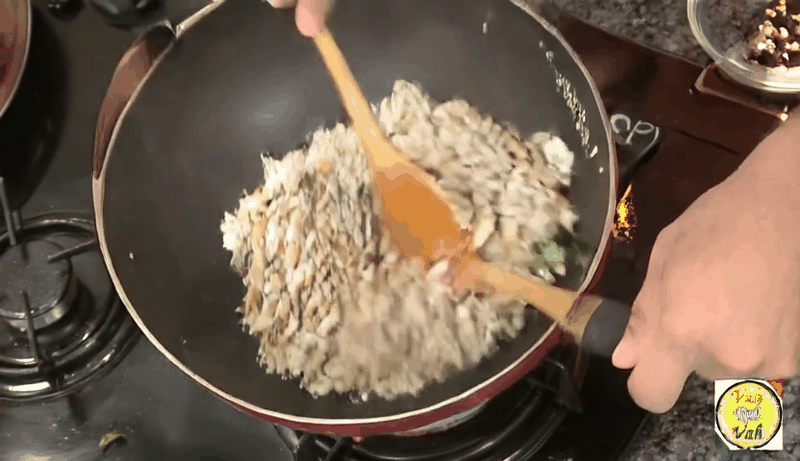} &
        \includegraphics[width=0.2\linewidth]{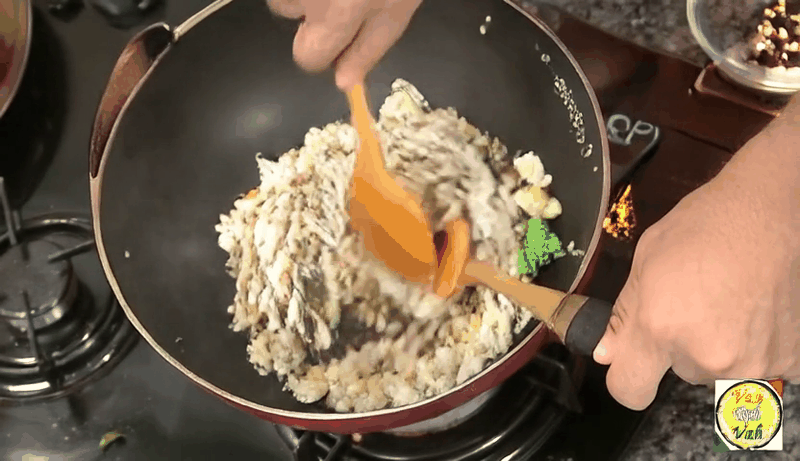} &
        \includegraphics[width=0.2\linewidth]{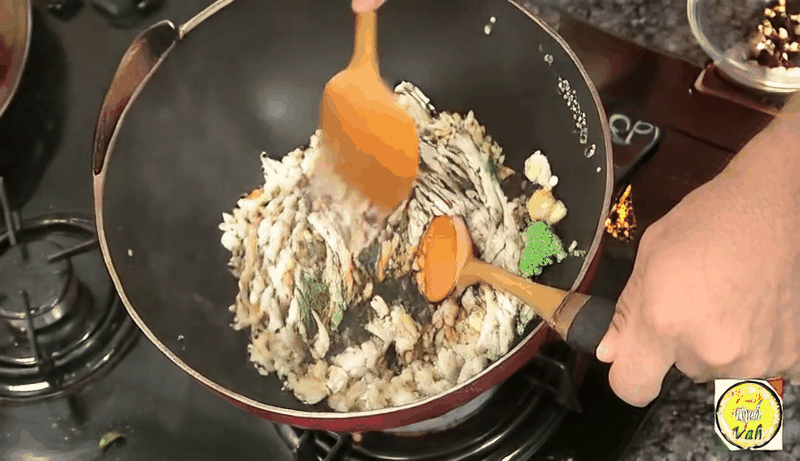} \\
        \multicolumn{5}{c}{\small\textbf{Seed 3}} \\
        \includegraphics[width=0.2\linewidth]{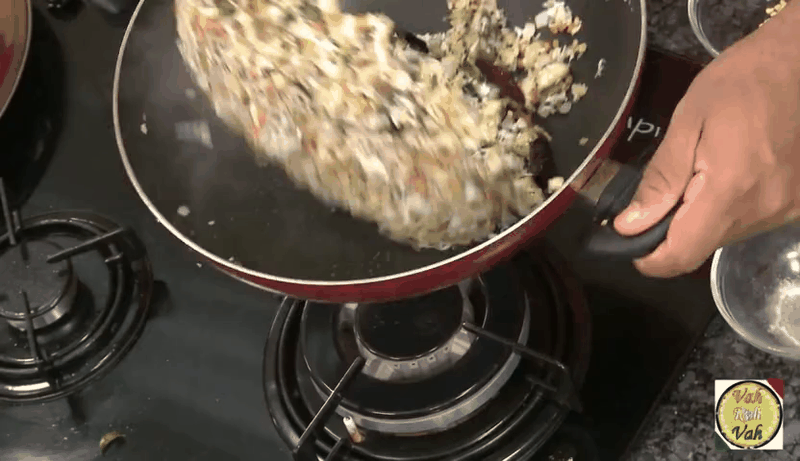} &
        \includegraphics[width=0.2\linewidth]{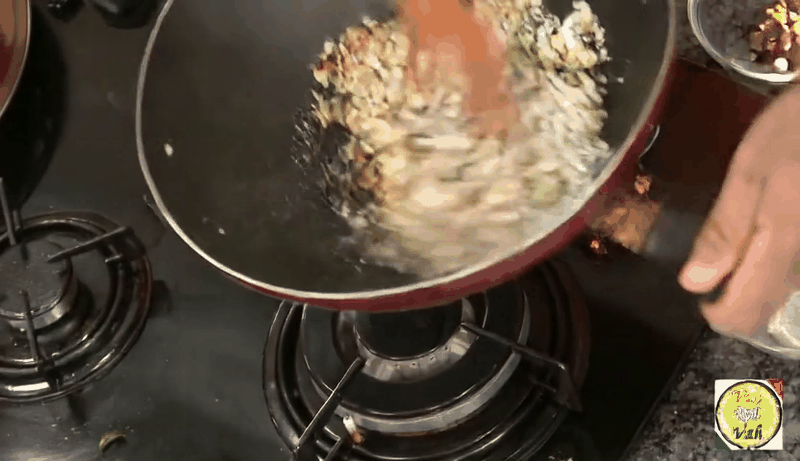} &
        \includegraphics[width=0.2\linewidth]{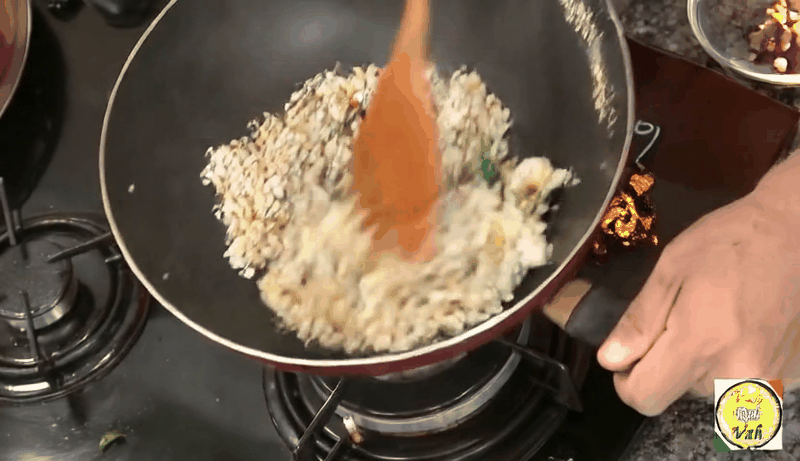} &
        \includegraphics[width=0.2\linewidth]{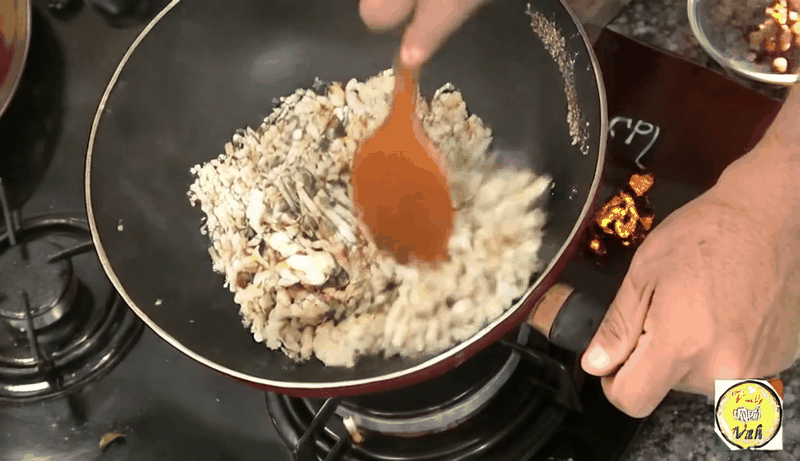} &
        \includegraphics[width=0.2\linewidth]{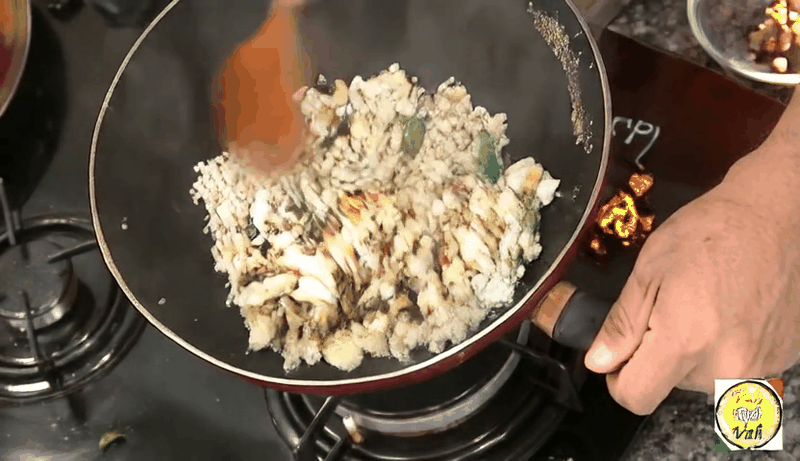} \\
        \multicolumn{5}{c}{\small\textbf{Seed 4}} \\
        \includegraphics[width=0.2\linewidth]{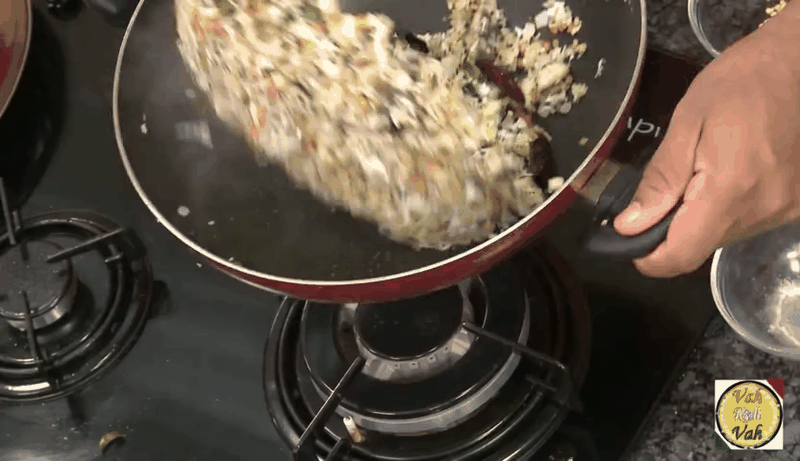} &
        \includegraphics[width=0.2\linewidth]{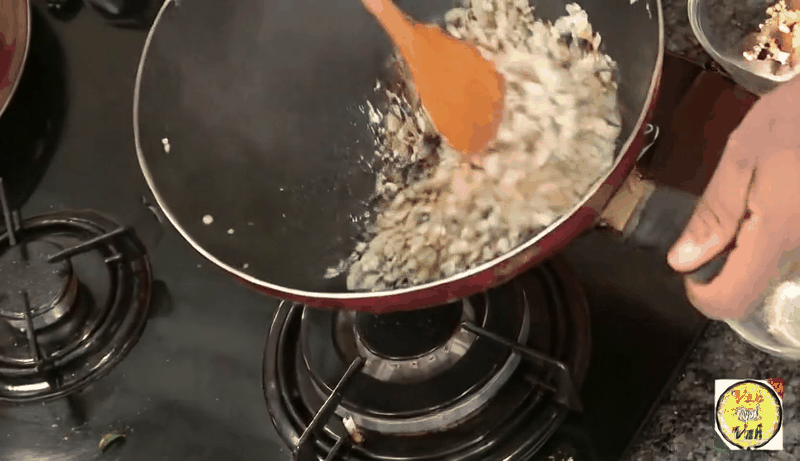} &
        \includegraphics[width=0.2\linewidth]{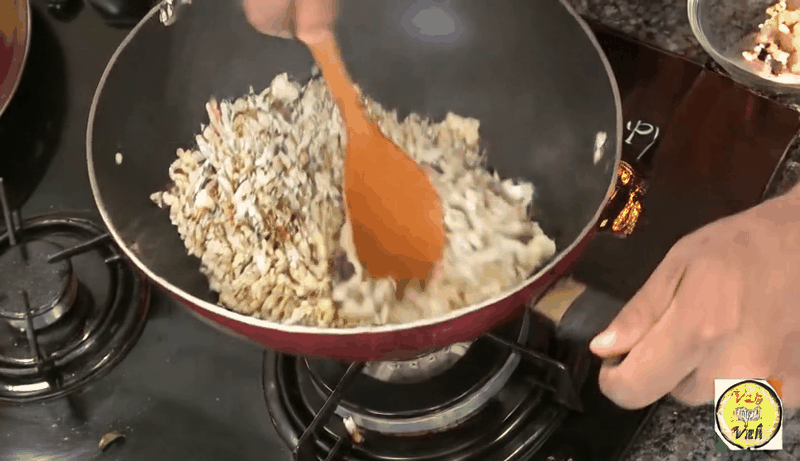} &
        \includegraphics[width=0.2\linewidth]{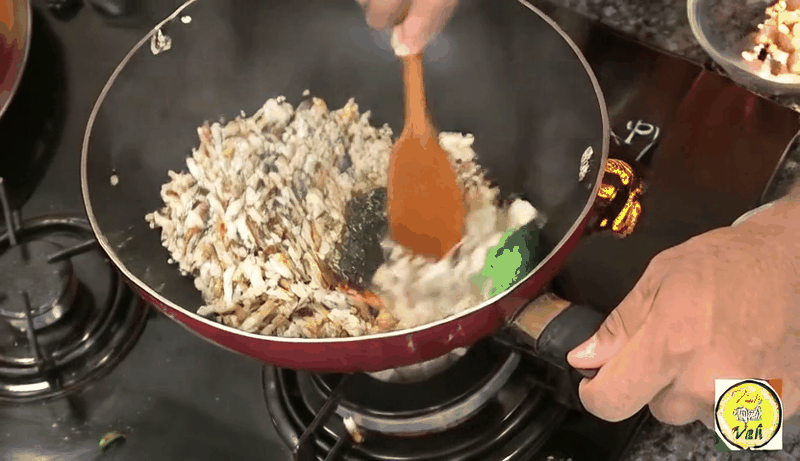} &
        \includegraphics[width=0.2\linewidth]{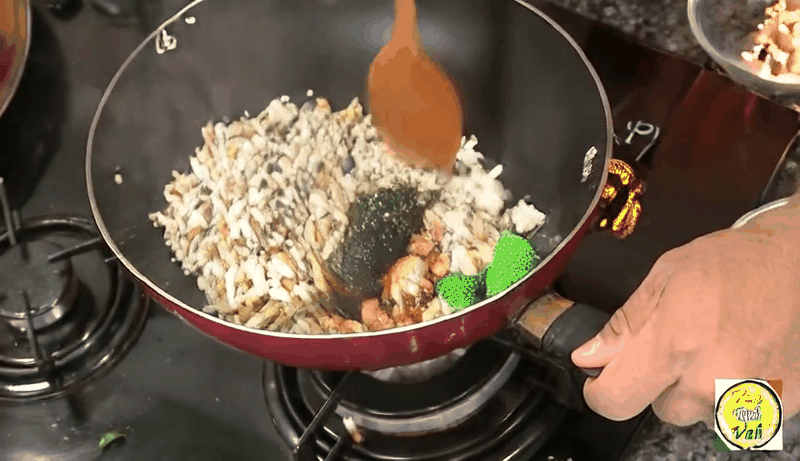} \\
    \end{tabular}
    \caption{\textbf{Generated frames across random seeds.} All five sequences share the same first frame and 3D tracks and differ only in seed. Trajectory-governed regions stay consistent, while underspecified content such as entering hands and occluded interiors varies.}
    \label{fig:multi_seed}
\end{figure}

To probe the stochasticity of generation in our distilled student model, we fix the first frame and the motion trajectories and vary only the random seed. \cref{fig:multi_seed} shows five videos generated with seeds 0 to 4 under identical conditioning. Regions governed directly by the input trajectories move almost identically across seeds, indicating that trajectory control is a strong and reliable spatial signal.

Variation instead concentrates where the conditioning is ambiguous or underspecified. Newly entering objects such as hands, whose appearance and timing the trajectories do not dictate, differ across seeds in shape, pose, and moment of entry. Occluded regions such as the interior of the cooking pan likewise settle into different fine-grained states, for instance the arrangement and appearance of the food, since neither the first frame nor the trajectories constrain them. This behavior is both expected and desirable: the model follows the provided control signals faithfully while using the seed to sample plausible completions for the remaining degrees of freedom, producing outputs that are diverse yet coherent.

\subsection{Effect of the Number of Control Points}\label{sec:supp:pointnum}

\begin{figure}[t!]
    \centering
    \includegraphics[width=\linewidth]{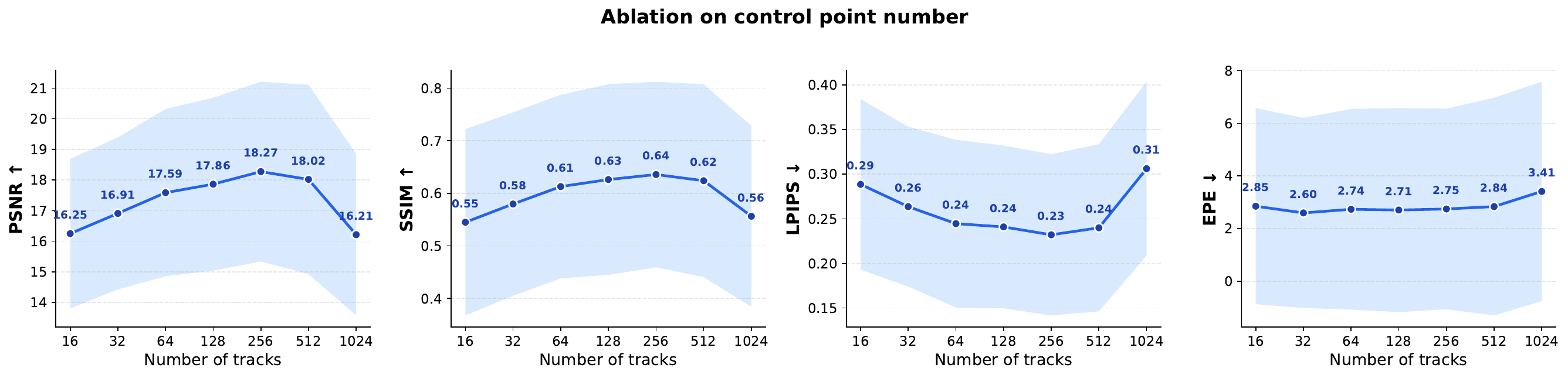}
    \caption{\textbf{Effect of the number of control points on student model generation.} PSNR, SSIM, LPIPS, and EPE as the inference-time track count varies from 16 to 1024. Quality peaks at the training count of 256.}
    \label{fig:pointnum}
\end{figure}

We study how the number of control tracks used at student inference affects generation quality and motion following, varying it over $\{16, 32, 64, 128, 256, 512, 1024\}$ with all other settings fixed (\cref{fig:pointnum}).

From 16 to 256 tracks, quality improves consistently: PSNR rises from 16.25 to 18.27, SSIM from 0.55 to 0.64, and LPIPS falls from 0.29 to 0.23, so denser control points provide richer spatial guidance. EPE stays stable across this range, fluctuating between 2.60 and 2.75.

Beyond 256 tracks all metrics degrade, moderately at 512 (PSNR 18.02, SSIM 0.62) and sharply at 1024 (PSNR 16.21, LPIPS 0.31, EPE 3.41). Since the model is trained with a fixed 256 tracks, this is expected: inference-time counts far from the training distribution give suboptimal results. The degradation is more severe above the training count than below it, as the model handles sparser subsets of the learned conditioning pattern more gracefully. We therefore use 256 tracks by default in all other experiments, matching the training configuration and giving the best balance of quality and motion accuracy.

\subsection{Effect of Attention Sink}\label{sec:attention_sink}

Our autoregressive generation uses a sliding context window, so earlier tokens are progressively evicted from the KV cache as the sequence extends. To keep a stable reference, we retain the initial token as a persistent anchor throughout generation.

\begin{figure}[t!]
    \centering
    \setlength{\tabcolsep}{0pt}
    \renewcommand{\arraystretch}{0.8}
    \begin{tabular}{cccc}
        \multicolumn{4}{c}{\small\textbf{With attention sink}} \\
        \includegraphics[width=0.25\linewidth]{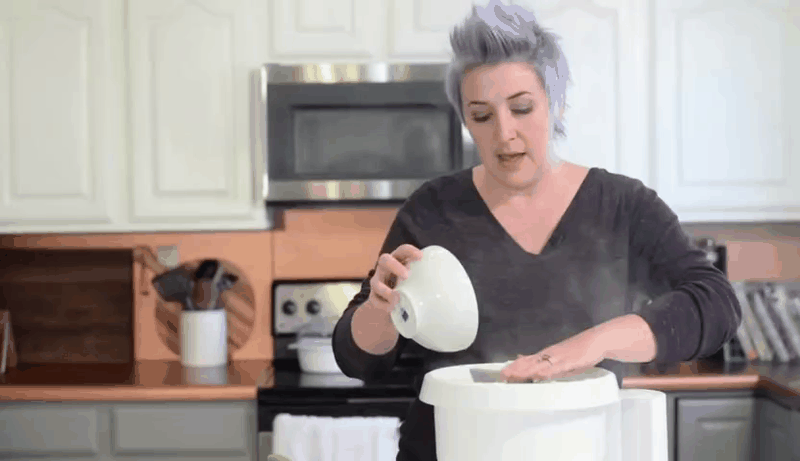} &
        \includegraphics[width=0.25\linewidth]{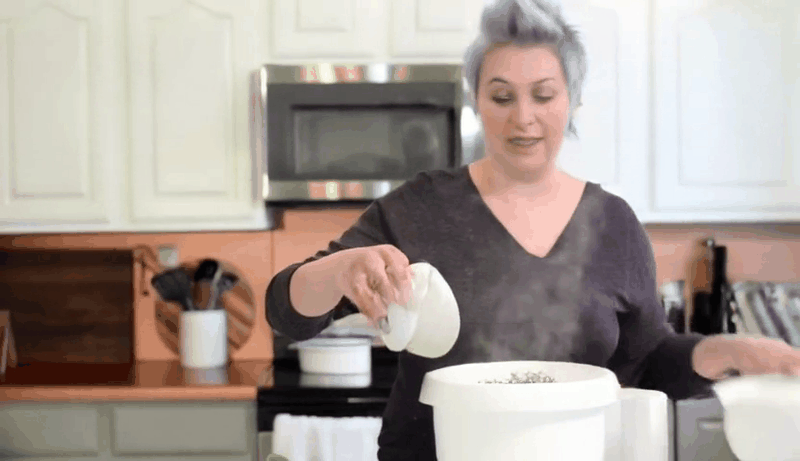} &
        \includegraphics[width=0.25\linewidth]{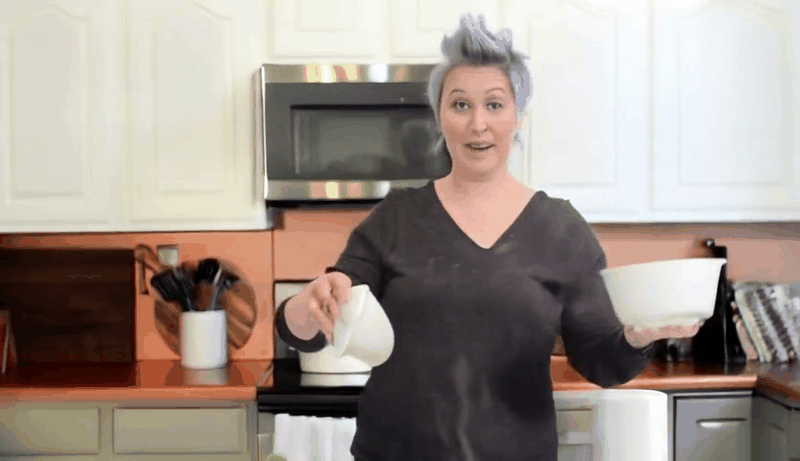} &
        \includegraphics[width=0.25\linewidth]{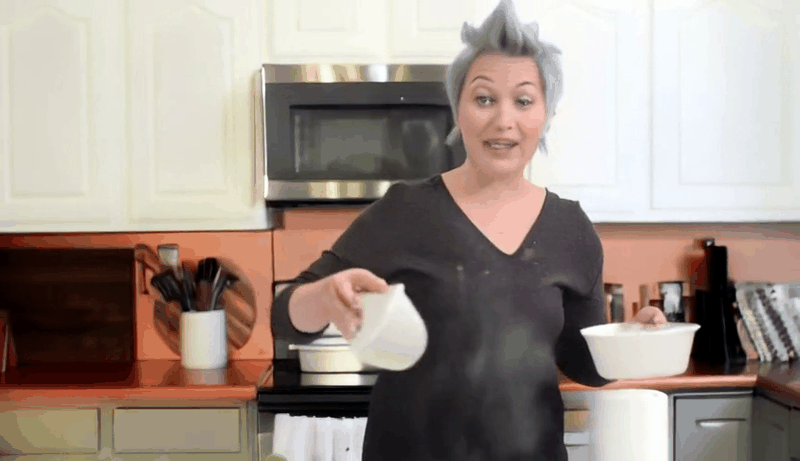} \\
        \multicolumn{4}{c}{\small\textbf{Without attention sink}} \\
        \includegraphics[width=0.25\linewidth]{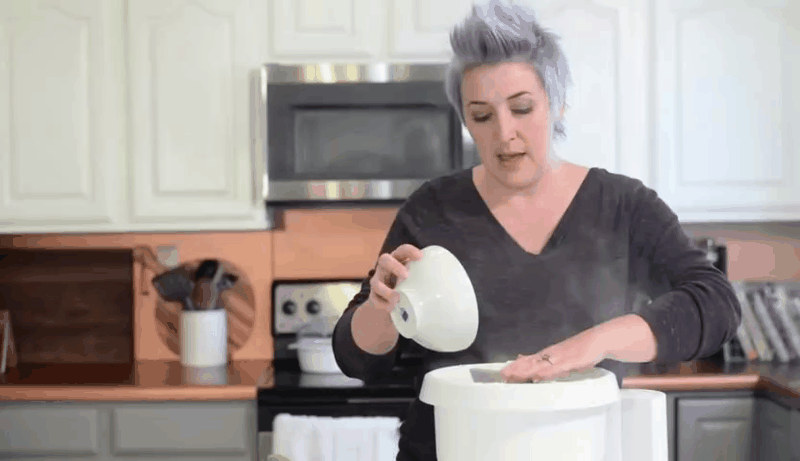} &
        \includegraphics[width=0.25\linewidth]{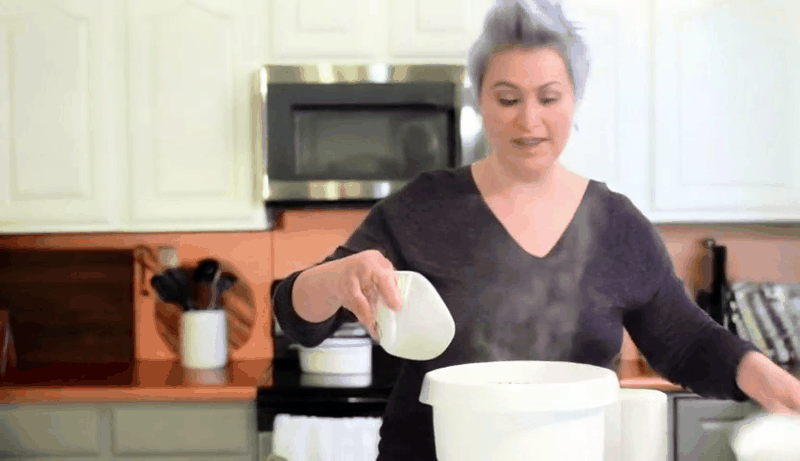} &
        \includegraphics[width=0.25\linewidth]{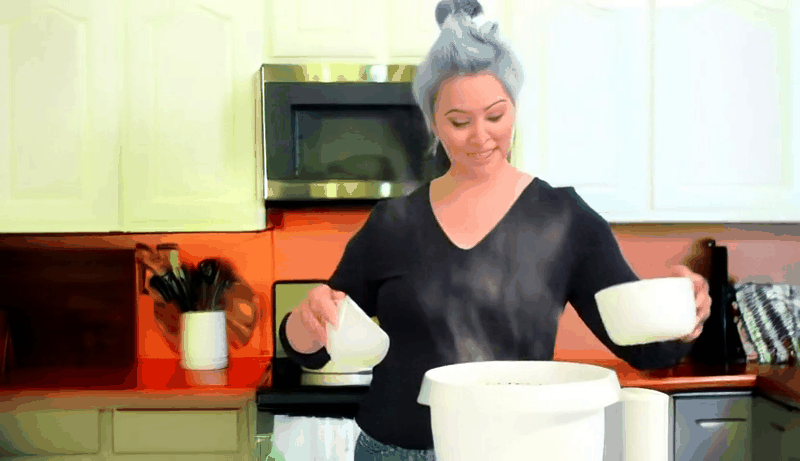} &
        \includegraphics[width=0.25\linewidth]{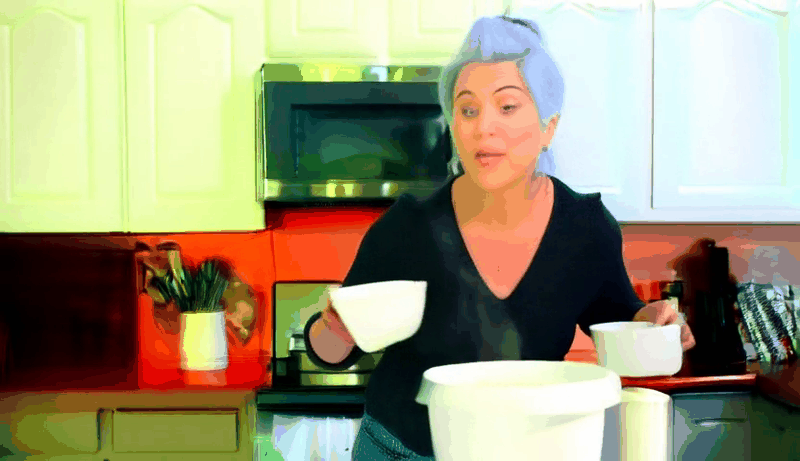} \\
    \end{tabular}
    \caption{\textbf{Effect of the attention sink.} Frames sampled at equal intervals from a 350-frame generated video. Without the sink, color and identity drift as the sequence extends.}
    \label{fig:attention_sink}
\end{figure}

\cref{fig:attention_sink} shows the effect. Without the sink, frames drift cumulatively: color saturation intensifies, facial identity departs from the initial appearance, and fine detail degrades into an overly smooth rendering. With the sink retained, the model holds color, identity, and visual quality stable across the entire sequence.

\section{Limitations and Future Work}\label{sec:limitations}

\paragraph{Dependency on monocular 3D estimation.}
Our data pipeline relies on monocular 3D point tracking \citep{xiao2025spatialtrackerv2} for scalable supervision without calibrated sensors. This enables large-scale curation from internet video, but the resulting 3D labels carry inherent scale ambiguity and grow less reliable under rapid motion or severe occlusion. Data diversity absorbs most of this during training, though it can surface at inference as subtle depth-scale inconsistency under extreme camera baselines. Adding multi-view geometric constraints \citep{schonberger2016colmap,wang2024dust3r} or metric depth priors \citep{yin2023metric3d} during curation could sharpen the supervision.

\paragraph{Fixed-density trajectory representation.}
Our $32{\times}32$ track grid balances compute against spatial coverage for most scenarios. Scenes with highly localized fine-grained interaction, such as dexterous hand manipulation or dense multi-body contact, would benefit from spatially adaptive track allocation. Extending the interface to variable-density or hierarchical layouts is a natural direction that could improve controllability in such cases without raising the average budget.

\paragraph{Streaming student versus teacher gap.}
The causal student achieves a $12.5\times$ speedup with only modest quality loss relative to the bidirectional teacher (about 0.5\,dB PSNR and a marginal EPE increase on DAVIS, \cref{tab:davis}). The fixed local attention window suffices for most motion patterns in our evaluation, though dependencies reaching well beyond it could be captured better with adaptive memory \citep{xiao2023streamingllm,zhang2024h2o}. We view closing this gap as an engineering challenge rather than a fundamental limit of the framework.

\paragraph{Future directions.}
Building on the current system, we see several promising extensions:
\begin{itemize}[nosep,leftmargin=*]
    \item \textbf{Closed-loop perception and generation.} Integrating real-time visual trackers \citep{karaev2024cotracker3} that observe the generated output and feed corrective 3D tracks back into the conditioning pipeline would give self-correcting generation, analogous to model-predictive control, compensating for drift without user intervention.
    \item \textbf{Physics-informed trajectory synthesis.} Coupling our interface with differentiable physics engines \citep{hu2019difftaichi} or learned dynamics models \citep{li2019learning} could generate physically plausible trajectories directly from high-level intent, reducing the manual effort of specifying control signals.
    \item \textbf{Embodied AI deployment.} Our streaming architecture interfaces naturally with robotic planning stacks, where action-conditioned 3D motion goals must be rendered as photorealistic futures for decision-making \citep{du2024learning,yang2024unisim,black2024pi0}. Extending \modelname into a real-time visual imagination module for model-based policy learning is a compelling direction.
    \item \textbf{4D-consistent multi-view generation.} Our 3D track conditioning enforces single-view geometric consistency; combining it with multi-view diffusion \citep{shi2024mvdream} or 4D scene representations \citep{wu20244dgaussians} could yield generators with strict cross-view consistency, enabling VR and AR content creation and digital twin synthesis.
    \item \textbf{Semantic trajectory understanding.} Trajectories currently encode pure kinematics without causal semantics. Incorporating world-model reasoning \citep{pmlr-v235-bruce24a,ha2018worldmodels} that couples ``where to move'' with ``why to move'' could improve physical plausibility and enable intent-level control beyond explicit point specification.
\end{itemize}

\end{document}